\documentclass[twoside]{article}
\usepackage{tikz,bm,algorithmic,algorithm}
\newcommand{\E}{\mathbb{E}}
\newcommand{\R}{\mathbb{R}}

\newcommand{\prob}{\mathbb{P}}
 
\newcommand{\greedy}{\text{greedy}}
\newcommand{\Greedy}{\bm\pi^{\text{\greedy}}}

\usepackage{array}    
\newcolumntype{K}[1]{>{\raggedright\arraybackslash}p{#1}}
\usepackage{microtype}
\usepackage{graphicx}
\usepackage{subfigure}
\usepackage{booktabs, enumitem} 
\usepackage{hyperref,natbib}
\hypersetup{
    colorlinks=true,       
    linkcolor=blue,        
    citecolor=blue,        
    filecolor=blue,        
    urlcolor=blue          
}

 \usepackage[accepted]{aistats2026}

\usepackage{amsmath}
\allowdisplaybreaks
\usepackage{amssymb}
\usepackage{mathtools}
\usepackage{amsthm}

 \usepackage[capitalize,noabbrev]{cleveref}
\usepackage{thm-restate}
\theoremstyle{plain}
\newtheorem{theorem}{Theorem}[section]

\newtheorem{lemma}[theorem]{Lemma}

\theoremstyle{definition}
\newtheorem{definition}[theorem]{Definition}
\newtheorem{assumption}[theorem]{Assumption}
\makeatletter
\def\th@remark{%
  \thm@headfont{\bfseries}
  \normalfont              
}
\makeatother
\newtheorem{remark}[theorem]{Remark}
\usepackage[textsize=tiny]{todonotes}

\begin{document}
\twocolumn[

\aistatstitle{From Restless to Contextual: A Thresholding Bandit Reformulation for 
Finite-horizon Improvement}

\aistatsauthor{ Jiamin Xu \And Ivan Nazarov\And Aditya Rastogi\And África Periáñez \And Kyra Gan }

\aistatsaddress{ Cornell Tech\And Causal Foundry\And Causal Foundry\And Causal Foundry\And  Cornell Tech} ]

\begin{abstract}
This paper addresses the poor finite-horizon performance of existing online \emph{restless bandit} (RB) algorithms, which stems from the prohibitive sample complexity of learning a full  \emph{Markov decision process} (MDP) for each agent.
We argue that superior finite-horizon performance requires \emph{rapid convergence} to a \emph{high-quality} policy. 
Thus motivated, we introduce a reformulation of online RBs as a
\emph{budgeted thresholding contextual bandit}, which simplifies the learning problem by
encoding long-term state transitions into a scalar reward.
We prove
the first non-asymptotic optimality of an oracle policy for a simplified finite-horizon setting.
We 
propose a practical learning policy under a heterogeneous-agent, multi-state setting, and show that it achieves 
a sublinear regret, achieving \emph{faster convergence} than existing methods. 
This 
directly translates to higher cumulative reward, as empirically validated by significant gains over state-of-the-art algorithms in large-scale heterogeneous environments. The code is provided in \href{https://github.com/jamie01713/EGT}{github}.
Our work provides a new pathway for achieving practical, sample-efficient learning in finite-horizon RBs.
\end{abstract}
 \section{Introduction}\label{sec:introduction}
Restless bandits 
extend classic contextual bandits by modeling state transitions and incorporating per-step budget constraints.
In a RB, 
a set of independent arms (agents) evolves over discrete time steps as MDP, with the goal of
dynamically allocating a fixed number of active plays to maximize cumulative rewards.
However, this modeling power comes with significant learning challenges. Even in 
the \emph{offline} setting with known model parameters, RB is PSPACE-hard \citep{b0e74184-2114-3e45-b092-dfbc8fefcf91}.
This computational intractability has directed much of the research toward policies that provide  \emph{asymptotic guarantees}, such as Whittle's Index Policy (WIP) \citep{whittle1988restless, Weber_Weiss_1990, 10.1109/ISIT.2018.8437712,7852321} and linear programming-based methods~\citep{verloop2016asymptotically,zhang2021restless,brown2023fluid,hong2024achievingexponentialasymptoticoptimality,gast2024linear,xiong2022reinforcement}, which are proven to approach
optimal as the number of agents approaches infinity.

This complexity carries over into the \emph{online} setting, where transition dynamics are unknown. Prior online RB algorithms largely seek to learn the optimal policy by first estimating the full underlying MDP for each agent. 
Consequently, online RB methods are largely inspired by, and mirror the standard frameworks of, \emph{reinforcement learning} (RL). 

They predominantly fall into two canonical RL categories. \emph{Model-based} approaches explicitly learn transition probabilities, typically within either \emph{episodic finite-horizon} \citep{jung2019regret,wang2023optimistic,liang2024bayesian} or infinite-horizon \citep{10.1007/978-3-642-34106-9_19, wang2020restless, xiong2022learning, akbarzadeh2023learning,xiong2022reinforcement} frameworks. \emph{Model-free} methods focus on learning the Q-function specifically in the infinite-horizon setting \citep{killian2021q, xiong2023finite,xiong2024whittle, kakarapalli2024faster}. A common strategy in both categories is to then integrate these learned models with WIP for decision-making.\footnote{We note that while LP-based methods are powerful in the offline setting, they are generally more difficult to adapt for online learning.} These online methods are designed to converge to their asymptotically optimal offline counterparts when agents are \emph{homogeneous}, where the \emph{convergence rate} is measured by the notion of \emph{regret} (formalized in Sec.~\ref{subsec:regret-convergence}).

This reliance on standard RL frameworks means that the performance guarantees for these online algorithms are inherently \emph{asymptotic} in the number of episodes or time steps.
However, this focus constitutes a fundamental limitation in practical applications,\footnote{Table~\ref{tab:examples} provides key application areas for RBs.} where horizons are 
finite due to budget cycles, program durations, or policy deadlines \citep{delrío2024, delrío2024Pharmacy}.
In domains like targeted patient interventions \citep{villar2015multi, biswas2021learn, mate2021} or resource allocation \citep{borkar2017opportunistic, brown2020index}, systems evolve continuously and cannot be reset, making the standard episodic RL framework impractical. 

Consequently, an algorithm's performance is judged by its cumulative reward within a \emph{single, limited-time episode}.
An asymptotically optimal policy that requires a long time to learn offers no practical value \emph{if it fails to achieve a high reward within the operational horizon}.
This critical limitation of prior work motivates a fundamental shift away from the objective of learning complex MDPs, and towards a reformulation designed for sample-efficient learning in the non-episodic, finite-horizon setting.
{Our contributions are fourfold}:
\begin{itemize}[leftmargin=12pt, itemsep=3pt, topsep=0pt, partopsep=0pt, parsep=1pt]
    \item \textbf{A Novel Reformulation}: we interpret regret as convergence rate (Sec.~\ref{subsec:regret-convergence}) and show that online algorithms may achieve higher cumulative reward in a finite horizon by
    targeting alternative design objectives, which leads to accelerated convergence (Sec.~\ref{subsec:need-for-reformulation}). This motivates us to introduce
a \emph{budgeted thresholding contextual multi-armed bandit} (BT-CMAB), which simplifies the learning objective of RBs to a contextual bandit problem through carefully designed rewards (Sec.~\ref{subsec:contextual_bandit}), enabling sample-efficient learning in finite horizons.
\item \textbf{First Non-asymptotic Finite-Horizon Optimality Guarantee}:
we establish the first finite-horizon non-asymptotic (valid for an arbitrary number of agents) optimality guarantee for a tractable RB setting, proving our oracle greedy policy is optimal for a
homogeneous 2-state model
(Theorem \ref{th:optimality-greedy-homogeneous}).
This provides a
theoretical foundation absent in prior work.

\item \textbf{A Fast-Converging Learning Algorithm}: 
we propose $\epsilon$-GT (\emph{$\epsilon$-Greedy Thresholding}, Algorithm~\ref{alg:lcb-greedy}, Sec.~\ref{section:thresholding-bandit-algorithm}), a novel learning algorithm for RB
with \emph{heterogeneous agents}. We establish that it achieves sublinear convergence rate
(Theorem \ref{th:sublinearregret}), enabled by
the careful choice of decaying $\epsilon$. 

\item \textbf{Empirical Validation}: we numerically demonstrate (in Section \ref{sec:experiments}) that our algorithm consistently outperform existing 
algorithms 
in various finite-horizon online RB settings across a wide range of instances \emph{with heterogeneous agents}.
\end{itemize}

\subsection{Additional Related Work}

\textbf{Thresholding Bandit/MDP\;\;} 
Our paper relates to regret-minimizing thresholding bandit problems, where regret is defined by the gap between the arm's mean and a predefined threshold when the mean falls below the threshold.
Under this setting, minimax optimal (max is taken over all possible bandit instances and min taken over all possible learning algorithms) constant regret is achievable \citep{tamatsukuri2019guaranteed,michel2022regret,feng2024satisficing}. 
Our problem setup differs from the traditional thresholding bandit framework by incorporating context, where the arrival of contexts follows an MDP.
In contrast, \citet{hajiabolhassan2023online} study threshold MDPs under an infinite horizon setting, seeking policies with average rewards exceeding a predefined threshold. We, however, focus on the finite-horizon setting.

\textbf{Episodic MDP\;\;} 
RB can be viewed as a single, massive MDP with a state space that is the combinatorial product of all agent states.
Thus, our work also relates to episodic MDP, which focuses on maximizing cumulative rewards across 
a finite horizon. Let $K$ be the number of episodes, each containing $T$ horizon. 
\citet{azar2017minimax} establishes a minimax regret of $\mathcal{O}(\sqrt{T^2K})$ for stationary transitions, and \citet{jin2018q} derives a regret of $\mathcal{O}(\sqrt{T^4K})$ for nonstationary transitions. 

\textbf{MAB Experimental Design\;\;} Our incremental reward explicitly incorporates the trade-off between reward maximization and treatment effect estimation (i.e., estimating the difference in rewards under $A=0$ and $A=1$), previously studied in MAB with Pareto-optimal algorithms provided \citep{doi:10.1287/mnsc.2023.00492,zhong2021achieving,zhang2024online}. Unlike prior work that treats these aspects separately, ours inherently combines both aspects through \emph{reward} that depends on intervention (i.e., $A=1$) and non-intervention (i.e., $A=0$) \emph{outcomes}.

\section{Finite-horizon Restless Bandits and Our 
Reformulation}\label{section:restless-bandit}
We introduce restless bandits in Sec.~\ref{subsec:online-restless-bandit} and reinterpret regret as convergence in Sec.~\ref{subsec:regret-convergence}. Sec.~\ref{subsec:need-for-reformulation} argues that better finite-sample performance requires new design objectives, while maintaining reward-based evaluation. We present our reformulation in Sec.~\ref{subsec:contextual_bandit}.



\subsection{Restless Bandits}\label{subsec:online-restless-bandit}
In a finite-horizon RB problem, the learner is provided with $M$ \emph{independent} arms, each representing one independent agent. 
%
Each agent $m$ is modeled as an
MDP represented by the tuple $\left(\mathcal{S},\mathcal{A},P_a^m,R^m\right)$, where 
$P_a^m(s,s'):\mathcal{S}\times \mathcal{S}\to[0,1]$ represents the transition probability from state $s$ to $s'$ under action $a$ for agent $m$,
and $R^m(s):\mathcal{S}\to\R$ represents the \emph{expected reward} that agent $m$ receives at state $s$. We 
focus on binary action,
i.e., $\mathcal{A}=\{0,1\}$. We highlight that in classical RBs,
the \emph{expected} reward, $R^m(s)$, is {independent} of the action, and the learner observes the reward \emph{immediately after arriving at} the state.

At each time $t$, given
the state $\bm{s}_t$, 
the learner's decision $\bm{a}_{t+1}$ follows a history-dependent random policy, $\pi_{t+1}$. 
Let
$\mathcal{H}_{t}:=\{(\bm{a}_{i}, \bm{s}_i, \bm{r}_i)\}_{i=1}^{t} \cup \{\bm s_0, \bm r_0\}$ be the history observed up to time $t$, and let $\mathcal{H}_{t}^+:=\mathcal{H}_{t} \cup \{\bm a_{t+1}\}$ be the augmented history up to taking action $\bm a_{t+1}$. 
Similarly, denote $\mathcal{F}_t$ as the $\sigma$-algebra generated by $\mathcal{H}_t$ and $\mathcal{F}_t^+$ as the $\sigma$-algebra generated by $\mathcal{H}_t^+$. 
Then,
$\pi_{t+1}:\mathcal{H}_{t}\to\Delta{\left(\{0,1\}^M\right)}$, where
$\Delta{\left(\{0,1\}^M\right)}$ denotes the probability simplex over all possible actions. The learner is provided with a budget constraint, allowing $B$ arms to be pulled at each time step,
i.e., $\sum_{m\in[M]} a_{t+1}^m= B$.
After pulling $B$ arms, the learner
observes the next state 
$s_{t+1}^m\sim P_{a_{t+1}^m}^m\left(s_t^m,s_{t+1}^m\right)$, and 
receives a feedback $\bm{r}_{t+1}$.
Suppose that the reward $\bm{r}_{t+1}$ is generated according to a distribution $\mathcal{P}_t$, we will assume stationarity of the \emph{expected} reward received at time $t+1$, $\bm{r}_{t+1}$, when conditioned on history $\mathcal{H}_{t}^+$ and the current state.
\begin{assumption}\label{assump:stationarity}
    $\E\left[r_{t+1}^m\mid s_{t+1}^m, \mathcal{H}_t^+\right]=R^m\left(s_{t+1}^m\right).$
\end{assumption}
We permit non-stationarity in the rewards, provided that Assumption~\ref{assump:stationarity} is satisfied, offering a relaxation of the standard stationarity assumption typically imposed in literature. 

The goal of the learner is to find the policy $\bm{\pi}:=\{\pi_t\}_{t\in[T]}$ that maximizes the expected cumulative reward, where the expectation is further taken over the randomized policy ($\bm{a}_{t}\sim \pi_{t}$), in addition to the states ($s_{t+1}^m\sim P^m_{a_{t+1}^m}(s_t^m,s_{t+1}^m)$), subject to the full budget utilization constraint:\footnote{Our model incorporates the standard full budget utilization constraint, consistent with the existing literature \citep{whittle1988restless,zhang2021restless,hong2024achievingexponentialasymptoticoptimality}.} 
\begin{align}
\vspace{-5pt}
    \max_{{\bm\pi}}\; &\E_{\bm{\pi}}\left[\sum_{m=1}^M\sum_{t=0}^TR^m\left(s_t^m\right)\mid\bm{s}_0
    \right]
     \notag\\
    \text{s.t.~}&\sum_{m=1}^M a_{t}^m= B,\;\forall t\in [T].\label{eq:optimization-problem-online}
\end{align}
We will use $\bm\pi^*$ to denote the optimal policy corresponding to Problem~\eqref{eq:optimization-problem-online}. Since $\bm\pi^*$ is intractable and unknown \citep{b0e74184-2114-3e45-b092-dfbc8fefcf91}, WIP \citep{whittle1988restless} and LP-based policies \citep{hong2024achievingexponentialasymptoticoptimality,xiong2022reinforcement} are proposed as approximations. These policies are proven to be asymptotically optimal as $M \to \infty$ for infinite-horizon rewards with homogeneous agents, a restrictive setting that often does not occur in practice. 
\subsection{Regret As Convergence Rate}\label{subsec:regret-convergence}
A critical shift occurs in online learning: \emph{the homogeneous agent assumption is often dropped to model more realistic, heterogeneous environments}. Under heterogeneity, these different approximation policies are no longer equivalent and will achieve different asymptotic performance levels. 
Consequently, comparing the regret magnitudes of different online algorithms is not meaningful when they target different reference policies (e.g., one converging to WIP versus another converging to an LP-based policy), as they are measuring convergence to fundamentally different benchmarks.
This reality reframes the meaning of regret in RBs. It is not an absolute measure of sub-optimality, but a relative measure of the \emph{convergence rate to a chosen reference policy $\bm\pi^e$}. We formalize this idea below:
\begin{definition}[Regret as Convergence Rate]\label{def:regret}
Let $\bm\pi^e$ be the reference policy that an online algorithm $\omega$ aims to converge to. We define the regret of $\omega$ relative to this reference policy over horizon $T$ as $\mathcal{R}_l^{\bm\pi^e}(\bm\omega)$:\footnote{Appendix~\ref{app:generalized_regret} extends this definition to episodic settings, which is common in online RBs.}
\begin{align*}
\mathcal{R}_l^{\bm\pi^e}(\bm{\omega}):&=
\E_{\bm{\pi}^{e}}\left[\sum_{m=1}^M\sum_{t=0}^Tl^m(s_t^m,a_{t+1}^m)\mid\bm{s}_0
    \right]\notag\\&-\E_{\bm{\omega}}\left[\sum_{m=1}^M\sum_{t=0}^Tl^m(s_t^m,a_{t+1}^m)\mid\bm{s}_0
    \right],
\end{align*}
where $l^m(s_t^m, a_{t+1}^m)$ serves as an instantaneous performance metric where the choice of $l^m$ depends on the objective of $\bm\pi^e$, which we elaborate below.
For heterogeneous agents with single-episode finite horizons, existing RB oracle policies are not optimal and hence not comparable. Thus, the regret $\mathcal{R}_l^{\bm{\pi}^e}(\bm{\omega})$ measures the \emph{convergence rate of the learning algorithm $\omega$  to its corresponding oracle (or reference) policy $\bm\pi^e$}, where sublinear regret implies that $\omega$ converges to $\bm\pi^e$ within sublinear time.
\end{definition}
The choice of the loss function $l^m$ is directly linked to the objective of $\bm\pi^e$.
For instance, in MAB, if $\bm\pi^e$ targets reward maximization, $l$ is immediate reward \citep{auer2002finite}; if $\bm\pi^e$ targets thresholding, $l$ penalizes action on arms below threshold \citep{feng2024satisficing}. Thus, $l$ always measures how well online algorithms are achieving the corresponding objectives of $\bm\pi^e$.
\begin{remark}
The interpretation of regret as a convergence rate appears implicitly in prior work. For example, \citet{wang2023optimistic} use Def.~\ref{def:regret} with $\bm\pi^e$ as WIP, while \citet{xiong2022reinforcement} use it with $\bm\pi^e$ as the infinite-horizon average-reward optimal policy.
\end{remark}
\subsection{The Need for Indirect Reward Maximization in Online RB}\label{subsec:need-for-reformulation}
Building on our reinterpretation of regret as convergence rate, we reconsider Problem~\eqref{eq:optimization-problem-online}'s reward maximization objective. In finite-horizon settings, algorithms targeting alternative objectives with faster convergence to high-quality reference policies may achieve superior empirical performance. This approach addresses the fundamental limitation of direct methods for online RB—including those learning $\bm\pi^*$, WIP \citep{jung2019regret,wang2023optimistic}, or LP-based policies \citep{xiong2022reinforcement}—which typically suffer from slow convergence in practice.\footnote{Although \citet{xiong2022reinforcement} claim sublinear convergence for finite-horizon RB, their analysis relies on a generative model that permits restarting the MDP from arbitrary states—an assumption rarely satisfied in practice. Without this, their algorithm would incur linear convergence.}


This slow convergence is well-characterized theoretically. For algorithms directly targeting reward maximization, \citet{auer2008near} establishes a $\Omega(T)$ lower bound when $\bm\pi^e=\bm\pi^*$ and $l^m=R^m$, while the tightest upper bound for learning WIP or LP-based policies remains $\mathcal{O}(T)$ \citep{wang2023optimistic}. This linear rate stems from the need to construct confidence intervals for \emph{transition probabilities}, which requires extensive exploration of state-action pairs $(s_t, a_t, s_{t+1})$ to achieve sufficient visitation counts.

Given these fundamental constraints on estimating transition probabilities, we pursue an alternative approach: proposing an oracle (or reference) policy that enables faster convergence by circumventing the need for direct transition probability estimation. We develop this in Sec.~\ref{subsec:contextual_bandit}, focusing on reference policies that avoid this bottleneck while maintaining strong \emph{empirical performance}.
\subsection{BT-CMAB
Reformulation}
\label{subsec:contextual_bandit}
Thus motivated, 
we reformulate restless bandit learning via two key modifications:
\begin{enumerate}[leftmargin=*, itemsep=0pt, topsep=0pt, partopsep=0pt, parsep=1pt]
    \item \emph{Contextual bandit framework}: Transition probabilities need not be explicitly estimated. It embeds nonstationarity into rewards, yielding policies that depend only on the current context. This reduces sample complexity when compared with MDPs \citep{simchi2022bypassing}.
    \item \emph{Thresholding objective}:
    At each step, we identify agents whose intervention benefit exceeds a threshold (vs. optimizing cumulative rewards), further lowering sample complexity \citep{feng2024satisficing} to achieve fast finite-horizon convergence.
\end{enumerate}
Under suitable assumptions, our approach with known
transitions/rewards recovers the optimal policy of \emph{offline} RB for any $M$ (Sec.~\ref{subsec:optimality}), unlike WIP's asymptotic optimality requiring $M,T\rightarrow\infty$.

\textbf{Regret-Minimizing BT-CMAB
Reduction\;\;}
In the reduced contextual bandit problem, 
the context space is the state space. 
At time $t$, the reward $\phi^m$ that agent $m$ receives at state $s_t^m$ under action $a_{t+1}^m$
is defined as the \emph{expected reward} that the agent will receive in the next time period:
\begin{align}
    \phi^m(s_t^m,a_{t+1}^m) &\coloneq \sum_{s_{t+1}^m\in\mathcal{S}}P^m_{a_{t+1}^m}(s_t^m,s_{t+1}^m)R^m\left(s_{t+1}^m\right)
    \notag \\
    &
    = \sum_{s_{t+1}^m\in\mathcal{S}}\prob\left(s_{t+1}^m\mid  \mathcal{H}_t^+\right) \E\left[r_{t+1}^m\mid s_{t+1}^m, \mathcal{H}_t^+\right]\notag\\
   & = \E\left[r_{t+1}^m\mid\mathcal{H}_t^+\right].
    \label{eq:c_reward}
\end{align}
The second equality
follows from conditional stationarity (Assump.~\ref{assump:stationarity}) and the Markovian property.
Eq.~\eqref{eq:c_reward} shows that the immediate reward, \(\phi^m(s_t^m, a_{t+1}^m)\), 
can be estimated without 
modeling the transitions,
consistent with the standard bandit reward formulation. 

As in online RBs, at each time $t$, the learner observes $\bm s_t$ and the corresponding reward $\bm {r}_t$.
The learner then updates the estimated uncertainty associated with the estimated $\phi^m(s_{t-1}^m, a_{t}^m) \;\forall m\in[M]$ according to Eq.~\eqref{eq:c_reward}, and uses it to guide the action $\bm a_{t+1}$, with $\sum_{m\in[M]} a_{t+1}^m= B$. 
%
%
%
The goal of the learner is to learn a policy that identifies and selects $B$ agents whose expected gain from taking the action (i.e., $a_{t+1}^m=1$) exceeds a predefined threshold, $\gamma$, in their current state $s_t^m$. To quantify this expected gain
in state $s\in\mathcal{S}$, we define $$I^m(s):=\phi^m(s,1)-\phi^m(s,0),$$ representing the \emph{incremental reward} an agent receives by taking the action versus not at state $s$.  
At time $t$, we say that an agent $m$ is in a good state if $I^m(s_t^m)\geq \gamma$. 

At each time $t$, let $G_t:=\{m\in[M]: I^m(s_t^m)\geq \gamma\}$ denote the set of agents currently in a good state. 
In this BT-CMAB
formulation, the optimal policy, $\bm\pi^b$, involves randomly selecting $B$ agents from $G_t$ for intervention 
when $|G_t|\geq B$ and selecting $B$ agents according to the ordering of incremental reward otherwise. 

In our reformulation, we define regret (Def.~\ref{def:regret}) for any policy $\bm\pi$ by setting the reference policy $\bm\pi^e=\bm\pi^b$ and using the loss function $l^m(s_t^m,a_{t+1}^m):=-c_t^m(a_{t+1}^m)$, where the cost function is:
\begin{equation}
    c_t^m\left(a_{t+1}^m\right):=\left(\gamma-I^m(s_t^m)\right)
    \mathbb{I}\left\{I^m(s_t^m)<\gamma,a_{t+1}^m=1\right\}.\label{eq:cost}
\end{equation}
This cost function has an intuitive interpretation: it penalizes allocations to agents in ``bad'' states (where $I^m(s_t^m)<\gamma$), with the penalty proportional to how far the state index falls below the threshold $\gamma$. The baseline policy $\bm\pi^b$, by construction, never incurs these costs since it only allocates to agents in good states.

We simplify notation by writing $\mathcal{R}_c^{\bm\pi^b}(\bm\pi)$ as $\mathcal{R}(\bm\pi)$. This regret directly measures how frequently $\bm\pi$ deviates from $\bm\pi^b$ by allocating to suboptimal states, thus quantifying the convergence rate to this reference policy. The learner's objective is to minimize cumulative regret $\mathcal{R}(\bm\pi)$ over horizon $T$, which equivalently maximizes the convergence rate to $\bm\pi^b$.

\section{Optimality of BT-CMAB
Reduction}\label{subsec:optimality}
Before establishing the sublinear convergence rate of our proposed algorithm for BT-CMAB with \emph{heterogeneous} agent, we first establish the optimality of our BT-CMAB formulation for 2-state homogeneous agents—the first non-asymptotic optimality result for finite-horizon RB. This demonstrates that the contextual bandit provides an exact reformulation of the offline RB problem under these conditions. \emph{We note that Sections~\ref{section:thresholding-bandit-algorithm} and \ref{sec:experiments} lift these assumptions.}

We assume \emph{binary states} and we further assume that all agents share a good state with higher $R^m$. Without loss of generality (WLOG), we will use 1 to denote the good state and 0 for the bad state. 
%
We first describe the greedy policy.
WLOG,
suppose that at each time step $t$, the incremental rewards are ordered decreasingly as $I^1\left(s_t^1\right)\geq I^2\left(s_t^2\right)\geq\cdots\geq I^M\left(s_t^M\right)$. The corresponding greedy policy $\pi_t^{\text{greedy}}: \mathcal{S}\rightarrow\{0,1\}^M$ is to give actions to the first $B$ agents according to the order. Formally, with a slight abuse of notation, let $\pi_t^{\text{greedy}}({s}_t^i)$ be the
$i$\textsuperscript{th} entry of $\pi_t^{\text{greedy}}$, defined as:
%
\begin{equation*}
 \pi_t^{\text{greedy}}({s}_t^i)=\begin{cases}
        1&i\leq B\\
        0&i>B
    \end{cases}.
\end{equation*}
Let $\bm\pi^{\text{greedy}}:=\{\pi_t^{\text{greedy}}\}_{t\in[T]}$.
We note that $\bm\pi^{\text{greedy}}=\bm{\pi}^b$ with suitable choices of $\gamma$. A trivial choice of $\gamma$ would be $\max_{m,s} I^m(s)$.
Now we are ready to show that the greedy policy is optimal for any $M$.
\begin{restatable}{theorem}{optimalityhomogeneous}\label{th:optimality-greedy-homogeneous}
Assuming all agents are homogeneous, i.e., $P^m=P^{m'}, R^m=R^{m'},\forall m,m'\in[M]$ and $\mathcal{S}=\{0,1\}$,
    $\bm{\pi}^{\text{greedy}}=\bm{\pi}^*$ for any $0\leq B\leq M<\infty$.
\end{restatable}
Theorem~\ref{th:optimality-greedy-homogeneous} establishes that our reference policy $\bm\pi^b$—which corresponds to the greedy policy $\bm\pi^{\text{greedy}}$ with appropriate threshold selection—achieves non-asymptotic optimality for homogeneous, 2-state finite-horizon restless bandits. This provides stronger guarantees than prior reference policies like WIP or LP-based methods \citep{hong2024achievingexponentialasymptoticoptimality,whittle1988restless,zhang2021restless}, which only offer asymptotic optimality for infinite-horizon settings under homogeneous agents.

While our analysis builds on the practical two-state assumption commonly used in recent work \citep{liang2024bayesian,killian2023robust,wang2023scalable}, proving Theorem~\ref{th:optimality-greedy-homogeneous} required developing novel analytical techniques. Existing tools are designed for analyzing value function-based policies (Eq.~\eqref{eq:value-function}) in infinite-horizon settings \citep{whittle1988restless,hong2024achievingexponentialasymptoticoptimality}, making them unsuitable for our problem setup.
\begin{remark}[Difficulty of Relaxation]\label{remark:significance_thm_optimality}
Theorem~\ref{th:optimality-greedy-homogeneous} does not extend to heterogeneous agents (Appendix~\ref{sec:counter-example-heterogeneous}) or multi-state systems (Appendix~\ref{sec:counter-three-state}). Our proof technique—which reduces Bellman optimality to conditions on value functions—relies critically on the two-state property that transition dynamics can be fully characterized by the probability of transitioning to state 1. This characterization breaks down in multi-state systems, where transitions require tracking probabilities to multiple states.
While we conjecture that appropriately designed greedy policies may achieve constant-factor approximations even in these more general settings, such extensions are beyond the scope of this paper. 
\end{remark}
\vspace{-5pt}
Although $\bm\pi^b$ is only guaranteed to be optimal for homogeneous agents and a high threshold, our BT-CMAB reduction is valid for heterogeneous agents and an arbitrary threshold $\gamma$. As mentioned in Section~\ref{subsec:contextual_bandit}, the threshold is associated with fast convergence. We will show in Section~\ref{subsec:theoretical-guarantee} that a higher threshold will lead to slower convergence for heterogeneous agents. The detailed discussion on how to choose the threshold is deferred to Remark~\ref{remark:trade-off-gamma}.

\textbf{Theorem \ref{th:optimality-greedy-homogeneous} Proof Sketch\;\;} The proof consists of
two steps: 1) We reformulate the problem as an MDP where the state space is the Cartesian product of the individual state space, i.e., $\mathcal{S}^M$; the action space is the subset of $\{0,1\}^M$ that
satisfies the budget constraint, and the transition matrix is the product of $P^m$ as all agents are independent. 2) We show that the proposed greedy policy $\pi_t^\greedy$ 
satisfies the Bellman 
equation (Eq.~\eqref{eq:bellman-optimality}) for all $t\leq T$ 
by using backward induction. 
We prove
the induction hypothesis 
by 
carefully controlling the difference of the value function (Eq.~\eqref{eq:value-function}).
\section{Learning in BT-CMAB}
\label{section:thresholding-bandit-algorithm}

We now present our learning algorithm (Algorithm \ref{alg:lcb-greedy}) for BT-CMAB
(as reformulated in Sec.~\ref{subsec:contextual_bandit}) in a general setting, and establish that it
achieves sublinear regret for this new problem. The sublinear regret bound explicitly characterizes the algorithm's convergence rate, demonstrating its fast convergence in finite sample
when compared with those of existing restless bandit algorithms.
We note that \emph{the results hold without
 agent homogeneity or  a binary state space}.
Learning in BT-CMAB framework is challenging because it requires estimating the incremental reward—the outcome difference between acting and not acting—which necessitates observing both actions for each agent-state pair (paired exploration).
However, to minimize regret, the policy must also exploit this knowledge by taking the optimal action (which could be either 0 or 1, depending on the state and agent). The tension between these two requirements—thorough exploration for accurate estimation and greedy exploitation for regret minimization—defines the core difficulty of the problem.

\textbf{$\epsilon$-Greedy Thresholding Algorithm\;\;} To resolve this, we propose an \emph{$\epsilon$-Greedy Thresholding algorithm}, Algorithm \ref{alg:lcb-greedy}, that explicitly balances this trade-off. By using a decaying exploration rate $\epsilon_t$, it ensures we sample both actions for each state-agent pair enough for learning while progressively prioritizing the optimal action to minimize regret, as formalized in our proof.


\begin{algorithm*}[!t]
  \caption{$\epsilon$-Greedy Thresholding ($\epsilon$-GT)}
    \label{alg:lcb-greedy}
    \begin{algorithmic}[1]
        \STATE{\bfseries Input:}  initial state vector $\bm{s}$, round budget $B$, exploration function $f(t)$,
        threshold $\gamma$, instance parameter $\eta$, scaling constant $D$, exploration decay rate $p$
        \STATE Initialize: initial states $\bm{s}_0\gets \bm{s}$; 
        $\forall (s, a, m) \in (\mathcal{S} \times \mathcal{A} \times [M]):$
        agent state visit count $N_{s,m}^{(0)}\gets 0$, agent state-action visit count $N_{s,m,a}^{(0)}\gets 0$, incremental reward
        $\widehat{I}_0^m(s)\gets 0$,
        cumulative immediate rewards $\hat{\phi}^m(s,a)\gets 0$,
        UCB of incremental reward $\mathrm{UCB}_0^m(s)\gets \infty$
        \FOR{$t=0, 1,\cdots,T$}
        \STATE Retrieve  
        $\bm s_t$; identify good agents above the threshold:
        $\widetilde{G}_{t+1} \gets \left\{m\in[M] \colon \widehat{I}_t^m({s}_t^m) \geq \gamma\right\}$.
        \STATE Initialize the candidate set $C\gets\widetilde{G}_{t+1}$;
        \STATE 
        Sort the remaining agents: \([M] \setminus \widetilde{G}_{t+1}\), in descending order of \(\text{UCB}_t^m(s_t^m)\); append top
        \(\max(0, B - |\widetilde{G}_{t+1}|)\) agents from sorted list to
        \(C\).
        \STATE $\widetilde{C} \gets B$ agents in $C$ with highest $\widehat{I}_{t}^m(s_t^m)$.
        \STATE With probability $\epsilon_t=\min\left\{1,\min_{m}\frac{D}{\left(N_{s_t^m,m}^{(t)}+1\right)^p\min\{\eta^2,1/2\}}\right\}$:
        \STATE $\quad$ Random exploration: uniformly choose $B$ agents from $[M]$ and set their 
        $a_{t+1}^m\gets1$.
        \STATE Otherwise greedy: set $a_{t+1}^m = 1$ for $m \in \widetilde{C}$, $0$ otherwise\;
            \STATE Observe $(\bm{s}_{t+1}, \bm r_{t+1})$ \;
        \STATE For each agent-state pair $(m,s)\in[M]\times\mathcal{S}$, update the UCB of the incremental rewards as:
\STATE \quad $\forall a\in\mathcal{A}: \hat{\phi}^m(s_t^m, a) = \hat{\phi}^m(s_t^m, a) +
        r^m_{t+1}\mathbb{I}\left\{a_{t+1}^m = a\right\} $.
        \STATE \quad 
        $\forall a\in\mathcal{A}: N_{s, m,a}^{(t+1)} = 
        N_{s, m,a}^{(t)} + \mathbb{I}
        \left\{a_{t+1}^m = a\right\}\mathbb{I}
        \left\{s = s_{t}^m \right\}$\;
       \STATE\quad $N_{s,m}^{(t+1)}=N_{s,m}^{(t)}+\mathbb{I}\{s=s_t^m\}$\;
        \STATE  \quad 
    $\widehat{I}_{t+1}^m(s)=\frac{\hat{\phi}^m(s, 1)}{N_{s, m,1}^{(t+1)}} - \frac{\hat{\phi}^m(s,0)}{N_{s,m,0}^{(t+1)}}$
        \STATE\quad$\mathrm{UCB}_{t+1}^m(s)=\widehat{I}_{t+1}^m(s)+\sqrt{\frac{f\left(N_{s,m,1}^{(t+1)},T\right)}{N_{s,m,1}^{(t+1)}+2}}+\sqrt{\frac{f\left(N_{s,m,0}^{(t+1)},T\right)}{N_{s,m,0}^{(t+1)}+2}}$
        \ENDFOR\;
    \end{algorithmic}
\end{algorithm*}
At each time step $t$, Algorithm~\ref{alg:lcb-greedy} proceeds as follows:
With probability $\epsilon_t$, the algorithm explores by uniformly selecting $B$ agents to act upon, regardless of their state or estimated reward. The exploration rate $\epsilon_t$ decays with the inverse of the minimum time an agent has spent in its current state, encouraging more exploration in under-sampled contexts. With probability $1-\epsilon_t$, the algorithm exploits its knowledge: it computes the empirical mean incremental reward for each agent in its current state, forms a candidate set $\widetilde{G}_{t+1}$ of agents exceeding the threshold $\gamma$. The algorithm then proceeds as follows: if
$|\widetilde{G}_{t+1}| \geq B$, it gives actions to the top-$B$ agents within $\widetilde{G}_{t+1}$ (e.g., those with the highest empirical means);
if $|\widetilde{G}_{t+1}| < B$, it gives actions to all agents in $\widetilde{G}_{t+1}$ and then allocates the remaining $B - |\widetilde{G}_{t+1}|$ actions by selecting the agents with the highest \emph{upper confidence bound} (UCB) on the incremental reward from the rest agents. After the actions are taken, the algorithm observes the new state vector $\bm{s}_{t+1}$ and the corresponding reward vector $\bm{r}_{t+1}$.

Next, the algorithm updates its internal estimates using the newly observed rewards and states. For each agent, we update the cumulative rewards and visitation counts for the state-action pairs that were just observed. These are used to compute a new empirical mean for the incremental reward, $\widehat{I}_{t+1}^m(s)$, for each agent-state pair. Finally, the upper confidence bound is recalculated. The new $\mathrm{UCB}_{t+1}^m(s)$ for the next round adds an optimism bonus to the empirical incremental reward. This bonus is derived from an anytime-valid confidence bound \citep{10.1214/20-AOS1991} and is inversely proportional to the square root of the state-action visit counts, ensuring that estimates for less-explored states have higher uncertainty. The specific function used is $f(t,T) = \log\log(2t + 0.72\log(10.4T))$.

As we address a thresholding contextual bandit with MDP-governed contexts instead of MAB, Alg.~\ref{alg:lcb-greedy} differs from \citet{auer2002finite}.
We tailor $\epsilon_t$ to decay with the minimum state occupancy time rather than the current timestep, ensuring sufficient exploration across states.

\subsection{Theoretical Guarantee}\label{subsec:theoretical-guarantee}
We establish that Algorithm~\ref{alg:lcb-greedy} achieves a $\mathcal{O}(\sqrt{T})$ regret upper bound under two assumptions.
 Assump.~\ref{assumption:regular-transition-matrix} ensures strictly positive transition probabilities (even if arbitrarily small), while Assump~\ref{asssumption:enough-good-arm} guarantees the existence of a policy with sufficient good agent-state pairs at each timestep.
\begin{assumption}[Positive Transition]\label{assumption:regular-transition-matrix}
\( P_a^m(s, s') > 0 \) for all \( s, s' \in \mathcal{S} \), \( a \in \mathcal{A} \), and \( m \in [M] \).
\end{assumption}
Assump.~\ref{assumption:regular-transition-matrix} guarantees that any state $\bm{s}' \in \mathcal{S}^M$ is reachable from any given state $\bm{s}$ in a single step. Importantly, we only require these one-step transition probabilities to be positive; we do not assume they are bounded away from zero, and our result \emph{does not} depend on the minimum transition probability.
\begin{assumption}[Exists A Good Policy]\label{asssumption:enough-good-arm}
    There exists a policy achieving zero cumulative cost.
\end{assumption}

Assump.~\ref{asssumption:enough-good-arm} serves primarily to simplify our theoretical analysis by ensuring problem instances are well-behaved. This condition guarantees that an oracle policy exists which can spend its full budget of $B$ actions per round without being forced to assign actions with negative expected returns.
In practice, this assumption
can be satisfied for any problem instance by carefully choosing the threshold parameter $\gamma$; for example, setting $\gamma = -\infty$ guarantees the condition is met trivially. Thus, $\gamma$ serves as a \emph{tunable parameter} to be optimized for overall performance, a point we discuss further in Remark~\ref{remark:trade-off-gamma}.  

 When combined with Assump.~\ref{assumption:regular-transition-matrix}, 
Assump.~\ref{asssumption:enough-good-arm} yields a stronger guarantee (Lemma~\ref{le:enough-good-arm-all-states})
 that is inherently assumed by any algorithm required to spend the full budget $B$ at each round. This is common in prior RB algorithms \citep{wang2020restless,wang2023optimistic}.

Let  $\mathcal{G}:=\{(m,s): m\in[M], s\in\mathcal{S}, I^m(s)\geq \gamma\}$ be the set of good agent-state pairs.
Lemma~\ref{le:enough-good-arm-all-states} ensures that for any state realization $\bm{s} \in \mathcal{S}^M$, there are always at least $B$ good agent-state pairs.
\begin{restatable}[Existence of Sufficient Good Agent-State Pairs]{lemma}{sufficientgoodarms}\label{le:enough-good-arm-all-states}
\vspace{-2pt}
   Under Assumptions \ref{asssumption:enough-good-arm} and \ref{assumption:regular-transition-matrix}, for any joint state $\bm{s} \in \mathcal{S}^M$, the number of good agent-state pairs observed in $\bm{s}$ is at least $B$. That is,
    $\left|\left\{m\in[M]:(m,s^m)\in \mathcal{G}\right\} \right|\geq B$.
\end{restatable}
Despite its strong guarantee, Lemma \ref{le:enough-good-arm-all-states} (proof in Appendix~\ref{proof:enough-good-arm-all-states}) is derived from two relatively mild assumptions. This result is critical for two reasons. First, it ensures that any algorithm required to spend a full budget of $B$ actions is never forced to assign actions whose incremental reward is below the threshold. Second, it guarantees that while choosing a suboptimal agent may affect future state distributions, this effect does not accumulate: sufficient good agents are always available.
Thus, a policy achieving zero per-step regret is attainable, which forms the foundation for our algorithm’s performance.

Let $\Delta_s^m := |\gamma - I^m(s)|$. Define $(m_{\min}, s_{\min})$ as the minimizer of $I^m(s) - \gamma$ over $(m,s) \in \mathcal{G}$, and set $\Delta := \Delta_{s_{\min}}^{m_{\min}}$.
Next, we present our main theorem
on the regret upper bound, which characterizes the finite-horizon convergence rate of Algorithm~\ref{alg:lcb-greedy}:
\begin{restatable}[Regret Upperbound]{theorem}{regretbound}\label{th:sublinearregret}
Assume $r_t^m$ is 1-subgaussian random variable for all $t\in[T]$. Let $D:=\max\left\{\frac{8M}{B},\frac{8M}{M-B},2\right\},p=1,\eta=\Delta$. 
Under Assump.s~\ref{asssumption:enough-good-arm} and \ref{assumption:regular-transition-matrix}, the regret, $\mathcal{R}(\pi)$,
 defined by Def.~\ref{def:regret} with the loss function $l^m$ given by Eq.~\eqref{eq:cost}, satisfies:
\vspace{-5pt}
\begin{equation}\label{eq:regret-upper-bound}
\begin{split}
\mathcal{R}(\pi)
&= \min\Biggl\{
\ \mathcal{O}\!\left(\frac{|\mathcal{S}|\,(M|\mathcal{S}|-|\mathcal{G}|)\,MD}{\Delta}\,\log T\right),\\
&\ \mathcal{O}\!\left(\sqrt{MDB\bigl(|\mathcal{S}|\,(M|\mathcal{S}|-|\mathcal{G}|)\bigr)\,T\log T}\right)
\Biggr\}.
\end{split}
\end{equation}
\end{restatable}
Eq.~\eqref{eq:regret-upper-bound} achieves a sublinear convergence rate to $\bm\pi^b$, contrasting the linear rate to WIP in \citet{wang2023optimistic}. This bound is independent of $\gamma$ and the smallest transition probability, with the constant formalized in Appendix~\ref{app:proof-thresholding}. Although Theorem~\ref{th:sublinearregret} sets $\eta = \Delta$, empirical results show algorithm stability across $\eta$ values in terms of average cumulative reward (Appendix~\ref{subsec:ablation}).
\textbf{Theorem~\ref{th:sublinearregret} Proof Sketch\;\;} 
Our analysis deviates from standard CMAB analysis because estimating the incremental reward requires pulling suboptimal arms; consequently, we must \emph{lower} bound the number of times suboptimal arms are pulled, in addition to the usual upper bound.
The argument proceeds in three steps. 1) Regret equivalence: we show that 
regret is equivalent to 
the expected number of 
interventions 
given to bad agents. 2) Bounding the interventions: we prove that this expected number can be bounded 
either by the overestimation of the incremental reward of bad agents or by the underestimation of that of good agents.
We further establish that at each time step, both the number of interventions and non-interventions assigned to agents are lower bounded. 3) Concentration: we apply concentration inequality (Lemma~\ref{le:concentration-empirical}) 
to obtain the desired result. 
Importantly, stationarity of the reward is not required here, since the concentration inequality we employ 
holds without such an assumption. 
\begin{remark}[How to Select the Threshold $\gamma$?]\label{remark:trade-off-gamma}
In our problem, there is an inherent trade-off between satisfying Assump.~\ref{asssumption:enough-good-arm} and maximizing cumulative reward. While our reformulation focuses on the threshold identification problem, maximizing cumulative reward remains our ultimate goal. Specifically, a smaller threshold \(\gamma\) makes it easier to satisfy Assump.~\ref{asssumption:enough-good-arm}, as more state-agent pairs qualify as good. However, a low \(\gamma\) also makes any arm above the threshold optimal, potentially leading to lower rewards. Empirically, we observe that there exists a universal choice of $\gamma$ which satisfies the Assump.~\ref{asssumption:enough-good-arm} that achieves the highest reward as shown in Appendix~\ref{paragraph:experiments-gamma}.
\end{remark}
  \begin{figure*}[!t]
       \centering
\includegraphics[width=\linewidth]{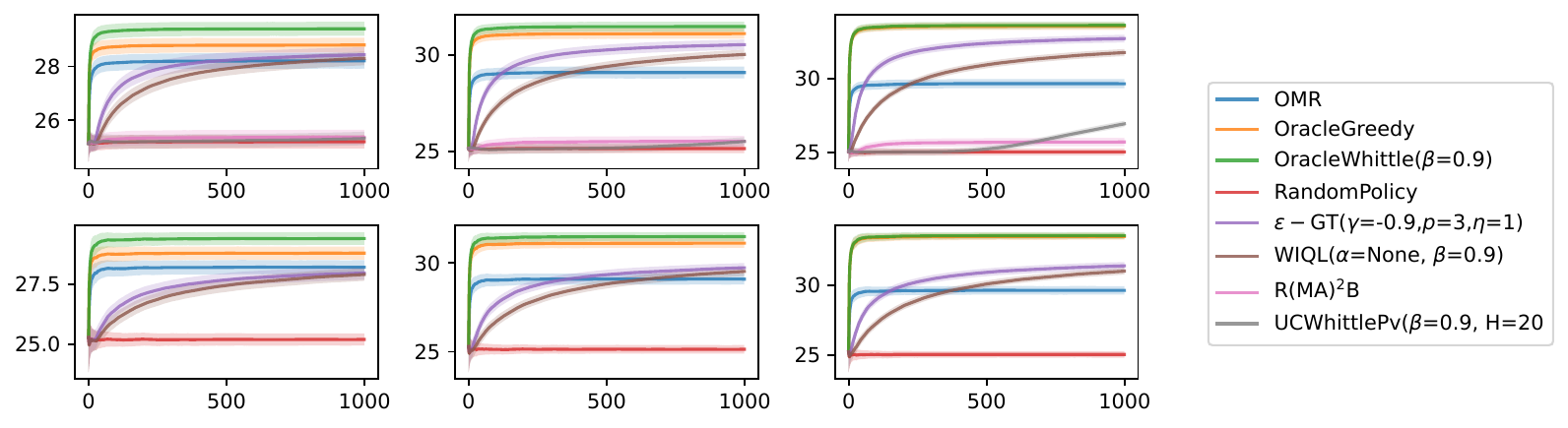}
        \caption{Average (over time) cumulative reward for budgets $B=5$ (left), $10$ (middle), $20$ (right), averaged over 50 instances with 13 repetitions each. Top: noiseless reward; bottom: noisy reward. Exclude R$(\text{MA})^2)$B and UCWhittlePV in the noisy setting since they require known reward.}
        \label{fig:sub5}
\end{figure*}
\section{Experiments}\label{sec:experiments}
We numerically evaluate the performance of our algorithm, showing: 1) sublinear regret under the 
BT-CMAB
reformulation (results in Appendix~\ref{subsec:implementation-detail-regret}), and 2) the 
achieved
average cumulative reward, 
as defined in the finite-horizon RB setting,
outperforms that of existing online RB
algorithms. 
These results underscore the practicality of the proposed reformulation and algorithms in finite-horizon restless bandit problems.
\textbf{Benchmarks\;\;} 
We compare the average cumulative reward obtained by our algorithm with the following benchmarks in online RB settings: WIP based: 1) 
WIQL \citep{biswas2021learn}, 2) UCWhittlePv \citep{wang2023optimistic}; LP-based: 3) R$(\text{MA})^2$B \citep{xiong2022reinforcement}; and 4) \emph{Random Policy} (uniformly pick $B$ agents), which follows the experimental setting of \citet{wang2023optimistic,xiong2022reinforcement}.\footnote{Our results differ from those of \citet{wang2023optimistic} due to detected bugs in their code; a detailed discussion is provided in Appendix~\ref{subsec:exp-detail-cumulative}.}

With heterogeneous agents,
no algorithm has been established as optimal. We incorporate three benchmark algorithms with full MDP knowledge:
5) \emph{Oracle Whittle} \citep{whittle1988restless}, 6) \emph{Oracle Greedy} and an LP-based policy 7) \emph{OMR} \citep{xiong2022reinforcement}. 
%
Baseline details and implementation are in Appendix~\ref{subsec:exp-detail-cumulative}.

\textbf{Setup\;\;} We set $M=50, \mathcal{S}=\{0,1\}$, and compare the results under three different budget settings: $B=5,10,20$. We use: 1) \textbf{Synthetic Instances}: The expected reward is set to be $R^m(s)=s$ and randomly generate 50 instances with different transition matrices, 2) \textbf{Wireless Communication Dataset \citep{https://doi.org/10.1002/sat.964}}: We use the same wireless communication dataset as \citet{wang2020restless}. Details will be deferred to Section~\ref{subsec:exp-real-data}.  We also consider the setting where $\mathcal{S}=[10]$ and randomly generate 50 instances. The results are deferred to Appendix~\ref{paragraph:experiments-10-states}.

In Figure~\ref{fig:sub5}, each line is averaged over $50$ instances, each with
$13$ repetitions.
We consider two reward settings: 1) a noiseless case, and 2) a noisy setting where the rewards at state $s\in\mathcal{S}$ follow a normal distribution with mean $R^m(s)$ and variance 1 for the synthetic dataset and 0.1 for the real-world dataset. We note that we did not fine-tune our hyperparameters. Specifically, we universally choose a threshold that satisfies Assump.~\ref{asssumption:enough-good-arm}. We have also shown that our algorithm is robust across different $\eta$ and we have done ablations on the choice of threshold (Appendix~\ref{paragraph:experiments-gamma}).
\subsection{Synthetic Instances}
\textbf{Results Under Noiseless Rewards\;\;} In Figure~\ref{fig:sub5} (top row), we observe that Alg.~\ref{alg:lcb-greedy} consistently outperforms online 
RB
algorithms within a horizon of $1000$  in all budget settings.
As shown in Figure~\ref{fig:sub5}, Alg.~\ref{alg:lcb-greedy} quickly converges to a stable yet suboptimal policy, whereas WIQL converges more slowly. However, given a sufficiently long horizon, WIQL will eventually converge to the \emph{Oracle Whittle} policy.
Nevertheless, Alg.~\ref{alg:lcb-greedy} outperforms WIQL when the horizon is short, even when the budget is low, highlighting the practicality of our algorithm.
UCWhittlePv requires extensive exploration to converge, and R$(\text{MA})^2$B performs similarly to random policy.

\textbf{Results Under Noisy Rewards} 
In Figure~\ref{fig:sub5} (bottom row), we excluded UCWhittlePv 
and R$(\text{MA})^2$B since they underperformed WIQL in the noiseless setting.
In all three budget settings, Alg.~\ref{alg:lcb-greedy} consistently outperforms WIQL. 

Lastly, we compare \emph{Oracle Whittle}, \emph{Oracle Greedy}, and \emph{OMR}. \emph{Oracle Whittle} consistently outperforms \emph{Oracle Greedy} across all budgets, with the gap narrowing as budget increases. Despite this, Alg.~\ref{alg:lcb-greedy} surpasses all online benchmarks, demonstrating the benefits of rapid convergence. \emph{OMR} performed worst, likely because its LP-based construction leverages homogeneity and falters in heterogeneous settings.
  \begin{figure*}[!t]
       \centering
\includegraphics[width=\linewidth]{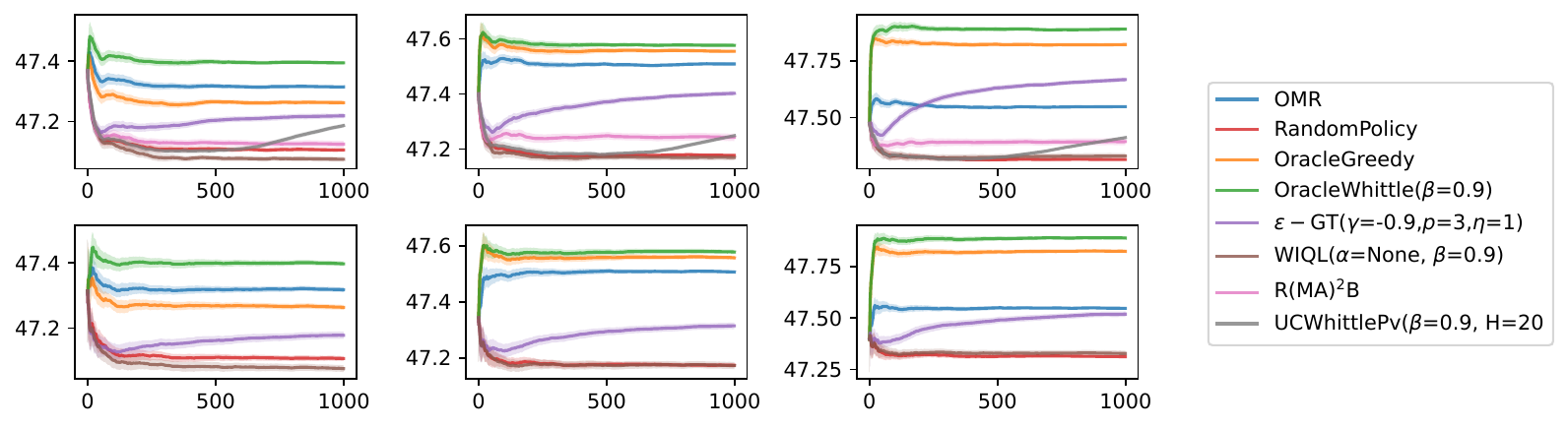}
        \caption{Average (over time) cumulative reward for budgets $B=5$ (left), $10$ (middle), $20$ (right), averaged over 100 repetitions of the real-world dataset. Top: noiseless reward; bottom: noisy reward. Exclude R$(\text{MA})^2)$B and UCWhittlePV in a noisy setting since they require known reward.}
        \label{fig:subreal}
\end{figure*}
\subsection{Wireless Communication Dataset}\label{subsec:exp-real-data}
This problem \citep{https://doi.org/10.1002/sat.964} involves digital video broadcasting satellite services to handheld devices via land mobile satellite. In this problem, the environment is modeled as a two-state system (“good’’ and “bad’’), with five different transition models corresponding to different satellite locations. In their original problem, each agent is provided with multiple actions which is modeled as the angle of the satellite. Similar to \citet{wang2020restless}, we choose two angles as the action: 40 degrees as action 0 and 60 degrees as action 1. The transition probability is chosen according to Table IV in \citet{https://doi.org/10.1002/sat.964}. The original reward is dependent on both action and state, but the reward difference between different actions are low (Table III in \citet{https://doi.org/10.1002/sat.964}). Further, the reward difference between the good state and bad state is also small. Therefore, we reconstruct the reward and set the reward for good state as 1 and bad state as 0.9 to mimic the real-world situations where signal is weak. We randomly assign 50 agents with one of the five transition probabilities. 

In Figure~\ref{fig:subreal}, we observe that in this problem, Our method consistently outperforms all benchmarks within a horizon of 1000 in all budget settings under noiseless and noisy rewards. As shown in Figure~\ref{fig:subreal},  Alg.~\ref{alg:lcb-greedy} converges to its oracle policy (\emph{OracleGreedy}) quickly while other online RB algorithms converge to their oracle policies more slowly. This translates to the superior performance of Alg.~\ref{alg:lcb-greedy}, demonstrating our observation that quick convergence translates to better finite sample performance.
\textbf{Discussions\;\;} 
We establish greedy policy optimality in finite-horizon RBs and reformulate online RB as a BT-CMAB with a learning algorithm. Our approach leverages the insight that fast convergence to a suboptimal policy can outperform slow convergence to a better one in finite horizons. 
As the BT-CMAB formulation is novel, our work appropriately focuses on establishing its theoretical and empirical value. Deriving a problem-specific lower bound is a natural and important direction for future work. Further, we plan to analyze the optimality of the oracle policy in multi-state setting.
\bibliography{ref}
\bibliographystyle{icml2025}
\section*{Checklist}
\begin{enumerate}

  \item For all models and algorithms presented, check if you include:
  \begin{enumerate}
    \item A clear description of the mathematical setting, assumptions, algorithm, and/or model. [Yes]
    \item An analysis of the properties and complexity (time, space, sample size) of any algorithm. [Yes]
    \item (Optional) Anonymized source code, with specification of all dependencies, including external libraries. [Yes]
  \end{enumerate}

  \item For any theoretical claim, check if you include:
  \begin{enumerate}
    \item Statements of the full set of assumptions of all theoretical results. [Yes]
    \item Complete proofs of all theoretical results. [Yes]
    \item Clear explanations of any assumptions. [Yes]     
  \end{enumerate}

  \item For all figures and tables that present empirical results, check if you include:
  \begin{enumerate}
    \item The code, data, and instructions needed to reproduce the main experimental results (either in the supplemental material or as a URL). [Yes]
    \item All the training details (e.g., data splits, hyperparameters, how they were chosen). [Yes]
    \item A clear definition of the specific measure or statistics and error bars (e.g., with respect to the random seed after running experiments multiple times). [Yes]
    \item A description of the computing infrastructure used. (e.g., type of GPUs, internal cluster, or cloud provider). [Yes]
  \end{enumerate}

  \item If you are using existing assets (e.g., code, data, models) or curating/releasing new assets, check if you include:
  \begin{enumerate}
    \item Citations of the creator If your work uses existing assets. [Not Applicable]
    \item The license information of the assets, if applicable. [Not Applicable]
    \item New assets either in the supplemental material or as a URL, if applicable. [Not Applicable]
    \item Information about consent from data providers/curators. [Not Applicable]
    \item Discussion of sensible content if applicable, e.g., personally identifiable information or offensive content. [Not Applicable]
  \end{enumerate}

  \item If you used crowdsourcing or conducted research with human subjects, check if you include:
  \begin{enumerate}
    \item The full text of instructions given to participants and screenshots. [Not Applicable]
    \item Descriptions of potential participant risks, with links to Institutional Review Board (IRB) approvals if applicable. [Not Applicable]
    \item The estimated hourly wage paid to participants and the total amount spent on participant compensation. [Not Applicable]
  \end{enumerate}
\end{enumerate}
\newpage
\appendix
\onecolumn
\aistatstitle{Appendix}
\renewcommand{\theequation}{\thesection.\arabic{equation}}
\renewcommand{\thetable}{\thesection.\arabic{table}}
\renewcommand{\thealgorithm}{\thesection.\arabic{algorithm}}
\renewcommand{\thefigure}{\thesection.\arabic{figure}}
\section{Real-World Restless Bandit Examples}
\begin{table*}[!th]
\centering
\renewcommand{\arraystretch}{1.2} 
\caption{Examples of good and bad states and corresponding budgeted interventions in various domains}
\label{tab:fixedwidth}
\vspace{5pt}
{\hyphenpenalty=10000\exhyphenpenalty=10000
\begin{tabular}{
K{0.105\textwidth} 
K{0.26\textwidth} 
K{0.25\textwidth} 
K{0.28\textwidth}
}
\hline
\textbf{Domain} & \textbf{Good state} & \textbf{Bad state} & \textbf{Budgeted Intervention Example} \\
\hline

General Usage & 
Frequent logins, feature exploration, meaningful use & 
Rare/brief logins, session abandonment & 
Reach out, offer help, gather feedback \\

E-commerce & 
High purchase frequency, browsing multiple categories & 
Low/no purchases, high cart abandonment & 
Provide a personalized discount or free shipping \\

E-learning & 
Completes modules, passes quizzes, revisits content & 
Seldom opens modules, fails quizzes & 
Progress‐based reward (finish by deadline for free certificate) \\

Loyalty & 
Completes surveys, redeems rewards, advances loyalty & 
Not collecting points, ignoring surveys, no rewards & 
Special bonus for feedback (earn points by reviewing new features) \\

Patient management & 
Updates records, completes follow‐ups, tracks adherence & 
Rarely logs updates, misses assessments or follow‐ups & 
Points or bonus for timely follow‐ups \\

\hline
\end{tabular}}
\label{tab:examples}
\end{table*}
We will further use an example to show case how the state, action, reward and budget is defined in practice. SwipeRX is an online platform in Indonesia where pharmacists typically place orders on Mondays. The platform aims to increase engagement and revenue by providing targeted coupons to pharmacists. Since pharmacists operate in different local markets with varying demand patterns and competition from other platforms, we model them as heterogeneous agents, each represented by an MDP.
\begin{itemize}
    \item \textbf{State}: The pharmacist's engagement level with the platform
\item \textbf{Action}: Binary decision of whether to provide a coupon to a given pharmacist
\item \textbf{Action-dependent State Transitions}: Based on whether an action is taken or not, the agent's engagement evolves accordingly based on environmental factors and platform interactions
\item \textbf{Reward}: Engagement metrics (e.g., time spent on platform, order frequency)
\item \textbf{Budget}: The platform has a budget on the number of coupons available at each day.
\end{itemize}

\section{Whittle's Index }\label{app:whittle}
Whittle's index is computed for each agent. 
At each time $t$, given $\bm s_t$, the top $B$ agents with the highest indices are selected.
Whittle's index policy has two variants, each aimed at maximizing a different objective in an \emph{infinite} horizon setting: one for maximizing average reward and the other for maximizing discounted reward.

\textbf{Average Reward:} In this case, Whittle's index policy is designed to approximately find the policy $\bm{\pi}_d$ that maximizes the average reward,
\begin{align}
    \max_{\bm{\pi}_d}~&\liminf_{T\to\infty}\frac{1}{TM}\E\left[\sum_{m=1}^M\sum_{t=0}^TR^m\left(s_t^m\right)\mid\bm{s}_0
    \right]\notag\\
    \text{s.t.~}&\sum_{m=1}^M a_t^m= B,\forall t\in[T]\label{eq:optimization-average-reward}.
\end{align}
We will denote $J^*$ to be the optimal value of Problem \ref{eq:optimization-average-reward}. Instead of directly solving Problem \ref{eq:optimization-average-reward}, Whittle deals with the Lagrangian relaxation of Problem \ref{eq:optimization-average-reward}, which is stated in Problem \ref{eq:optimization-average-reward-relaxation}
\begin{equation}
    \max_{\bm{\pi}_d}~\liminf_{T\to\infty}\frac{1}{TM}\E\left[\sum_{m=1}^M\sum_{t=0}^TR^m\left(s_t^m\right)+\lambda\left(1-a_{t+1}^m\right)\mid\bm{s}_0
    \right].\label{eq:optimization-average-reward-relaxation}
\end{equation}
The key observation to derive Whittle's index policy is that Problem \ref{eq:optimization-average-reward-relaxation} can be solved by addressing the following dynamic programming separately for each agent
\begin{align*}
    V^m_\lambda(s)&=\max_{a\in\{0,1\}} Q^m(s,a)\\
    Q^m_\lambda(s,a)+\beta^*_m&=R^m(s)-\lambda\mathbb{I}\{a=1\}\\&+\sum_{s'\in\mathcal{S}}P^m_a(s,s')V^m(s'),
\end{align*}
where $\beta^*_m$ is the maximized average reward for arm $m$ without any constraint, i.e.
\begin{equation*}
    \beta^*_m:=\max~ \liminf_{T\to\infty}\frac{1}{T}\E\left[\sum_{t=0}^TR^m\left(s_t^m\right)-\lambda\mathbb{I}\{a=1\}\mid s_0^m
    \right].
\end{equation*}
The Whittle's index $\lambda^m_s$ is defined as the multiplier $\lambda_s^m$ when it is equally favorable between choosing action $1$ and action $0$, i.e. $Q^m_{\lambda^m_s}(s,1)=Q^m_{\lambda^m_s}(s,0)$. The Whittle's index policy would then give action to $B$ agents with the highest Whittle's index at each time step $t$. Denote this as $\bm{\pi}^{\text{Whittle}}$ and the average reward of Whittle's index policy as $J^{\text{Whittle}}$. The performance of Whittle's index policy is characterized in Theorem \ref{th:optimality-whittle}.
\begin{theorem}[Theorem 2, \citealt{Weber_Weiss_1990}]\label{th:optimality-whittle}
Suppose all arms are homogeneous and assumptions therein, i.e., $P_a^m=P_a^{m'}, R^m=R^{m'},\forall m\neq m'$, then the Whittle's index policy is asymptotically optimal,
\begin{equation*}
    \lim_{M\to\infty}J^{\text{Whittle}}=\lim_{M\to\infty}J^*.
\end{equation*}
\end{theorem}

\textbf{Discounted Reward:} In this case, Whittle's index policy is designed to approximately find the policy $\bm{\pi}_d$ that maximizes the discounted reward for a fixed discount factor $\beta$,
\begin{align}
    \max_{\bm{\pi}_d}~&\sum_{t=0}^{\infty}\frac{1}{M}\E\left[\sum_{m=1}^M\beta^tR^m\left(s_t^m\right)\mid\bm{s}_0
    \right]\notag\\
    \text{s.t.~}&\sum_{m=1}^M a_t^m= B,\forall t\in[T]\label{eq:optimization-discounted-reward}.
\end{align}
Similarly, instead of directly solving Problem \ref{eq:optimization-discounted-reward}, Whittle deals with the Lagrangian relaxation of Problem \ref{eq:optimization-discounted-reward}, which is stated in Problem \ref{eq:optimization-discounted-reward-relaxation}
\begin{equation}
    \max_{\bm{\pi}_d}~\sum_{t=0}^{\infty}\frac{1}{M}\E\left[\sum_{m=1}^M\beta^t\left(R^m\left(s_t^m\right)+\lambda\left(1-a_{t+1}^m\right)\right)\mid\bm{s}_0
    \right].\label{eq:optimization-discounted-reward-relaxation}
\end{equation}
Again, the same key observation is that Problem \ref{eq:optimization-discounted-reward-relaxation} can be solved by addressing the following dynamic programming separately for each agent
\begin{align}
    V^m_\lambda(s)&=\max_{a\in\{0,1\}} Q^m(s,a)\notag\\
    Q^m_\lambda(s,a)&=R^m(s)-\lambda\mathbb{I}\{a=1\}\notag\\&+\beta\sum_{s'\in\mathcal{S}}P^m_a(s,s')V^m(s').\label{eq:dynamic-programming-discounted}
\end{align}
The definition of Whittle's index will be the same in the average reward case.
\subsection{Generalized Regret in Online Restless Bandits}\label{app:generalized_regret}
Given $K$ episodes each with a horizon length of $T$ and the set of online Whittle's index policy at each episode $k\in[K]$, $\bm\omega:=\{\bm \omega_k\}_{k\in[K]}$, the regret $\mathcal{R}_{\text{Restless}}^{\bm\pi^{\text{Whittle}}}(\bm{\omega})$:
\begin{align}\label{eq:regret-finite-horizon-general}
\mathcal{R}_{f}(\bm{\omega}):&=\frac{1}{K}\sum_{k=1}^K\Bigg(\E_{\bm{\pi}^{\text{Whittle}}}\left[\sum_{m=1}^M\sum_{t=0}^TR^m\left(s_t^m\right)\mid\bm{s}_0
    \right]-\E_{\bm{\omega}_k}\left[\sum_{m=1}^M\sum_{t=0}^TR^m\left(s_t^m\right)\mid\bm{s}_0
    \right]\Bigg).
\end{align}
We note that our problem setting corresponds to the scenario where $K=1$.

Existing regret bounds on online Whittle's index policy are only meaningful when the number of episodes is sufficiently large. 
To illustrate this, \citet{wang2023optimistic} shows that their algorithm has a regret bound of the order $\mathcal{O}\left(T\sqrt{K}\right)$. If we only have $1$ episode, this will be equivalent to saying that the regret bound is $\mathcal{O}(T)$. The hardness lies in estimating the MDP:  \citealt{auer2008near} show that the lower bound of the regret is $\Omega(T\sqrt{K})$. This lower bound indicates that when directly dealing with an MDP, it is impossible to get a regret bound that is sublinear in $T$.

\section{Counter Example for Heterogeneous Agents}\label{sec:counter-example-heterogeneous}
In this section, we will construct an instance involving heterogeneous agents where the greedy policy is not optimal. Say that each agent is associated with two state: State 0 and State 1, and the agent can
take either Action $1$ (intervention) or Action $0$ (no intervention). Assume for now that there are \textbf{4} agents with different transition probabilities. 
The first two agents fall under Type 1:
their transition probability under Action 1 is shown in Figure \ref{fig:transition-1-1} 
and their transition probability under Action 0
is shown in Figure \ref{fig:transition-1-0}. 
The other two agents fall under Type 2:
their transition probability under Action 1
is shown in Figure \ref{fig:transition-2-1} and their transition probability under Action 0 
is shown in Figure \ref{fig:transition-2-0}. 
\begin{figure}[!h]
\centering
\begin{tikzpicture}
 \node[circle, draw] (0) at (0,0) {$0$};
  \node[circle, draw] (1) at (4,0) {$1$};
  
  \draw[->] (0) edge[out=150, in=200, loop] node[left] {$p=\frac{1}{3}$}  (1);
  \draw[->] (1) edge[out=45, in=355, loop] node[right] {$p=\frac{1}{2}$}  (0);
  \draw[->] (0) to[bend left]  node[above] {$p=\frac{2}{3}$}(1);
  \draw[->] (1) to[bend left] node[below] {$p=\frac{1}{2}$}(0);
   \node at (0,1) {$r=0$};
  \node at (4,1) {$r=1$};
\end{tikzpicture}
\caption{Dynamic of Agent Type 1 with Intervention}
\label{fig:transition-1-1}
\end{figure}
\begin{figure}[!h]
\centering
\begin{tikzpicture}
 \node[circle, draw] (0) at (0,0) {$0$};
  \node[circle, draw] (1) at (4,0) {$1$};
  
  \draw[->] (0) edge[out=150, in=200, loop] node[left] {$p=1$}  (1);
  \draw[->] (1) edge[out=45, in=355, loop] node[right] {$p=\frac{1}{3}$}  (0);
  \draw[->] (0) to[bend left]  node[above] {$p=0$}(1);
  \draw[->] (1) to[bend left] node[below] {$p=\frac{2}{3}$}(0);
   \node at (0,1) {$r=0$};
  \node at (4,1) {$r=1$};
\end{tikzpicture}
\caption{Dynamic of Agent Type 1 without Intervention}
\label{fig:transition-1-0}
\end{figure}
\begin{figure}[!h]
\centering
\begin{tikzpicture}
 \node[circle, draw] (0) at (0,0) {$0$};
  \node[circle, draw] (1) at (4,0) {$1$};
  
  \draw[->] (0) edge[out=150, in=200, loop] node[left] {$p=\frac{1}{4}$}  (1);
  \draw[->] (1) edge[out=45, in=355, loop] node[right] {$p=3/4$}  (0);
  \draw[->] (0) to[bend left]  node[above] {$p=\frac{3}{4}$}(1);
  \draw[->] (1) to[bend left] node[below] {$p=1/4$}(0);
   \node at (0,1) {$r=0$};
  \node at (4,1) {$r=1$};
\end{tikzpicture}
\caption{Dynamic of Agent Type 2 with Intervention}
\label{fig:transition-2-1}
\end{figure}
\begin{figure}[!ht]
\centering
\begin{tikzpicture}
 \node[circle, draw] (0) at (0,0) {$0$};
  \node[circle, draw] (1) at (4,0) {$1$};
  
  \draw[->] (0) edge[out=150, in=200, loop] node[left] {$p=1$}  (1);
  \draw[->] (1) edge[out=45, in=355, loop] node[right] {$p=0$}  (0);
  \draw[->] (0) to[bend left]  node[above] {$p=0$}(1);
  \draw[->] (1) to[bend left] node[below] {$p=1$}(0);
   \node at (0,1) {$r=0$};
  \node at (4,1) {$r=1$};
\end{tikzpicture}
\caption{Dynamic of Agent Type 2 without Intervention}
\label{fig:transition-2-0}
\end{figure}

Suppose the budget equals one, meaning that we can only give one agent intervention at each time. 
We observe that under our problem setup Type 2 agents always have a higher incremental reward $\phi^2(1,1):=\frac{3}{4}\cdot 1, \phi^2(0,1):=\frac{3}{4}\cdot 1,\phi^2(s,0)=0,I^2(1)=I^2(0)=\frac{3}{4}$, opposed to $\phi^1(0,1)=\frac{2}{3}\cdot 1,\phi^1(1,1)=\frac{1}{2}\cdot 1,\phi^1(0,0)=0,\phi^1(1,0)=\frac{1}{3}, I(1)=\frac{2}{3},I(0)=\frac{1}{6}$. 
However, this does not imply that the optimal policy is to always take Action 1 for Type 2 agents. If all resources are allocated to Type 2 agents, the expected reward is $3/4$ per time step summed across all agents. Alternatively, if resources are devoted to Type 1 agents, alternating actions between the two Type 1 agents, the expected reward is $2/3 + 2/3 \times 1/3 = 8/9$ per Type 1 agent over two time periods (where an action is taken in one period but not in the other). To illustrate this point, we ran the following experiment:
we set the horizon to be $20$ and repeated the simulation $300$ times. We compare the performance of the following two policies in terms of cumulative rewards: Policy 1) choosing one of the Type 2 agents for intervention at each time, and Policy 2) alternating intervention between the two Type 1 agents. The observed cumulative for Policies 1 and 2 are 15.02 and 16.78, respectively. This shows that the greedy policy is not optimal under all heterogeneous agents cases.
\section{Counter Example for Three-State, Homogeneous Agents}\label{sec:counter-three-state}
In this section, we will construct an instance involving homogeneous agents with three states where the greedy policy is not optimal. Say that each agent is associated with three states: State 0, State 1 and State 2, and the agent can take either Action 1 (intervention) or Action 0 (no intervention). Assume for now that there are two agents with the same transition probabilities. Their transition probability under Action 1 is shown in Figure~\ref{fig:transition-1} and their transition probability under Action 0 is shown in Figure~\ref{fig:transition-2}.
\begin{figure}[!h]
\centering
\begin{tikzpicture}
 \node[circle, draw] (0) at (0,0) {$0$};
  \node[circle, draw] (1) at (4,0) {$1$};
  \node[circle, draw] (2) at (8,0) {$2$};
  \draw[->] (0) edge[loop below] node[below] {$p=1$}  (1);
  \draw[->] (1) edge[loop below] node[below] {$p=\frac{1}{2}$}  (0);
  \draw[->] (2) edge[loop below] node[below] {$p=\frac{1}{7}$} (2);
  \draw[->] (1) to[bend left] node[below] {$p=\frac{1}{2}$}(2);
    \draw[->] (2) to[bend left] node[below] {$p=\frac{1}{7}$} (1);
     \draw[->] (2) to[bend left] node[below] {$p=\frac{5}{7}$} (0);
   \node at (0,1) {$r=0$};
  \node at (4,1) {$r=1$};
  \node at (8,1) {$r=2$};
\end{tikzpicture}
\caption{Dynamic with Intervention}
\label{fig:transition-1}
\end{figure}
\begin{figure}[!h]
\centering
\begin{tikzpicture}
 \node[circle, draw] (0) at (0,0) {$0$};
  \node[circle, draw] (1) at (4,0) {$1$};
  \node[circle, draw] (2) at (8,0) {$2$};
  \draw[->] (0) edge[loop below] node[below] {$p=1$}  (1);
  \draw[->] (1) edge[loop below] node[below] {$p=\frac{3}{4}$}  (0);
  \draw[->] (1) to[bend left] node[below] {$p=\frac{1}{4}$}(2);
    \draw[->] (2) to[bend left] node[below] {$p=\frac{5}{28}$} (1);
     \draw[->] (2) to[bend left] node[below] {$p=\frac{23}{28}$} (0);
   \node at (0,1) {$r=0$};
  \node at (4,1) {$r=1$};
  \node at (8,1) {$r=2$};
\end{tikzpicture}
\caption{Dynamic without Intervention}
\label{fig:transition-2}
\end{figure}

Suppose that the budget equals one, we observe that State 2 and State 1 have the same incremental reward $I(2)=\frac{3}{7}-\frac{5}{28}=\frac{1}{4}=\frac{3}{2}-\frac{5}{4}=I(1)$. However, we will show that when $T=3$, greedy policy is not optimal. Specifically, we will show that when $t=0$ and $\bm s_0=(1,2)$, the optimal policy would be giving State 2 Action 1 instead of State 1 Action 1. This will show that greedy policy is not optimal as greedy policy could give State 1 Action 1 by definition. Suppose we are giving State 1 Action 1, then we have
\begin{align*}
    V((1,2))&=3+\frac{5}{56}V_1^*((1,2))+\frac{5}{56}V_1^*((1,1))+\frac{23}{28}V_1^*((1,0))\\
    &=3+\frac{5}{56}\frac{47+3*28}{28}+\frac{5}{56}\frac{19}{4}+\frac{23}{28}\frac{5}{2}.
\end{align*}
Similarly, if we are giving State 2 Action 1, then we have
\begin{align*}
    V((1,2))&=3+\frac{1}{28}V_1^*((2,2))+\frac{1}{7}V_1^*((1,2))+\frac{15}{28}V_1^*((1,0))+\frac{5}{28}V_1^*((2,0))+\frac{3}{28}V_1^*((1,1))\\
    &=3 + \frac{1}{28}\left(\frac{17}{28}+4\right)
  + \frac{1}{7}\left(3+\frac{47}{28}\right)
  + \frac{15}{28}\left(\frac{5}{2}\right)
  + \frac{5}{28}\left(\frac{3}{7}+2\right)
  + \frac{3}{28}\left(\frac{11}{4}+2\right).
\end{align*}
By calculation, we will have the results. This shows that the greedy policy is not optimal under three states setting.
\section{Technical Details In Section \ref{subsec:optimality}}\label{appendix:proof-of-reduction}
First, recall that
at each time $t$ of an offline restless bandit problem, 
the learner observes the set of states of $M$ agents, denoted by $\bm{s}_t:=\left(s_t^1,s_t^2,\cdots,s_t^M\right)$, and receives the corresponding reward $\bm{r}_t:=(r_t^1,\cdots, r_t^M)$. 
Based on the observed states, the learner determines the action $\bm{a}_{t+1}:=\left(a_{t+1}^1,a_{t+1}^2,\cdots,a_{t+1}^M\right)$,  according to a deterministic policy $\pi^d_{t+1}(\bm{s}):\mathcal{S}^M\rightarrow 
\{0,1\}^M$.
The learner is provided with a budget constraint, allowing at most $B$ arms to be pulled at each time step,
i.e., $\sum_{m\in[M]} a_{t+1}^m= B$.
After pulling up to $B$ arms, the learner
observes the next state $\bm{s}_{t+1}$, 
generated according to $s_{t+1}^m\sim P_{a_{t+1}^m}^m\left(s_t^m,s_{t+1}^m\right)$, and 
receives a feedback $\bm{r}_{t+1}$.

To ease notation, in Section \ref{appendix:proof-of-reduction} we will use $q^a_{ss'}(m)$ to denote $P_a^m(s,s')$.
Before presenting the proofs, we reformulate the online restless bandit problem as a single large MDP, where the state and action are defined as tuples comprising the composite states and actions of all agents. This reformulation is valid because the agents/arms in the restless bandit problem are independent. This MDP can be described as
$\left(\mathcal{S}^M,\mathcal{F},q,\bm{R}\right)$, where $\mathcal{F}$ is defined as
\begin{align*}
    \mathcal{F}:=\bigg\{&\left(a^1,a^2,\cdots,a^M\right):\left(a^1,a^2,\cdots,a^M\right)\in\mathcal{A}^M, \sum_{m=1}^M a^m\leq B\bigg\}.
\end{align*}
Let $q_{\bm s \bm s'}^{\bm a}:=\mathbb{P}(\bm s'|\bm s, \bm a) = P_{\bm a}(\bm s, \bm s')= \prod_{m=1}^M 
q^{a}_{ss'}(m)$, and let
%
$\bm{R}(\bm{s})=\sum_{m=1}^M r^m\left(s^m\right)$.
Then for each policy $\bm{\pi}$ and each time-step $h$, we can define the value function $V_h^{\bm{\pi}}(\bm{s})$ as
\begin{equation}\label{eq:value-function}
    \E_{\bm{\pi}}\left[\sum_{t=h}^T \bm{R}(\bm{s}_t)\mid \bm{s}_h=\bm{s}\right].
\end{equation}
With this definition, Problem \eqref{eq:optimization-problem-online} is equivalent to
\begin{equation*}
    \max_{\bm{\pi}} V_0^{\bm{\pi}}\left(\bm{s}_0\right).
\end{equation*}
Again, we use $\bm \pi^*:= \{\pi_{t}^*\}_{t\in[T]}$ to denote the maximizer, and let $V_h^*$ be the value function under policy $\bm \pi^*$. We will use this notation in Appendix~ \ref{app:thm-optimality-greedy-homogeneous}, and \ref{app:proof-thresholding}.
\subsection{Proof of Theorem \ref{th:optimality-greedy-homogeneous}}\label{app:thm-optimality-greedy-homogeneous}
\optimalityhomogeneous*
\begin{proof}
WLOG, assume that $I(1)\leq I(0)$. For simplicity, we first look at the simple case where $M=2,B=1$. The generalization to multi-agent case is straightforward and will be deferred to the last part of the proof. First of all, we will derive some useful equations that are useful in the proof.
\begin{align}
    &\sum_{s_{h+1}^1,s_{h+1}^2}q^0_{1s_{h+1}^1}q^1_{0s_{h+1}^2}V_{h+1}^*\left(\left(s_{h+1}^1,s_{h+1}^2\right)\right)-  \sum_{s_{h+1}^1,s_{h+1}^2}q^1_{1s_{h+1}^1}q^0_{0s_{h+1}^2}V_{h+1}^*\left(\left(s_{h+1}^1,s_{h+1}^2\right)\right)\notag\\
    &=q^0_{11}q^1_{01}V_{h+1}^*\left((1,1)\right)+q^0_{11}q^1_{00}V_{h+1}^*\left((1,0)\right)+q^0_{10}q^1_{01}V_{h+1}^*\left((1,0)\right)+q^0_{10}q^1_{00}V_{h+1}^*\left((0,0)\right)\notag\\
    &-q^1_{11}q^0_{01}V_{h+1}^*\left((1,1)\right)-q^1_{11}q^0_{00}V_{h+1}^*\left((1,0)\right)-q^1_{10}q^0_{01}V_{h+1}^*\left((1,0)\right)-q^1_{10}q^0_{00}V_{h+1}^*\left((0,0)\right)\notag\\
    &=q^0_{11}q^1_{01}V_{h+1}^*\left((1,1)\right)-q^0_{11}q^1_{01}V_{h+1}^*\left((1,0)\right)+q^0_{11}V_{h+1}^*\left((1,0)\right)\notag\\
    &+q^1_{01}V_{h+1}^*\left((1,0)\right)-q^0_{11}q^1_{01}V_{h+1}^*\left((1,0)\right)\notag\\
    &+V_{h+1}^*\left((0,0)\right)+q^0_{11}q^1_{01}V_{h+1}^*\left((0,0)\right)-\left(q^0_{11}+q^1_{01}\right)V_{h+1}^*\left((0,0)\right)\notag\\
     &-q^1_{11}q^0_{01}V_{h+1}^*\left((1,1)\right)+q^1_{11}q^0_{01}V_{h+1}^*\left((1,0)\right)-q^1_{11}V_{h+1}^*\left((1,0)\right)\notag\\
    &-q^0_{01}V_{h+1}^*\left((1,0)\right)+q^1_{11}q^0_{01}V_{h+1}^*\left((1,0)\right)\notag\\
    &-V_{h+1}^*\left((0,0)\right)-q^1_{11}q^0_{01}V_{h+1}^*\left((0,0)\right)+\left(q^1_{11}+q^0_{01}\right)V_{h+1}^*\left((0,0)\right)\notag\\
    &= \left(q^0_{11}+q^1_{01}-q^1_{11}-q^0_{01}\right)\left(V_{h+1}^*\left((1,0)\right)-V_{h+1}^*\left((0,0)\right)\right)\notag\\&+\left(q^0_{11}q^1_{01}-q^0_{01}q^1_{11}\right)\left(V_{h+1}^*\left((1,1)\right)-2V_{h+1}^*\left((1,0)\right)+V_{h+1}^*\left((0,0)\right)\right).\label{eq:simplification-optimality-equation}
\end{align}

Next, we will simplify the difference between the value function of different states, assuming that the optimal policy is the greedy policy. For simplicity, we will only give the detailed derivation of the equation for $V_{h}^*\left((1,1)\right)-V_h^*\left((1,0)\right)$. The equation for $V_{h}^*\left((1,0)\right)-V_h^*\left((0,0)\right)$ holds for similar reasons.
\begin{align}
    V_{h}^*\left((1,1)\right)-V_h^*\left((1,0)\right)&=2R(1)+\sum_{s_{h+1}^1,s_{h+1}^2}q^1_{1s_{h+1}^1}q\left(s_{h+1}^2\mid 1,0\right)V_{h+1}^*\left(\left(s_{h+1}^1,s_{h+1}^2\right)\right)\notag\\
    &-R(1)-R(0)-\sum_{s_{h+1}^1,s_{h+1}^2}q^0_{1s_{h+1}^1}q^1_{0s_{h+1}^2}V_{h+1}^*\left(\left(s_{h+1}^1,s_{h+1}^2\right)\right)\notag\\
    &=R(1)-R(0)+q^1_{11}q^0_{11}V_{h+1}^*\left((1,1)\right)+q^1_{11}q^0_{10}V_{h+1}^*\left((1,0)\right)\notag\\&+q^1_{10}q^0_{11}V_{h+1}^*\left((1,0)\right)+q^0_{10}q^1_{10}V_{h+1}^*\left((0,0)\right)\notag\\
    &-q^0_{11}q^1_{01}V_{h+1}^*\left((1,1)\right)-q^0_{11}q^1_{00}V_{h+1}^*\left((1,0)\right)\notag\\&-q^0_{10}q^1_{01}V_{h+1}^*\left((1,0)\right)-q^0_{10}q^1_{00}V_{h+1}^*\left((0,0)\right)\notag\\
    &=R(1)-R(0)+q^1_{11}q^0_{11}V_{h+1}^*\left((1,1)\right)+q^1_{11}V_{h+1}^*\left((1,0)\right)-q^1_{11}q^0_{11}V_{h+1}^*\left((1,0)\right)\notag\\
    &+q^0_{11}V_{h+1}^*\left((1,0)\right)-q^1_{11}q^0_{11}V_{h+1}^*\left((1,0)\right)\notag\\&+V_{h+1}^*\left((0,0)\right)+q^0_{11}q^1_{11}V_{h+1}^*\left((0,0)\right)-\left(q^1_{11}+q^0_{11}\right)V_{h+1}^*\left((0,0)\right)\notag\\
    &-q^0_{11}q^1_{01}V_{h+1}^*\left((1,1)\right)-q^0_{11}V_{h+1}^*\left((1,0)\right)+q^1_{01}q^0_{11}V_{h+1}^*\left((1,0)\right)\notag\\
    &-q^1_{01}V_{h+1}^*\left((1,0)\right)+q^0_{11}q^1_{01}V_{h+1}^*\left((1,0)\right)\notag\\&-V_{h+1}^*\left((0,0)\right)-q^0_{11}q^1_{01}V_{h+1}^*\left((0,0)\right)+\left(q^0_{11}+q^1_{01}\right)V_{h+1}^*\left((0,0)\right)\notag\\
    &=R(1)-R(0)+\left(q^1_{11}-q^1_{01}\right)\left(V_{h+1}^*\left((1,0)\right)-V_{h+1}^*\left((0,0)\right)\right)\notag\\
    &+q^0_{11}\left(q^1_{11}-q^1_{01}\right)\left(V_{h+1}^*\left((1,1)\right)-2V_{h+1}^*\left((1,0)\right)+V_{h+1}^*\left((0,0)\right)\right).\label{eq:simplification-difference-1110}
\end{align}
Similarly, we can have
\begin{align}
    V_{h}^*\left((1,0)\right)-V_h^*\left((0,0)\right)&=R(1)-R(0)+\left(q^0_{11}-q^0_{01}\right)\left(V_{h+1}^*\left((1,0)\right)-V_{h+1}^*\left((0,0)\right)\right)\notag\\
    &+q^1_{01}\left(q^0_{11}-q^0_{01}\right)\left(V_{h+1}^*\left((1,1)\right)-2V_{h+1}^*\left((1,0)\right)+V_{h+1}^*\left((0,0)\right)\right)\label{eq:simplification-difference-1000}.
\end{align}

We will derive the final useful equation for the incremental reward.
\begin{align}
    I(1)&=\phi(1,1)-\phi(1,0)\notag\\
    &=q^1_{11}R(1)+q^1_{10}R(0)-q^0_{11}R(1)-q^0_{10}R(0)\\
    &=\left(q^1_{11}-q^0_{11}\right)\left(R(1)-R(0)\right).\label{eq:simplification-incremental-reward-1}
\end{align}
Similarly, we also have
\begin{equation}
    I(0)=\left(q^1_{01}-q^0_{01}\right)\left(R(1)-R(0)\right).\label{eq:simplification-incremental-reward-0}
\end{equation}

In the following, we will consider four cases. 
\paragraph{Case 1:}$q^0_{11}\geq q^0_{01},q^1_{11}\leq q^1_{01},q_{11}^1\geq q_{11}^0$.
We will use backward induction to prove the following statements:
\begin{align}
    \forall t\leq T, &0\leq V_t^*\left((1,0)\right)-V_t^*((0,0))\leq \frac{R(1)-R(0)}{q^0_{10}}\label{eq:monotonicity-10-case1}\\
    &0\leq V_t^*\left((1,1)\right)-V_t^*((1,0))\leq R(1)-R(0)\label{eq:monotonicity-11-case1}\\
    &\pi_{t+1}^{\text{greedy}}=\pi_{t+1}^*.\label{eq:optimality-lemma3-case1}
\end{align}
The statements hold trivially when $t=T$. Suppose the statements hold when $t\geq h+1$, when $t=h$, we will first prove Eq.~\eqref{eq:optimality-lemma3-case1} for $t=h$. Recall that the policy satisfying the following Bellman optimality equation is the optimal policy:
    \begin{equation}\label{eq:bellman-optimality}
    V_h^*\left((s_h^1,s_h^2)\right)=R(s_h^1)+R(s_h^2)+\max_{a_{h+1}^1,a_{h+1}^2}\sum_{s_{h+1}^1,s_{h+1}^2}q^{a_{h+1}^1}_{s_h^1s_{h+1}^1}q^{a_{h+1}^2}_{s_h^2s_{h+1}^2}V_{h+1}^*\left(s_{h+1}^1,s_{h+1}^2\right).
    \end{equation}
By the problem formulation, one of the $a_{h+1}^i=1$. Therefore, it's easy to see that when $s_h^1=s_h^2$, randomly giving action will satisfy this equation. We will focus on the case where $s_h^1\neq s_h^2$. In that case, to show Eq.~\eqref{eq:optimality-lemma3-case1}, we only need to show that
      \begin{equation}
    \sum_{s_{h+1}^1,s_{h+1}^2}q^0_{1s_{h+1}^1}q^1_{0s_{h+1}^2}V_{h+1}^*\left(s_{h+1}^1,s_{h+1}^2\right)\geq   \sum_{s_{h+1}^1,s_{h+1}^2}q^1_{1s_{h+1}^1}q^0_{0s_{h+1}^2}V_{h+1}^*\left(s_{h+1}^1,s_{h+1}^2\right).\label{eq:equation-showing-optimality}
    \end{equation}
By Eq.~\eqref{eq:simplification-optimality-equation}, this is equivalent to showing
\begin{align*}
    &\left(q^0_{11}+q^1_{01}-q^1_{11}-q^0_{01}\right)\left(V_{h+1}^*\left((1,0)\right)-V_{h+1}^*\left((0,0)\right)\right)\\&+\left(q^0_{11}q^1_{01}-q^0_{01}q^1_{11}\right)\left(V_{h+1}^*\left((1,1)\right)-2V_{h+1}^*\left((1,0)\right)+V_{h+1}^*\left((0,0)\right)\right)\geq 0.
\end{align*}
Then we have
\begin{align*}
    &\left(q^0_{11}+q^1_{01}-q^1_{11}-q^0_{01}\right)\left(V_{h+1}^*\left((1,0)\right)-V_{h+1}^*\left((0,0)\right)\right)\\&+\left(q^0_{11}q^1_{01}-q^0_{01}q^1_{11}\right)\left(V_{h+1}^*\left((1,1)\right)-2V_{h+1}^*\left((1,0)\right)+V_{h+1}^*\left((0,0)\right)\right)\\
    &\geq \left(q^0_{11}+q^1_{01}-q^1_{11}-q^0_{01}\right)\left(V_{h+1}^*\left((1,0)\right)-V_{h+1}^*\left((0,0)\right)\right)\\
    &-\left(q^0_{11}q^1_{01}-q^0_{01}q^1_{11}\right)\left(V_{h+1}^*\left((1,0)\right)-V_{h+1}^*\left((0,0)\right)\right)\\
    &=\frac{I(0)-I(1)}{R(1)-R(0)}\left(V_{h+1}^*\left((1,0)\right)-V_{h+1}^*\left((0,0)\right)\right)-\frac{\left(q^1_{11}I(0)-q^1_{01}I(1)\right)}{R(1)-R(0)}\left(V_{h+1}^*\left((1,0)\right)-V_{h+1}^*\left((0,0)\right)\right)\\
    &=\frac{\left(q^1_{10}I(0)-q^1_{00}I(1)\right)}{R(1)-R(0)}\left(V_{h+1}^*\left((1,0)\right)-V_{h+1}^*\left((0,0)\right)\right)\\
    &\geq 0,
\end{align*}
where the first inequality holds because of the induction hypothesis and $q^0_{11}\geq q^0_{01},q^1_{01}\geq q^1_{11}$, the first equality holds because of Eq.~\eqref{eq:simplification-incremental-reward-0} and Eq.~\eqref{eq:simplification-incremental-reward-1} and the last inequality holds because $q^1_{10}\geq q^1_{00}$, $I(0)\geq I(1)$ and the assumption that Eq.~\eqref{eq:monotonicity-10-case1} holds for $t=h+1$. 

Next, we first show the LHS of Eq.~\eqref{eq:monotonicity-10-case1} for $t=h$ under the assumption that Eq.~\eqref{eq:optimality-lemma3-case1} holds for $t=h$ and Eq.~\eqref{eq:monotonicity-10-case1} holds for $t=h+1$.
\begin{align*}
    V_{h}^*\left((1,0)\right)-V_h^*\left((0,0)\right)&=R(1)-R(0)+\left(q^0_{11}-q^0_{01}\right)\left(V_{h+1}^*\left((1,0)\right)-V_{h+1}^*\left((0,0)\right)\right)\\
    &+q^1_{01}\left(q^0_{11}-q^0_{01}\right)\left(V_{h+1}^*\left((1,1)\right)-2V_{h+1}^*\left((1,0)\right)+V_{h+1}^*\left((0,0)\right)\right)\\
    &\geq R(1)-R(0)+\left(q^0_{11}-q^0_{01}\right)\left(V_{h+1}^*\left((1,0)\right)-V_{h+1}^*\left((0,0)\right)\right)\\
    &-q^1_{01}\left(q^0_{11}-q^0_{01}\right)\left(V_{h+1}^*\left((1,0)\right)-V_{h+1}^*\left((0,0)\right)\right)\\
    &\geq R(1)-R(0)+q^1_{00}\left(q^0_{11}-q^0_{01}\right)\left(V_{h+1}^*\left((1,0)\right)-V_{h+1}^*\left((0,0)\right)\right)\\
    &\geq 0,
\end{align*}
where the first equality holds because of Eq.~\eqref{eq:simplification-difference-1000} and the last inequality holds since $q^0_{11}\geq q^0_{01}$. Under the same condition, for the RHS of Eq.~\eqref{eq:monotonicity-10-case1}, we have
\begin{align*}
    q^0_{10}\left(V_{h}^*\left((1,0)\right)-V_h^*\left((0,0)\right)\right)&=q^0_{10}(R(1)-R(0))\\
    &+q^0_{10}q^1_{00}\left(q^0_{11}-q^0_{01}\right)\left(V_{h+1}^*\left((1,0)\right)-V_{h+1}^*\left((0,0)\right)\right)\\
    &+q^0_{10}q^1_{01}\left(q^0_{11}-q^0_{01}\right)\left(V_{h+1}^*\left((1,1)\right)-V_{h+1}^*\left((1,0)\right)\right)\\
    &\leq q^0_{10}(R(1)-R(0))\\
    &+q^1_{00}\left(q^0_{11}-q^0_{01}\right)(R(1)-R(0))\\
    &+q^0_{10}q^1_{01}\left(q^0_{11}-q^0_{01}\right)(R(1)-R(0))\\
    &\leq   q^0_{10}(R(1)-R(0))\\&+q^1_{00}q^0_{11}(R(1)-R(0))+q^1_{01}q^0_{11}(R(1)-R(0))\\
    &=R(1)-R(0).
\end{align*}

Then, we only need to show Eq.~\eqref{eq:monotonicity-11-case1} for $t=h$. Assuming that Eq.~\eqref{eq:optimality-lemma3-case1} holds for $t=h$ and Eq.~\eqref{eq:monotonicity-11-case1} holds for $t=h+1$, for the RHS, we have
\begin{align*}
    V_{h}^*\left((1,1)\right)-V_h^*\left((1,0)\right)&=R(1)-R(0)+\left(q^1_{11}-q^1_{01}\right)\left(V_{h+1}^*\left((1,0)\right)-V_{h+1}^*\left((0,0)\right)\right)\\
    &+q^0_{11}\left(q^1_{11}-q^1_{01}\right)\left(V_{h+1}^*\left((1,1)\right)-2V_{h+1}^*\left((1,0)\right)+V_{h+1}^*\left((0,0)\right)\right)\\
    &=R(1)-R(0)+q^0_{11}\left(q^1_{11}-q^1_{01}\right)\left(V_{h+1}^*\left((1,1)\right)-V_{h+1}^*\left((1,0)\right)\right)\\&+q^0_{10}\left(q^1_{11}-q^1_{01}\right)\left(V_{h+1}^*\left((1,0)\right)-V_{h+1}^*\left((0,0)\right)\right)\\
    &\leq R(1)-R(0),
\end{align*}
where the last inequality holds since $q^1_{11}\leq q^1_{01}$. 
Under the same condition, for the LHS, we have
\begin{align*}
    V_{h}^*\left((1,1)\right)-V_h^*\left((1,0)\right)&=R(1)-R(0)+\left(q^1_{11}-q^1_{01}\right)\left(V_{h+1}^*\left((1,0)\right)-V_{h+1}^*\left((0,0)\right)\right)\\
    &+q^0_{11}\left(q^1_{11}-q^1_{01}\right)\left(V_{h+1}^*\left((1,1)\right)-2V_{h+1}^*\left((1,0)\right)+V_{h+1}^*\left((0,0)\right)\right)\\
    &= R(1)-R(0)+q^0_{11}\left(q^1_{11}-q^1_{01}\right)\left(V_{h+1}^*\left((1,1)\right)-V_{h+1}^*\left((1,0)\right)\right)\\&+q^0_{10}\left(q^1_{11}-q^1_{01}\right)\left(V_{h+1}^*\left((1,0)\right)-V_{h+1}^*\left((0,0)\right)\right)\\
    &\geq R(1)-R(0)+q^0_{11}\left(q^1_{11}-q^1_{01}\right)(R(1)-R(0))\\
    &+\left(q^1_{11}-q^1_{01}\right)(R(1)-R(0))\\
    &=\left(q^1_{00}+q^1_{11}-q^0_{11}q^1_{01}+q^0_{11}q^1_{11}\right)(R(1)-R(0))\\
    &\geq 0,
\end{align*}
where the last inequality holds since $q^1_{11}\geq q^0_{11}$.

\paragraph{Case 2:} $q^0_{11}\leq q^0_{01},q^1_{11}\leq q^1_{01}$. 

First we consider the sub-case where $\boxed{\bm{q^1_{11}q^0_{01}\leq q^1_{01}q^0_{11}}}$. We will use backward induction to prove the following statements:
\begin{align}
    \forall t\leq T, &0\leq V_t^*\left((1,0)\right)-V_t^*((0,0))\leq R(1)-R(0)\label{eq:monotonicity-10-case2-sub1}\\
    &0\leq V_t^*\left((1,1)\right)-V_t^*((1,0))\leq R(1)-R(0)\label{eq:monotonicity-11-case2-sub1}\\
    &\pi_{t+1}^{\text{greedy}}=\pi_{t+1}^*.\label{eq:optimality-lemma3-case2-sub1}
\end{align}
When $t=T$, the statements hold trivially. Suppose the statements hold for $t\geq h+1$, when $t=h$, we will first prove Eq.~\eqref{eq:optimality-lemma3-case2-sub1} under the assumption that Eq.~\eqref{eq:monotonicity-10-case2-sub1} and Eq.~\eqref{eq:monotonicity-11-case2-sub1} holds for $t=h+1$. Similarly we only need to show Eq.~\eqref{eq:equation-showing-optimality}. 
\begin{align*}
    &\left(q^0_{11}+q^1_{01}-q^1_{11}-q^0_{01}\right)\left(V_{h+1}^*\left((1,0)\right)-V_{h+1}^*\left((0,0)\right)\right)\\&+\left(q^0_{11}q^1_{01}-q^0_{01}q^1_{11}\right)\left(V_{h+1}^*\left((1,1)\right)-2V_{h+1}^*\left((1,0)\right)+V_{h+1}^*\left((0,0)\right)\right)\\
    &\geq \frac{\left(q^1_{10}I(0)-q^1_{00}I(1)\right)}{R(1)-R(0)}\left(V_{h+1}^*\left((1,0)\right)-V_{h+1}^*\left((0,0)\right)\right)\\
    &\geq 0,
\end{align*}
where the first inequality holds because the assumption we make and the last inequality holds because $q^1_{10}\geq q^1_{00}$, $I(0)\geq I(1)$ and the assumption that Eq.~\eqref{eq:monotonicity-10-case2-sub1} holds.

Next we want to prove the RHS of Eq.~\eqref{eq:monotonicity-11-case2-sub1} under the assumption that Eq.~\eqref{eq:optimality-lemma3-case2-sub1} holds for $t=h$ and that Eq.~\eqref{eq:monotonicity-10-case2-sub1} and Eq.~\eqref{eq:monotonicity-11-case2-sub1} hold for $t=h+1$ .
\begin{align*}
     V_{h}^*\left((1,1)\right)-V_h^*\left((1,0)\right)&=R(1)-R(0)+\left(q^1_{11}-q^1_{01}\right)\left(V_{h+1}^*\left((1,0)\right)-V_{h+1}^*\left((0,0)\right)\right)\\
    &+q^0_{11}\left(q^1_{11}-q^1_{01}\right)\left(V_{h+1}^*\left((1,1)\right)-2V_{h+1}^*\left((1,0)\right)+V_{h+1}^*\left((0,0)\right)\right)\\
    &=R(1)-R(0)+q^0_{11}\left(q^1_{11}-q^1_{01}\right)\left(V_{h+1}^*\left((1,1)\right)-V_{h+1}^*\left((1,0)\right)\right)\\&+q^0_{10}\left(q^1_{11}-q^1_{01}\right)\left(V_{h+1}^*\left((1,0)\right)-V_{h+1}^*\left((0,0)\right)\right)\\
    &\leq R(1)-R(0).
\end{align*}
Under the same condition, for the LHS of Eq.~\eqref{eq:monotonicity-11-case2-sub1}, we have
\begin{align*}
     V_{h}^*\left((1,1)\right)-V_h^*\left((1,0)\right)
    &=R(1)-R(0)+q^0_{11}\left(q^1_{11}-q^1_{01}\right)\left(V_{h+1}^*\left((1,1)\right)-V_{h+1}^*\left((1,0)\right)\right)\\&+q^0_{10}\left(q^1_{11}-q^1_{01}\right)\left(V_{h+1}^*\left((1,0)\right)-V_{h+1}^*\left((0,0)\right)\right)\\
    &\geq R(1)-R(0)+q^0_{10}\left(q^1_{11}-q^1_{01}\right)(R(1)-R(0))\\&+q^0_{11}\left(q^1_{11}-q^1_{01}\right)(R(1)-R(0))\\
    &=(q^1_{11}+q^1_{00})(R(1)-R(0))\\
    &\geq 0.
\end{align*}
Similarly, we can show Eq.~\eqref{eq:monotonicity-10-case2-sub1} for $t=h$ under the same condition. 

Secondly, we consider the sub-case where $\boxed{\bm{q^1_{11}q^0_{01}\geq q^1_{01}q^0_{11}}}$. We want to use backward induction to prove the following statements:
\begin{align}
    \forall t\leq T, &0\leq V_t^*\left((1,0)\right)-V_t^*((0,0))\leq{R(1)-R(0)}\label{eq:monotonicity-10-case2-sub2}\\
    &0\leq V_t^*\left((1,1)\right)-V_t^*((1,0))\leq R(1)-R(0)\label{eq:monotonicity-11-case2-sub2}\\
    &V_t^*\left((1,1)\right)-V_t^*((1,0))\leq V_t^*\left((1,0)\right)-V_t^*((0,0))\label{eq:monotonicity-1110-case2-sub2}\\
    &\pi_{t+1}^{\text{greedy}}=\pi_{t+1}^*.\label{eq:optimality-lemma3-case2-sub2}
\end{align}
When $t=T$, the statements hold trivially. Suppose this holds for $t\geq h+1$, when $t=h$, we will first show Eq.~\eqref{eq:optimality-lemma3-case2-sub2} for $t=h$ under the assumption that Eq.~\eqref{eq:monotonicity-10-case2-sub2}, Eq.~\eqref{eq:monotonicity-11-case2-sub2} and Eq.~\eqref{eq:monotonicity-1110-case2-sub2} hold for $t=h+1$. Similarly, this is equivalent to show Eq.~\eqref{eq:equation-showing-optimality}.
\begin{align*}
    &\left(q^0_{11}+q^1_{01}-q^1_{11}-q^0_{01}\right)\left(V_{h+1}^*\left((1,0)\right)-V_{h+1}^*\left((0,0)\right)\right)\\&+\left(q^0_{11}q^1_{01}-q^0_{01}q^1_{11}\right)\left(V_{h+1}^*\left((1,1)\right)-2V_{h+1}^*\left((1,0)\right)+V_{h+1}^*\left((0,0)\right)\right)\\
    &=\frac{I(0)-I(1)}{R(1)-R(0)}\left(V_{h+1}^*(1,0)-V_{h+1}^*(0,0)\right)\\
    &+\left(q^0_{11}q^1_{01}-q^0_{01}q^1_{11}\right)\left(V_{h+1}^*\left((1,1)\right)-2V_{h+1}^*\left((1,0)\right)+V_{h+1}^*\left((0,0)\right)\right)\\
    &\geq 0,
\end{align*}
where the first equality holds because of Eq.~\eqref{eq:simplification-incremental-reward-0} and Eq.~\eqref{eq:simplification-incremental-reward-1} and the first inequality holds since $q^0_{11}q^1_{01}-q^0_{01}q^1_{11}\leq 0$ and the assumption that Eq.~\eqref{eq:monotonicity-1110-case2-sub2} holds for $t=h+1$.

Using the same logic as the previous case, we can prove that 
\begin{align*}
    &0\leq V_h^*\left((1,0)\right)-V_h^*((0,0))\leq{R(1)-R(0)}\\
    &0\leq V_h^*\left((1,1)\right)-V_h^*((1,0))\leq R(1)-R(0).
\end{align*}
Therefore, the only thing we need to show is the Eq.~\eqref{eq:monotonicity-1110-case2-sub2} for $t=h$ under the assumption that Eq.~\eqref{eq:optimality-lemma3-case2-sub2} holds for $t=h$ and that Eq.~\eqref{eq:monotonicity-1110-case2-sub2} holds for $t=h+1$. We have
\begin{align*}
    &V_h^*\left((1,1)\right)-2V_h^*((1,0))+V_h^*((0,0))\\&=q^0_{11}\left(q^1_{11}-q^1_{01}\right)\left(V_{h+1}^*\left((1,1)\right)-2V_{h+1}^*((1,0))+V_{h+1}^*((0,0))\right)\\
    &+\left(q^1_{11}-q^1_{01}\right)\left(V_{h+1}^*\left((1,0)\right)-V_{h+1}^*\left((0,0)\right)\right)\\
    &-q^1_{01}\left(q^0_{11}-q^0_{01}\right)\left(V_{h+1}^*\left((1,1)\right)-2V_{h+1}^*((1,0))+V_{h+1}^*((0,0))\right)\\
    &-\left(q^0_{11}-q^0_{01}\right)\left(V_{h+1}^*\left((1,0)\right)-V_{h+1}^*\left((0,0)\right)\right)\\
    &=q^0_{11}\left(q^1_{11}-q^1_{01}\right)\left(V_{h+1}^*\left((1,1)\right)-2V_{h+1}^*((1,0))+V_{h+1}^*((0,0))\right)\\
    &-q^1_{01}\left(q^0_{11}-q^0_{01}\right)\left(V_{h+1}^*\left((1,1)\right)-2V_{h+1}^*((1,0))+V_{h+1}^*((0,0))\right)\\
    &+\frac{I(1)-I(0)}{R(1)-R(0)}\left(V_{h+1}^*\left((1,0)\right)-V_{h+1}^*\left((0,0)\right)\right).
\end{align*}
Since $I(1)\leq I(0),V_{h+1}^*\left((1,0)\right)-V_{h+1}^*\left((0,0)\right)\geq 0$, we only need to prove that 
\begin{equation*}
    q^0_{11}\left(q^1_{11}-q^1_{01}\right)\geq q^1_{01}\left(q^0_{11}-q^0_{01}\right).
\end{equation*}
Since we have
\begin{equation*}
    q^1_{11}q^0_{01}\geq q^1_{01}q^0_{11},
\end{equation*}
we have
\begin{align*}
    q^1_{11}q^0_{01}-q^1_{11}q^0_{11}&\geq q^1_{01}q^0_{11}-q^1_{11}q^0_{11}\\
    q^1_{11}\left(q^0_{01}-q^0_{11}\right)&\geq q^0_{11}(q^1_{01}-q^1_{11})\\
    q^1_{01}\left(q^0_{01}-q^0_{11}\right)\geq q^1_{11}\left(q^0_{01}-q^0_{11}\right)&\geq q^0_{11}(q^1_{01}-q^1_{11})\\
    q^1_{01}\left(q^0_{11}-q^0_{01}\right)&\leq q^0_{11}(q^1_{11}-q^1_{01}).
\end{align*}
This will give us the desired result.
\paragraph{Case 3:} $q^1_{11}\geq q^1_{01}, q^0_{11}\geq q^0_{01}$. 

We will use backward induction to prove the following statements:
\begin{align}
    \forall t\leq T, &\frac{R(1)-R(0)}{q^1_{10}+q^1_{01}}\left(1-\left(q^1_{11}-q^1_{01}\right)^{T-t+1}\right)\leq V_t^*\left((1,0)\right)-V_t^*((0,0))\label{eq:lowerbound-11-case3}\\
    &\frac{R(1)-R(0)}{q^0_{10}+q^0_{01}}\left(1-\left(q^0_{11}-q^0_{01}\right)^{T-t+1}\right)\geq V_t^*\left((1,0)\right)-V_t^*((0,0))\label{eq:upperbound-11-case3}\\
    &\frac{R(1)-R(0)}{q^1_{10}+q^1_{01}}\left(1-\left(q^1_{11}-q^1_{01}\right)^{T-t+1}\right)\leq V_t^*\left((1,0)\right)-V_t^*((0,0))\label{eq:lowerbound-10-case3}\\
    &\frac{R(1)-R(0)}{q^0_{10}+q^0_{01}}\left(1-\left(q^0_{11}-q^0_{01}\right)^{T-t+1}\right)\geq V_t^*\left((1,0)\right)-V_t^*((0,0))\label{eq:upperbound-10-case3}\\
    &\pi_{t}^{\text{greedy}}=\pi_{t}^*.\label{eq:optimality-lemma3-case3}
\end{align}
When $t=T$, this holds trivially, suppose this holds for $t\geq h+1$. When $t=h$, we will first show that Eq.~\eqref{eq:optimality-lemma3-case3} holds for $t=h$ under the assumption that Eq.~\eqref{eq:lowerbound-11-case3}, Eq.~\eqref{eq:upperbound-11-case3}, Eq.~\eqref{eq:lowerbound-10-case3} and Eq.~\eqref{eq:upperbound-10-case3} hold for $t=h+1$. This again is equivalent to show that
\begin{align*}
    &\left(q^0_{11}+q^1_{01}-q^1_{11}-q^0_{01}\right)\left(V_{h+1}^*\left((1,0)\right)-V_{h+1}^*\left((0,0)\right)\right)\\&+\left(q^0_{11}q^1_{01}-q^0_{01}q^1_{11}\right)\left(V_{h+1}^*\left((1,1)\right)-2V_{h+1}^*\left((1,0)\right)+V_{h+1}^*\left((0,0)\right)\right)\geq 0.
\end{align*}
We have
\begin{align*}
    &\left(q^0_{11}+q^1_{01}-q^1_{11}-q^0_{01}\right)\left(V_{h+1}^*\left((1,0)\right)-V_{h+1}^*\left((0,0)\right)\right)\\&+\left(q^0_{11}q^1_{01}-q^0_{01}q^1_{11}\right)\left(V_{h+1}^*\left((1,1)\right)-2V_{h+1}^*\left((1,0)\right)+V_{h+1}^*\left((0,0)\right)\right)\\
    &=\left(q^0_{11}+q^1_{01}-q^1_{11}-q^0_{01}-q^0_{11}q^1_{01}-q^0_{01}q^1_{11}\right)\left(V_{h+1}^*\left((1,0)\right)-V_{h+1}^*\left((0,0)\right)\right)\\
    &+\left(q^0_{11}q^1_{01}-q^0_{01}q^1_{11}\right)\left(V_{h+1}^*\left((1,1)\right)-V_{h+1}^*\left((1,0)\right)\right).
\end{align*}
If $q^0_{11}+q^1_{01}-q^1_{11}-q^0_{01}-q^0_{11}q^1_{01}-q^0_{01}q^1_{11}\geq 0$, then using our assumption that, $V_{h+1}^*\left((1,1)\right)-V_{h+1}^*\left((1,0)\right)\geq 0, V_{h+1}^*\left((1,0)\right)-V_{h+1}^*\left((0,0)\right)\geq 0$, will give us the desired result.

If $q^0_{11}+q^1_{01}-q^1_{11}-q^0_{01}-q^0_{11}q^1_{01}-q^0_{01}q^1_{11}\leq 0$, with our hypothesis, we have
\begin{align}
    &\left(q^0_{11}+q^1_{01}-q^1_{11}-q^0_{01}-q^0_{11}q^1_{01}-q^0_{01}q^1_{11}\right)\left(V_{h+1}^*\left((1,0)\right)-V_{h+1}^*\left((0,0)\right)\right)\notag\\
    &+\left(q^0_{11}q^1_{01}-q^0_{01}q^1_{11}\right)\left(V_{h+1}^*\left((1,1)\right)-V_{h+1}^*\left((1,0)\right)\right)\notag\\
    &\geq \left(q^0_{11}+q^1_{01}-q^1_{11}-q^0_{01}-q^0_{11}q^1_{01}-q^0_{01}q^1_{11}\right)\notag\\&\times\frac{R(1)-R(0)}{q^0_{10}+q^0_{01}}\left(1-\left(q^0_{11}-q^0_{01}\right)^{T-h}\right)\notag\\
    &+\left(q^0_{11}q^1_{01}-q^0_{01}q^1_{11}\right)\frac{R(1)-R(0)}{q^1_{10}+q^1_{01}}\left(1-\left(q^1_{11}-q^1_{01}\right)^{T-h}\right).\label{eq:state-1-better-optimality}
\end{align}
Denote $\frac{R(1)-R(0)}{q^0_{10}+q^0_{01}}\left(1-\left(q^0_{11}-q^0_{01}\right)^{T-h}\right)$ as $d_{\text{upper}}$ and let $k$ be defined as\\ $\frac{R(1)-R(0)}{q^1_{10}+q^1_{01}}\left(1-\left(q^1_{11}-q^1_{01}\right)^{T-h}\right)/d_{\text{upper}}$. Then to show that Eq.~\eqref{eq:state-1-better-optimality} is larger than 0, we only need to show that
\begin{equation*}
    q^0_{11}+q^1_{01}-q^1_{11}-q^0_{01}-(k-1)\left(q^0_{11}q^1_{01}-q^0_{01}q^1_{11}\right)\geq 0,
\end{equation*}
which is equivalent to show $k\geq \frac{q^0_{11}q^1_{01}-q^0_{01}q^1_{11}-q^0_{11}-q^1_{01}+q^1_{11}+q^0_{01}}{q^0_{11}q^1_{01}-q^0_{01}q^1_{11}}$. 

First, we will show 
\begin{equation*}
    \frac{q^0_{10}+q^0_{01}}{q^1_{10}+q^1_{01}}\geq \frac{q^0_{11}q^1_{01}-q^0_{01}q^1_{11}-q^0_{11}-q^1_{01}+q^1_{11}+q^0_{01}}{q^0_{11}q^1_{01}-q^0_{01}q^1_{11}}.
\end{equation*}
To do this, we only need to show 
\begin{align*}
    &\left(q^0_{10}+q^0_{01}\right)\left(q^0_{11}q^1_{01}-q^0_{01}q^1_{11}\right)\\&-\left(q^1_{10}+q^1_{01}\right)\left(q^0_{11}q^1_{01}-q^0_{01}q^1_{11}\right)\\&-\left(q^1_{10}+q^1_{01}\right)\left(q^1_{11}+q^0_{01}-q^0_{11}-q^1_{01}\right)\geq 0.
\end{align*}
We have
\begin{align*}
    &\left(q^0_{10}+q^0_{01}\right)\left(q^0_{11}q^1_{01}-q^0_{01}q^1_{11}\right)\\&-\left(q^1_{10}+q^1_{01}\right)\left(q^0_{11}q^1_{01}-q^0_{01}q^1_{11}\right)\\&-\left(q^1_{10}+q^1_{01}\right)\left(q^1_{11}+q^0_{01}-q^0_{11}-q^1_{01}\right)\\
    &=\left(q^0_{10}+q^0_{01}-q^1_{10}-q^1_{01}\right)\left(q^0_{11}q^1_{01}-q^0_{01}q^1_{11}\right)\\
    &-\left(q^1_{10}+q^1_{01}\right)\left(q^1_{11}+q^0_{01}-q^0_{11}-q^1_{01}\right)\\
    &=\left(q^1_{11}+q^0_{01}-q^0_{11}-q^1_{01}\right)\left(q^0_{11}q^1_{01}-q^0_{01}q^1_{11}-q^1_{10}-q^1_{01}\right).
\end{align*}
Since $q^1_{11}+q^0_{01}-q^0_{11}-q^1_{01}=\frac{I(1)-I(0)}{R(1)-R(0)}\leq 0$ and $q^0_{11}q^1_{01}-q^1_{01}\leq 0$, we have the desired result.

Second, we want to show that $1-\left(q^1_{11}-q^1_{01}\right)^{T-h}\geq 1-\left(q^0_{11}-q^0_{01}\right)^{T-h}$. To do this, we only need to show that $\left(q^1_{11}-q^1_{01}\right)^{T-h}\leq \left(q^0_{11}-q^0_{01}\right)^{T-h}$, which comes directly from the fact that $0\leq q^1_{11}-q^1_{01}\leq q^0_{11}-q^0_{01}$.

Then we have
\begin{align*}
    k&=  \frac{q^0_{10}+q^0_{01}}{q^1_{10}+q^1_{01}}\frac{1-\left(q^1_{11}-q^1_{01}\right)^{T-h}}{1-\left(q^0_{11}-q^0_{01}\right)^{T-h}}\\&\geq \frac{q^0_{10}+q^0_{01}}{q^1_{10}+q^1_{01}}\\
    &\geq \frac{q^0_{11}q^1_{01}-q^0_{01}q^1_{11}-q^0_{11}-q^1_{01}+q^1_{11}+q^0_{01}}{q^0_{11}q^1_{01}-q^0_{01}q^1_{11}}.
\end{align*}
This will give us the desired result.

Next, we will first prove Eq.~\eqref{eq:lowerbound-11-case3} for $t=h$ under the assumption that Eq.~\eqref{eq:lowerbound-11-case3}, Eq.~\eqref{eq:upperbound-11-case3}, Eq.~\eqref{eq:lowerbound-10-case3} and Eq.~\eqref{eq:upperbound-10-case3} hold for $t=h+1$ and Eq.~\eqref{eq:optimality-lemma3-case2-sub2} holds for $t=h$ . We have
\begin{align*}
    V_{h}^*\left((1,1)\right)-V_h^*\left((1,0)\right)&=R(1)-R(0)+\left(q^1_{11}-q^1_{01}\right)\left(V_{h+1}^*\left((1,0)\right)-V_{h+1}^*\left((0,0)\right)\right)\\
    &+q^0_{11}\left(q^1_{11}-q^1_{01}\right)\left(V_{h+1}^*\left((1,1)\right)-2V_{h+1}^*\left((1,0)\right)+V_{h+1}^*\left((0,0)\right)\right)\\
    &= R(1)-R(0)+q^0_{11}\left(q^1_{11}-q^1_{01}\right)\left(V_{h+1}^*\left((1,1)\right)-V_{h+1}^*\left((1,0)\right)\right)\\&+q^0_{10}\left(q^1_{11}-q^1_{01}\right)\left(V_{h+1}^*\left((1,0)\right)-V_{h+1}^*\left((0,0)\right)\right)\\
    &\geq R(1)-R(0)\\&+\left(q^1_{11}-q^1_{01}\right)\frac{R(1)-R(0)}{q^1_{10}+q^1_{01}}\left(1-\left(q^1_{11}-q^1_{01}\right)^{T-h}\right)\\
    &=\frac{R(1)-R(0)}{q^1_{10}+q^1_{01}}\left(1-\left(q^1_{11}-q^1_{01}\right)^{T-h+1}\right).
\end{align*}
Under the same condition, for Eq.~\eqref{eq:upperbound-11-case3} when $t=h$, we have
\begin{align*}
    V_{h}^*\left((1,1)\right)-V_h^*\left((1,0)\right)&=R(1)-R(0)+\left(q^1_{11}-q^1_{01}\right)\left(V_{h+1}^*\left((1,0)\right)-V_{h+1}^*\left((0,0)\right)\right)\\
    &+q^0_{11}\left(q^1_{11}-q^1_{01}\right)\left(V_{h+1}^*\left((1,1)\right)-2V_{h+1}^*\left((1,0)\right)+V_{h+1}^*\left((0,0)\right)\right)\\
    &= R(1)-R(0)+q^0_{11}\left(q^1_{11}-q^1_{01}\right)\left(V_{h+1}^*\left((1,1)\right)-V_{h+1}^*\left((1,0)\right)\right)\\&+q^0_{10}\left(q^1_{11}-q^1_{01}\right)\left(V_{h+1}^*\left((1,0)\right)-V_{h+1}^*\left((0,0)\right)\right)\\
    &\leq R(1)-R(0)\\&+\left(q^0_{11}-q^0_{01}\right)\frac{R(1)-R(0)}{q^0_{10}+q^0_{01}}\left(1-\left(q^0_{11}-q^0_{01}\right)^{T-h}\right)\\
    &=\frac{R(1)-R(0)}{q^0_{10}+q^0_{01}}\left(1-\left(q^0_{11}-q^0_{01}\right)^{T-h+1}\right),
\end{align*}
where the inequality holds because we have $q^1_{11}-q^1_{01}\leq q^0_{11}-q^0_{01}$. Using the same logic will give us the Eq.~\eqref{eq:upperbound-10-case3} and Eq.~\eqref{eq:lowerbound-10-case3} for $t=h$.
\paragraph{Case 4:} Suppose $q^1_{11}\leq q^1_{01},q_{11}^0\geq q_{01}^0,q_{11}^0\geq q_{11}^1$. We will directly show that $V_t^{\bm\pi^{\text{greedy}}}$ satisfies the optimal Bellman equation. With the same logic, we only need to prove that Eq.~\eqref{eq:equation-showing-optimality} holds for each $t\leq T$. We will first assume that the following holds and will prove this later:
\begin{equation}\label{eq:negative-11-10-00}
    V_t^{\bm\pi^{\text{greedy}}}((1,1))+V_t^{\bm{\pi}^{\text{greedy}}}((0,0))-2V_t^{\bm{\pi}^{\text{greedy}}}((1,0))\leq 0,~\forall t\leq T.
\end{equation}
Then we have 
\begin{align*}
    &\left(q_{11}^0+q_{01}^1-q_{11}^1-q_{01}^0\right)\left(V_{t+1}^{\Greedy}((1,0))-V_{t+1}^{\Greedy}((0,0))\right)\\
    &+\left(q_{01}^1q_{11}^0-q_{01}^0q_{11}^1\right)\left(V_{t+1}^{\bm\pi^{\text{greedy}}}((1,1))+V_{t+1}^{\bm{\pi}^{\text{greedy}}}((0,0))-2V_{t+1}^{\bm{\pi}^{\text{greedy}}}((1,0))\right)\\
    &=\left(q_{11}^0+q_{01}^1-q_{11}^1-q_{01}^0\right)\left(V_{t+1}^{\Greedy}((1,0))-V_{t+1}^{\Greedy}((0,0))\right)\\
    &+\left(q_{11}^1(q_{11}^0-q_{01}^0)-q_{11}^0(q_{11}^1-q_{01}^1)\right)\left(V_{t+1}^{\bm\pi^{\text{greedy}}}((1,1))+V_{t+1}^{\bm{\pi}^{\text{greedy}}}((0,0))-2V_{t+1}^{\bm{\pi}^{\text{greedy}}}((1,0))\right)\\
    &\geq \left(q_{11}^0+q_{01}^1-q_{11}^1-q_{01}^0\right)\left(V_{t+1}^{\Greedy}((1,0))-V_{t+1}^{\Greedy}((0,0))\right)\\
    &+\left(q_{01}^1(q_{11}^0-q_{01}^0)-q_{11}^0(q_{11}^1-q_{01}^1)\right)\left(V_{t+1}^{\bm\pi^{\text{greedy}}}((1,1))+V_{t+1}^{\bm{\pi}^{\text{greedy}}}((0,0))-2V_{t+1}^{\bm{\pi}^{\text{greedy}}}((1,0))\right)\\
    &=-V_t^{\bm\pi^{\text{greedy}}}((1,1))-V_t^{\bm{\pi}^{\text{greedy}}}((0,0))+2V_t^{\bm{\pi}^{\text{greedy}}}((1,0))\\
    &\geq 0.
\end{align*}

Next, we will prove Eq.~\eqref{eq:negative-11-10-00}. To ease notation, we will let $a:=1-q_{01}^1,b:=1-q_{11}^0, c=q_{11}^0-q_{01}^0, d=q_{11}^1-q_{01}^1$. By Eq.~\eqref{eq:simplification-difference-1110} and \eqref{eq:simplification-difference-1000}, we have the following iterative equations
\begin{align*}
    V_t^{\Greedy}((1,0))-V_t^{\Greedy}((0,0))&=R(1)-R(0)+ac\left( V_{t+1}^{\Greedy}((1,0))-V_{t+1}^{\Greedy}((0,0))\right)\\&+(1-a)c \left(V_{t+1}^{\Greedy}((1,1))-V_{t+1}^{\Greedy}((1,0))\right)\\
    V_t^{\Greedy}((1,1))-V_t^{\Greedy}((1,0))&=R(1)-R(0)+bd\left( V_{t+1}^{\Greedy}((1,0))-V_{t+1}^{\Greedy}((0,0))\right)\\&+(1-b)d \left(V_{t+1}^{\Greedy}((1,1))-V_{t+1}^{\Greedy}((1,0))\right).
\end{align*}
Let $\bm{V}_t=\begin{pmatrix}
    V_t^{\Greedy}((1,0))-V_t^{\Greedy}((0,0))\\V_t^{\Greedy}((1,1))-V_t^{\Greedy}((1,0))
\end{pmatrix}, A=\begin{pmatrix}
    ac&(1-a)c\\
    bd&(1-b)d
\end{pmatrix}$. We have 
\begin{equation*}
    \bm{V}_t=(R(1)-R(0))\bm{1}+A\bm{V}_{t+1}.
\end{equation*}
First we consider the subcase where $I-A$ is not invertible, which is $A$ has eigenvalue 1. This is equivalent to $\boxed{q_{11}^0=1,q_{01}^0=q_{01}^1=q_{11}^1=0}$. Under this case, we have $V_{t}^{\Greedy}((1,1))-V_t^{\Greedy}((1,0))=R(1)-R(0)$. We could also easily use induction to show that
\begin{equation*}
     V_t^{\Greedy}((1,0))-V_t^{\Greedy}((0,0))=R(1)-R(0)+ V_{t+1}^{\Greedy}((1,0))-V_{t+1}^{\Greedy}((0,0))\geq R(1)-R(0).
\end{equation*}
This will give us the desired result.

Next, we assume $I-A$ is invertible, then we have
\begin{align*}
    \bm{V}_t&=(R(1)-R(0))(I-A)^{-1}\bm{1}+(R(1)-R(0))A^{T-t}\left(\bm{1}-(I-A)^{-1}\bm{1}\right)\\
    &=(R(1)-R(0))(I-A)^{-1}\bm{1}-(R(1)-R(0))A^{T-t+1}(I-A)^{-1}\bm{1}\\
    &=\sum_{k=0}^{T-t}(R(1)-R(0))A^k\bm{1}.
\end{align*}
It is easy to see that $A$ has two eigenvalues:
\begin{equation*}
    \lambda_1=\frac{\eta+\sqrt{\eta^2-4\delta}}{2}~,\lambda_2=\frac{\eta-\sqrt{\eta^2-4\delta}}{2},
\end{equation*}
where $\eta=ac+(1-b)d, \delta=(a-b)cd$.

Then, we first consider the case $\boxed{\lambda_i\in\R}$. We have
\begin{align*}
    -V_t^{\bm\pi^{\text{greedy}}}((1,1))-V_t^{\bm{\pi}^{\text{greedy}}}((0,0))+2V_t^{\bm{\pi}^{\text{greedy}}}((1,0))&=(1,-1)\bm{V}_t\bm{1}\\
    &=(R(1)-R(0))(c-d)\sum_{k=0}^{T-t}\frac{\lambda_1^k-\lambda_2^k}{\lambda_1-\lambda_2}.
\end{align*}
Since under Case 4, $c\geq d$, we only need to prove that $\sum \frac{\lambda_1^k-\lambda_2^k}{\lambda_1-\lambda_2}\geq 0$.

When $\eta\geq 0$, $|\lambda_1|\geq |\lambda_2|$. Therefore, for all $k$, $\lambda_1^k\geq \lambda_2^k$, which will give us the desired result.

When $\eta\leq 0, \delta\geq 0$, we have $-1\leq\eta\leq \lambda_2\leq\lambda_1\leq 0$. First, we will prove $\lambda_1+\lambda_1^2\geq \lambda_2+\lambda_2^2$:
\begin{align*}
    \lambda_1+\lambda_1^2-\lambda_2-\lambda_2^2&=(\lambda_1-\lambda_2)(\lambda_1+\lambda_2+1)\\
    &=\delta(1+\eta)\\
    &\geq 0.
\end{align*}
Therefore, we have $\frac{\lambda_1}{\lambda_2}\leq \frac{1+\lambda_2}{1+\lambda_1}$. Then we can have for each odd $k$
\begin{align*}
    \frac{\lambda_1^k}{\lambda_2^k}\leq \frac{\lambda_1}{\lambda_2}\leq\frac{1+\lambda_2}{1+\lambda_1}.
\end{align*}
This means for each odd $k$, $\lambda_1^k+\lambda_1^{k+1}\geq \lambda_2^k+\lambda_2^{k+1}$. Then we have for all $T-t,\sum_{k=0}^{T-t}\lambda_1^k-\lambda_2^k\geq 0$. This will give us the desired result.

When $\eta\leq 0, \delta\leq 0$, we can numerically show that $\lambda_2\geq -1$. Specifically, we use gradient descent with a sufficient small step size ($10^{-5}$) and multiple initialization by choosing the grid points to calculate the solutions of the KKT conditions. The results shows that the minimum is larger than $-1$. Then we have $\sum_{k=0}^{T-t}\lambda_2^k\leq 1$. We also have that $\sum_{k=0}^{T-t} \lambda_1^k\geq 1$ since $\lambda_1\geq 0$. This will give us the desired result.

Finally, we consider the subcase where $\boxed{\lambda_i\in\mathbb{C}}$. We note that this will only happen when $\delta\geq 0$ which means $q_{11}^0\leq q_{01}^1$. We will first rewrite the eigenvalues using its spherical form, i.e. $\lambda_i=\rho\exp(\pm i\theta)$, where 
\begin{equation*}
    \rho=\sqrt{(a-b)cd}~, \cos(\theta)=\frac{ac+(1-b)d}{2\sqrt{(a-b)cd}}.
\end{equation*}
Then we have
\begin{align*}
     &-V_t^{\bm\pi^{\text{greedy}}}((1,1))-V_t^{\bm{\pi}^{\text{greedy}}}((0,0))+2V_t^{\bm{\pi}^{\text{greedy}}}((1,0))=(1,-1)\bm{V}_t\bm{1}\\
    &=(R(1)-R(0))(c-d)\sum_{k=0}^{T-t}\frac{\lambda_1^k-\lambda_2^k}{\lambda_1-\lambda_2}\\
    &=(R(1)-R(0))\frac{c-d}{1-2\rho\cos(\theta)+\rho^2}\left(1-\rho^{T-t}\frac{\sin((T-t+1)\theta-\rho\sin((T-t)\theta))}{\sin(\theta)}\right).
\end{align*}
Since $c-d\geq 0, 1-2\rho\cos(\theta)+\rho^2=|1-\rho \exp(i\theta)|^2\geq 0$, we only need to show that
\begin{equation*}
    1-\rho^{T-t}\frac{\sin((T-t+1)\theta-\rho\sin((T-t)\theta))}{\sin(\theta)}\geq 0.
\end{equation*}
Since $0\leq \rho\leq 1$, we only need to show the following:
\begin{align*}
    \rho^{T-t}\frac{\sin((T-t+1)\theta-\sin((T-t)\theta))}{\sin(\theta)}&\leq 1\\
    \rho^{T-t}\frac{\sin((T-t+1)\theta)}{\sin(\theta)}&\leq 1.
\end{align*}
We will use induction to prove the following:
\begin{align*}
    \rho^{k}\frac{\sin((k+1)\theta-\sin((k)\theta))}{\sin(\theta)}&\leq 1\\
    \rho^{k}\frac{\sin((k+1)\theta)}{\sin(\theta)}&\leq 1.
\end{align*}
When $k=0$, the statements trivially hold. Suppose it holds for $k=t$, when $k=t+1$, we have
\begin{align*}
    \rho^{t+1}\frac{\sin((t+2)\theta-\sin((t+1)\theta))}{\sin(\theta)}&=\rho^{t+1}\frac{\cos((t+1)\theta+\theta/2)}{\cos(\theta/2)}\\
    &=\rho^{t+1}\frac{\cos(t\theta+\theta/2)\cos(\theta)-\sin(t\theta+\theta/2)\sin(\theta)}{\cos(\theta/2)}\\
    &\leq \rho\cos(\theta)+2\sin(\theta/2)\rho.
\end{align*}

Similarly, we have
\begin{align*}
     \rho^{t+1}\frac{\sin((t+2)\theta)}{\sin(\theta)}&=\rho^{t+1}\frac{\sin((t+1)\theta)\cos(\theta)+\sin(\theta)\cos((t+1)\theta)}{\sin(\theta)}\\
     &\leq \rho\cos(\theta)+\rho\\
     &=\frac{1}{2}\left(ac+(1-b)d+2\sqrt{(q_{01}^1-q_{11}^1)(q_{01}^1-q_{11}^0)(q_{11}^0-q_{01}^0)}\right)\\
     &\leq \frac{1}{2}\left((1-q_{01}^1)q_{11}^0+q_{11}^0(q_{11}^1-q_{01}^1)+2(q_{01}^1-q_{11}^1)\right)\\
     &\leq \frac{1}{2}\left((1-q_{01}^1)q_{11}^0+(2-q_{11}^0)q_{01}^1\right)\\
     &\leq 1.
\end{align*}

Now we are going to prove the theorem for arbitrary number of agents and budget. We will again prove the result under three cases.

\paragraph{Case 1:} Suppose $q^1_{11}\leq q^1_{01},q^0_{11}\geq q^0_{01},q_{11}^1\geq q_{11}^0$.

 We will first define three sets: $\mathcal{S}^{M-2}_1,\mathcal{S}^{M-2}_2,\mathcal{S}^{M-2}_3$. 
\begin{align*}
    \mathcal{S}^{M-2}_1&:=\{\bm{s}:\bm{s}\in\mathcal{S}^{M-2}, n_{\bm{s}}\geq B\}\\
    \mathcal{S}^{M-2}_2&:=\{\bm{s}:\bm{s}\in\mathcal{S}^{M-2}, n_{\bm{s}}= B-1\}\\
    \mathcal{S}^{M-2}_3&:=\{\bm{s}:\bm{s}\in\mathcal{S}^{M-2}, n_{\bm{s}}\leq B-2\},
\end{align*}
where $n_{\bm{s}}$ denotes the number of state $0$ of $\bm{s}$.

We will use backward induction to prove that 
\begin{align}
    \forall t\leq T, \forall \bm{s}\in\mathcal{S}^{M-2}_3, &0\leq V_t^*((1,0,\bm{s}))-V_t^*((0,0,\bm{s}))\leq R(1)-R(0)\label{eq:monotonicity-103-case1}\\
    & 0\leq V_t^*((1,1,\bm{s}))-V_t^*((1,0,\bm{s}))\leq R(1)-R(0)\label{eq:monotonicity-113-case1}\\
    \forall \bm{s}\in\mathcal{S}^{M-2}_2, &0\leq V_t^*((1,0,\bm{s}))-V_t^*((0,0,\bm{s}))\leq \frac{R(1)-R(0)}{q^0_{10}}\label{eq:monotonicity-102-case1}\\
    & 0\leq V_t^*((1,1,\bm{s}))-V_t^*((1,0,\bm{s}))\leq R(1)-R(0)\label{eq:monotonicity-112-case1}\\
    \forall \bm{s}\in\mathcal{S}^{M-2}_1,&0\leq V_t^*((1,0,\bm{s}))-V_t^*((0,0,\bm{s}))\leq \frac{R(1)-R(0)}{q^0_{10}}\label{eq:monotonicity-101-case1}\\
    & 0\leq V_t^*((1,1,\bm{s}))-V_t^*((1,0,\bm{s}))\leq \frac{R(1)-R(0)}{q^0_{10}}\label{eq:monotonicity-111-case1}\\
    &\pi_{t+1}^{\greedy}=\pi_{t+1}^*.\label{eq:optimality-th-case1-ma}
\end{align}
This holds trivially when $t=T$. Suppose $t\geq h+1$, this holds, we will first prove Eq.~\eqref{eq:optimality-th-case1-ma} for $t=h$ under the assumption that Eq.~\eqref{eq:monotonicity-103-case1}, Eq.~\eqref{eq:monotonicity-113-case1}, Eq.~\eqref{eq:monotonicity-102-case1}, Eq.~\eqref{eq:monotonicity-112-case1}, Eq.~\eqref{eq:monotonicity-101-case1} and Eq.~\eqref{eq:monotonicity-111-case1} hold for $t=h+1$.
Similarly we have the following Bellman optimality equation.
\begin{align*}
    V_h^*\left(\left(1,0,\bm{s}_{h}\right)\right)&=R(1)+R(0)+\sum_{s\in\bm{s}_h}R(s)\\&+\max_{a_{h+1}^1,a_{h+1}^2,\cdots,a_{h+1}^M}\sum_{s_{h+1}^1,s_{h+1}^2}q_{1,s_{h+1}^1}^{a_{h+1}^1}q_{0,s_{h+1}^2}^{a_{h+1}^2}\sum_{\bm{s}_{h+1}}q_{\bm{s}_h,\bm{s}_{h+1}}^{\bm{a}_{h+1}}V_{h+1}^*\left((s_{h+1}^1,s_{h+1}^2,\bm{s}_{h+1})\right).
\end{align*}
We want to show that fix any policy of choosing $a_{h+1}^3,\cdots,a_{h+1}^M$, giving action 1 to agent who is in state 0 will give a higher value, which is
\begin{align}
    \forall \bm{a}_{h+1}, &\sum_{s_{h+1}^1,s_{h+1}^2}q^0_{1s_{h+1}^1}q^1_{0s_{h+1}^2}\sum_{\bm{s}_{h+1}}q_{\bm{s}_h,\bm{s}_{h+1}}^{\bm{a}_{h+1}}V_{h+1}^*\left((s_{h+1}^1,s_{h+1}^2,\bm{s}_{h+1})\right)\notag\\&\geq \sum_{s_{h+1}^1,s_{h+1}^2}q^1_{1s_{h+1}^1}q^0_{0s_{h+1}^2}\sum_{\bm{s}_{h+1}}q_{\bm{s}_h,\bm{s}_{h+1}}^{\bm{a}_{h+1}}V_{h+1}^*\left((s_{h+1}^1,s_{h+1}^2,\bm{s}_{h+1})\right).\label{eq:switching-is-better}
\end{align}
Suppose Eq.~\eqref{eq:switching-is-better} holds, starting from any policy of allocating action that is not greedy, we can find an agent who is in state 1 and gets action 1 while another agent who is in state 0 gets action 0. By using the result, switching the action between them will give us a better policy. Suppose there are less than $B$ agents who are in state 0, then we can keep switching until all agents who are in state 0 are given action 1 , this is the greedy policy. Suppose there are more than $B$ agents who are in state 0, then we can keep switching until $B$ agents who are in state 0 get action 1, which is the greedy policy.

Now, we need to prove Eq.~\eqref{eq:switching-is-better}. Eq.~\eqref{eq:switching-is-better} is equivalent to
\begin{align}
    &\left(q^0_{11}+q^1_{01}-q^1_{11}-q^0_{01}\right)\left(F_{\bm{a}_{h+1},h+1}((1,1))-F_{\bm{a}_{h+1},h+1}((1,0))\right)\notag\\&+\left(q^0_{11}q^1_{01}-q^0_{01}q^1_{11}\right)\left(F_{\bm{a}_{h+1},h+1}\left((1,1)\right)-2F_{\bm{a}_{h+1},h+1}\left((1,0)\right)+F_{\bm{a}_{h+1},h+1}\left((0,0)\right)\right)\geq 0,\label{eq:showing-optimality-multiagent}
\end{align}
where $F_{\bm{a}_{h+1},h+1}\left((s_{h+1}^1,s_{h+1}^2)\right)=\sum_{\bm{s}_{h+1}}q_{\bm{s}_h,\bm{s}_{h+1}}^{\bm{a}_{h+1}}V_{h+1}^*\left((s_{h+1}^1,s_{h+1}^2,\bm{s}_{h+1})\right)$. The reason why this holds is similar to what we proved in binary agent case. We know from the assumption that Eq.~\eqref{eq:monotonicity-103-case1}-\eqref{eq:monotonicity-111-case1} hold and the definition of $F$ that 
\begin{align*}
   0&\leq F_{\bm{a}_{h+1},h+1}((1,1))-F_{\bm{a}_{h+1},h+1}((1,0))\leq \frac{R(1)-R(0)}{q^0_{10}}\\
    0&\leq F_{\bm{a}_{h+1},h+1}((1,0))-F_{\bm{a}_{h+1},h+1}((0,0))\leq \frac{R(1)-R(0)}{q^0_{10}}.
\end{align*}
Then, we have
\begin{align*}
       &\left(q^0_{11}+q^1_{01}-q^1_{11}-q^0_{01}\right)\left(F_{\bm{a}_{h+1},h+1}\left((1,0)\right)-F_{\bm{a}_{h+1},h+1}\left((0,0)\right)\right)\\&+\left(q^0_{11}q^1_{01}-q^0_{01}q^1_{11}\right)\left(F_{\bm{a}_{h+1},h+1}\left((1,1)\right)-2F_{\bm{a}_{h+1},h+1}\left((1,0)\right)+F_{\bm{a}_{h+1},h+1}\left((0,0)\right)\right)\\
    &\geq \left(q^0_{11}+q^1_{01}-q^1_{11}-q^0_{01}\right)\left(F_{\bm{a}_{h+1},h+1}\left((1,0)\right)-F_{\bm{a}_{h+1},h+1}\left((0,0)\right)\right)\\
    &-\left(q^0_{11}q^1_{01}-q^0_{01}q^1_{11}\right)\left(F_{\bm{a}_{h+1},h+1}\left((1,0)\right)-F_{\bm{a}_{h+1},h+1}\left((0,0)\right)\right)\\
    &=\frac{I(0)-I(1)}{R(1)-R(0)}\left(F_{\bm{a}_{h+1},h+1}\left((1,0)\right)-F_{\bm{a}_{h+1},h+1}\left((0,0)\right)\right)-\frac{\left(q^1_{11}I(0)-q^1_{01}I(1)\right)}{R(1)-R(0)}\\&\left(F_{\bm{a}_{h+1},h+1}\left((1,0)\right)-F_{\bm{a}_{h+1},h+1}\left((0,0)\right)\right)\\
    &=\frac{\left(q^1_{10}I(0)-q^1_{00}I(1)\right)}{R(1)-R(0)}\left(F_{\bm{a}_{h+1},h+1}\left((1,0)\right)-F_{\bm{a}_{h+1},h+1}\left((0,0)\right)\right)\\
    &\geq 0.
\end{align*}
This shows Eq.~\eqref{eq:switching-is-better}, which, by our argument, shows Eq.~\eqref{eq:optimality-th-case1-ma} for $t=h$. Next, we will first define the specific greedy policy instead of randomly choosing agents who have higher incremental reward at time $h$ for different $\bm{s}$. When $\bm{s}\in\mathcal{S}^{M-2}_1$, we will only give action 1 to agent $3,4,\cdots,M$; when $\bm{s}\in\mathcal{S}^{M-2}_2$, we will give one action 1 to agent $1,2$; when $\bm{s}\in\mathcal{S}^{M-2}_3$, we will give two action 1 to agent $1,2$.

For $\bm{s}\in\mathcal{S}_1^{M-2}$, we will show Eq.~\eqref{eq:monotonicity-101-case1} holds for $t=h$ under the assumption that Eq.\eqref{eq:monotonicity-103-case1}-\eqref{eq:monotonicity-111-case1} hold for $t=h+1$ and Eq.~\eqref{eq:optimality-th-case1-ma} holds for $t=h$. We have
\begin{align*}
    V_h^*\left((1,0,\bm{s})\right)-V_h^*\left((0,0,\bm{s})\right)&=R(1)-R(0)+\left(q^0_{11}-q^0_{01}\right)\left(F_{h+1}^*\left((1,0)\right)-F_{h+1}^*\left((0,0)\right)\right)\\
    &+q^0_{01}\left(q^0_{11}-q^0_{01}\right)\left(F_{h+1}^*\left((1,1)\right)-2F_{h+1}^*\left((1,0)\right)+F_{h+1}^*\left((0,0)\right)\right)\\
    &= R(1)-R(0)+q^0_{01}\left(q^0_{11}-q^0_{01}\right)\left(F_{h+1}^*\left((1,1)\right)-F_{h+1}^*\left((1,0)\right)\right)\\&+q^0_{00}\left(q^0_{11}-q^0_{01}\right)\left(F_{h+1}^*\left((1,0)\right)-F_{h+1}^*\left((0,0)\right)\right)\\
    &\geq 0.
\end{align*}
Next, we will show the RHS of Eq.~\eqref{eq:monotonicity-101-case1} for $t=h$ under the same condition. 
\begin{align*}
    q_{10}^0\left(V_h^*\left((1,0,\bm{s})\right)-V_h^*\left((0,0,\bm{s})\right)\right)&=q_{10}^0(R(1)-R(0))+q_{10}^0\left(q^0_{11}-q^0_{01}\right)\left(F_{h+1}^*\left((1,0)\right)-F_{h+1}^*\left((0,0)\right)\right)\\
    &+q_{10}^0q^0_{01}\left(q^0_{11}-q^0_{01}\right)\left(F_{h+1}^*\left((1,1)\right)-2F_{h+1}^*\left((1,0)\right)+F_{h+1}^*\left((0,0)\right)\right)\\
    &= q_{10}^0(R(1)-R(0))+q^0_{10}q^0_{01}\left(q^0_{11}-q^0_{01}\right)\left(F_{h+1}^*\left((1,1)\right)-F_{h+1}^*\left((1,0)\right)\right)\\&+q_{10}^0q^0_{00}\left(q^0_{11}-q^0_{01}\right)\left(F_{h+1}^*\left((1,0)\right)-F_{h+1}^*\left((0,0)\right)\right)\\
    &\leq q_{10}^0(R(1)-R(0))+q_{01}^0q_{11}^0(R(1)-R(0))+q_{00}^0q_{11}^0(R(1)-R(0))\\
    &=R(1)-R(0).
\end{align*}
Meanwhile, when $t=h$, again under the same condition, the LHS of Eq.~\eqref{eq:monotonicity-111-case1} satisfies 
\begin{align*}
    V_h^*\left((1,1,\bm{s})\right)-V_h^*\left((1,0,\bm{s})\right)&=R(1)-R(0)+\left(q^0_{11}-q^0_{01}\right)\left(F_{h+1}^*\left((1,0)\right)-F_{h+1}^*\left((0,0)\right)\right)\\
    &+q^0_{11}\left(q^0_{11}-q^0_{01}\right)\left(F_{h+1}^*\left((1,1)\right)-2F_{h+1}^*\left((1,0)\right)+F_{h+1}^*\left((0,0)\right)\right)\\
    &= R(1)-R(0)+q^0_{11}\left(q^0_{11}-q^0_{01}\right)\left(F_{h+1}^*\left((1,1)\right)-F_{h+1}^*\left((1,0)\right)\right)\\&+q^0_{10}\left(q^0_{11}-q^0_{01}\right)\left(F_{h+1}^*\left((1,0)\right)-F_{h+1}^*\left((0,0)\right)\right)\\
    &\geq 0.
\end{align*}
With the same reasoning, we can show the RHS of Eq.~\eqref{eq:monotonicity-111-case1} under the same condition for $t=h$.

For $\bm{s}\in\mathcal{S}_3^{M-2}$, we have
\begin{align*}
    V_h^*\left((1,0,\bm{s})\right)-V_h^*\left((0,0,\bm{s})\right)&=R(1)-R(0)+\left(q^1_{11}-q^1_{01}\right)\left(F_{h+1}^*\left((1,0)\right)-F_{h+1}^*\left((0,0)\right)\right)\\
    &+q^1_{01}\left(q^1_{11}-q^1_{01}\right)\left(F_{h+1}^*\left((1,1)\right)-2F_{h+1}^*\left((1,0)\right)+F_{h+1}^*\left((0,0)\right)\right)\\
    &= R(1)-R(0)+q^1_{01}\left(q^1_{11}-q^1_{01}\right)\left(F_{h+1}^*\left((1,1)\right)-F_{h+1}^*\left((1,0)\right)\right)\\&+q^1_{00}\left(q^1_{11}-q^1_{01}\right)\left(F_{h+1}^*\left((1,0)\right)-F_{h+1}^*\left((0,0)\right)\right)\\
    &= \frac{q^0_{10}}{q^0_{10}}\left(R(1)-R(0)+q^1_{01}\left(q^1_{11}-q^1_{01}\right)\left(F_{h+1}^*\left((1,1)\right)-F_{h+1}^*\left((1,0)\right)\right)\right)\\&+\frac{q^0_{10}}{q^0_{10}}\left(q^1_{00}\left(q^1_{11}-q^1_{01}\right)\left(F_{h+1}^*\left((1,0)\right)-F_{h+1}^*\left((0,0)\right)\right)\right)\\
    &\geq  \frac{q^0_{10}}{q^0_{10}}\left(R(1)-R(0)+\frac{q^1_{01}}{q^0_{10}}\left(q^1_{11}-q^1_{01}\right)\left(R(1)-R(0)\right)\right)\\&+\frac{q^1_{00}}{q^0_{10}}\left(q^1_{11}-q^1_{01}\right)\left(R(1)-R(0)\right)\\
    &=\frac{1}{q^0_{10}}\left(q^0_{10}+q^1_{00}q^1_{11}-q^1_{00}q^1_{01}\right)\left(R(1)-R(0)\right)\\&+\frac{1}{q^0_{10}}\left(q^1_{01}\left(q^1_{11}-q^1_{01}\right)\right)\left(R(1)-R(0)\right)\\
    &\geq \frac{1}{q^0_{10}}\left(1-q^1_{01}+q^1_{11}-q^0_{11}\right)\left(R(1)-R(0)\right)\\
    &\geq 0.
\end{align*}
Meanwhile, we also have
\begin{align*}
    V_h^*\left((1,1,\bm{s})\right)-V_h^*\left((1,0,\bm{s})\right)&=R(1)-R(0)+\left(q^1_{11}-q^1_{01}\right)\left(F_{h+1}^*\left((1,0)\right)-F_{h+1}^*\left((0,0)\right)\right)\\
    &+q^1_{11}\left(q^1_{11}-q^1_{01}\right)\left(F_{h+1}^*\left((1,1)\right)-2F_{h+1}^*\left((1,0)\right)+F_{h+1}^*\left((0,0)\right)\right)\\
    &= R(1)-R(0)+q^1_{11}\left(q^1_{11}-q^1_{01}\right)\left(F_{h+1}^*\left((1,1)\right)-F_{h+1}^*\left((1,0)\right)\right)\\&+q^1_{10}\left(q^1_{11}-q^1_{01}\right)\left(F_{h+1}^*\left((1,0)\right)-F_{h+1}^*\left((0,0)\right)\right)\\
    &= \frac{q^0_{10}}{q^0_{10}}\left(R(1)-R(0)+q^1_{11}\left(q^1_{11}-q^1_{01}\right)\left(F_{h+1}^*\left((1,1)\right)-F_{h+1}^*\left((1,0)\right)\right)\right)\\&+\frac{q^0_{10}}{q^0_{10}}\left(q^1_{10}\left(q^1_{11}-q^1_{01}\right)\left(F_{h+1}^*\left((1,0)\right)-F_{h+1}^*\left((0,0)\right)\right)\right)\\
    &\geq  \frac{q^0_{10}}{q^0_{10}}\left(R(1)-R(0)+\frac{q^1_{11}}{q^0_{10}}\left(q^1_{11}-q^1_{01}\right)\left(R(1)-R(0)\right)\right)\\&+\frac{q^1_{10}}{q^0_{10}}\left(q^1_{11}-q^1_{01}\right)\left(R(1)-R(0)\right)\\
    &=\frac{1}{q^0_{10}}\left(q^0_{10}+q^1_{10}q^1_{11}-q^1_{10}q^1_{01}\right)\left(R(1)-R(0)\right)\\&+\frac{1}{q^0_{10}}\left(q^1_{11}\left(q^1_{11}-q^1_{01}\right)\right)\left(R(1)-R(0)\right)\\
    &\geq \frac{1}{q^0_{10}}\left(1-q^1_{01}+q^1_{11}-q^0_{11}\right)\left(R(1)-R(0)\right)\\
    &\geq 0.
\end{align*}

For $\bm{s}\in\mathcal{S}_2^{M-2}$, we can use the same reasoning used in the two agent case to get the result.

\paragraph{Case 2:} $q^0_{11}\leq q^0_{01},q^1_{11}\leq q^1_{01}$. 

First we consider the sub-case where $\boxed{\bm{q^1_{11}q^0_{01}\leq q^1_{01}q^0_{11}}}$. We will use backward induction to prove the following statements:
\begin{align}
    \forall t\leq T, &0\leq V_t^*\left((1,0,\bm{s})\right)-V_t^*((0,0,\bm{s}))\leq R(1)-R(0)\label{eq:monotonicity-10-case2-sub1-ma}\\
    &0\leq V_t^*\left((1,1,\bm{s})\right)-V_t^*((1,0,\bm{s}))\leq R(1)-R(0)\label{eq:monotonicity-11-case2-sub1-ma}\\
    &\pi_{t+1}^{\text{greedy}}=\pi_{t+1}^*.\label{eq:optimality-lemma3-case2-sub1-ma}
\end{align}
They hold trivially when $t=T$. Suppose they hold when $t\geq h+1$, when $t=h$, we will first prove Eq.~\eqref{eq:optimality-lemma3-case2-sub1-ma} for $t=h$ under the assumption that Eq.~\eqref{eq:monotonicity-10-case2-sub1-ma} and Eq.~\eqref{eq:monotonicity-11-case2-sub1} hold for $t=h+1$. Using the same reasoning, we only need to show Eq.~\eqref{eq:showing-optimality-multiagent}. Again, from the assumptions and the definition of $F$, we know that 
\begin{align*}
   0&\leq F_{\bm{a}_{h+1},h+1}((1,1))-F_{\bm{a}_{h+1},h+1}((1,0))\leq {R(1)-R(0)}\\
    0&\leq F_{\bm{a}_{h+1},h+1}((1,0))-F_{\bm{a}_{h+1},h+1}((0,0))\leq {R(1)-R(0)}.
\end{align*}
Therefore, we have
\begin{align*}
       &\left(q^0_{11}+q^1_{01}-q^1_{11}-q^0_{01}\right)\left(F_{\bm{a}_{h+1},h+1}\left((1,0)\right)-F_{\bm{a}_{h+1},h+1}\left((0,0)\right)\right)\\&+\left(q^0_{11}q^1_{01}-q^0_{01}q^1_{11}\right)\left(F_{\bm{a}_{h+1},h+1}\left((1,1)\right)-2F_{\bm{a}_{h+1},h+1}\left((1,0)\right)+F_{\bm{a}_{h+1},h+1}\left((0,0)\right)\right)\\
    &\geq \left(q^0_{11}+q^1_{01}-q^1_{11}-q^0_{01}\right)\left(F_{\bm{a}_{h+1},h+1}\left((1,0)\right)-F_{\bm{a}_{h+1},h+1}\left((0,0)\right)\right)\\
    &-\left(q^0_{11}q^1_{01}-q^0_{01}q^1_{11}\right)\left(F_{\bm{a}_{h+1},h+1}\left((1,0)\right)-F_{\bm{a}_{h+1},h+1}\left((0,0)\right)\right)\\
    &=\frac{I(0)-I(1)}{R(1)-R(0)}\left(F_{\bm{a}_{h+1},h+1}\left((1,0)\right)-F_{\bm{a}_{h+1},h+1}\left((0,0)\right)\right)-\frac{\left(q^1_{11}I(0)-q^1_{01}I(1)\right)}{R(1)-R(0)}\\&\left(F_{\bm{a}_{h+1},h+1}\left((1,0)\right)-F_{\bm{a}_{h+1},h+1}\left((0,0)\right)\right)\\
    &=\frac{\left(q^1_{10}I(0)-q^1_{00}I(1)\right)}{R(1)-R(0)}\left(F_{\bm{a}_{h+1},h+1}\left((1,0)\right)-F_{\bm{a}_{h+1},h+1}\left((0,0)\right)\right)\\
    &\geq 0.
\end{align*}
This will give us the desired result. Assuming Eq.~\eqref{eq:optimality-lemma3-case2-sub1-ma} holds for $t=h$,
we will select the greedy policy for $\mathcal{S}_i^{M-2}$ as we did in case 1. Again, the only case that is different from the 2-agent case is when $\bm{s}\in\mathcal{S}_1^{M-2},\mathcal{S}_3^{M-2}$. Therefore, we will only show Eq.~\eqref{eq:monotonicity-10-case2-sub1-ma} and Eq.~\eqref{eq:monotonicity-11-case2-sub1-ma} for those two cases when $t=h$ under the assumption that Eq.~\eqref{eq:monotonicity-10-case2-sub1-ma} and Eq.~\eqref{eq:monotonicity-11-case2-sub1-ma} hold for $t=h+1$ and Eq.~\eqref{eq:optimality-lemma3-case2-sub1-ma} holds for $t=h$. We will first prove the RHS of Eq.~\eqref{eq:monotonicity-10-case2-sub1-ma}.
When $\bm{s}\in\mathcal{S}_1^{M-2}$, we have
\begin{align*}
    V_h^*\left((1,0,\bm{s})\right)-V_h^*\left((0,0,\bm{s})\right)&=R(1)-R(0)+\left(q^0_{11}-q^0_{01}\right)\left(F_{h+1}^*\left((1,0)\right)-F_{h+1}^*\left((0,0)\right)\right)\\
    &+q^0_{01}\left(q^0_{11}-q^0_{01}\right)\left(F_{h+1}^*\left((1,1)\right)-2F_{h+1}^*\left((1,0)\right)+F_{h+1}^*\left((0,0)\right)\right)\\
    &= R(1)-R(0)+q^0_{01}\left(q^0_{11}-q^0_{01}\right)\left(F_{h+1}^*\left((1,1)\right)-F_{h+1}^*\left((1,0)\right)\right)\\&+q^0_{00}\left(q^0_{11}-q^0_{01}\right)\left(F_{h+1}^*\left((1,0)\right)-F_{h+1}^*\left((0,0)\right)\right)\\
    &\leq R(1)-R(0).
\end{align*}
Similarly, when $\bm{s}\in\mathcal{S}_3^{M-2}$, we have
\begin{align*}
    V_h^*\left((1,0,\bm{s})\right)-V_h^*\left((0,0,\bm{s})\right)&=R(1)-R(0)+\left(q^1_{11}-q^1_{01}\right)\left(F_{h+1}^*\left((1,0)\right)-F_{h+1}^*\left((0,0)\right)\right)\\
    &+q^1_{01}\left(q^1_{11}-q^1_{01}\right)\left(F_{h+1}^*\left((1,1)\right)-2F_{h+1}^*\left((1,0)\right)+F_{h+1}^*\left((0,0)\right)\right)\\
    &= R(1)-R(0)+q^1_{01}\left(q^1_{11}-q^1_{01}\right)\left(F_{h+1}^*\left((1,1)\right)-F_{h+1}^*\left((1,0)\right)\right)\\&+q^1_{00}\left(q^1_{11}-q^1_{01}\right)\left(F_{h+1}^*\left((1,0)\right)-F_{h+1}^*\left((0,0)\right)\right)\\
    &\leq R(1)-R(0).
\end{align*}
Now, we are going to prove the LHS of Eq.~\eqref{eq:monotonicity-10-case2-sub1-ma}. 
When $\bm{s}\in\mathcal{S}_1^{M-2}$, we have
\begin{align*}
    V_h^*\left((1,0,\bm{s})\right)-V_h^*\left((0,0,\bm{s})\right)&=R(1)-R(0)+\left(q^0_{11}-q^0_{01}\right)\left(F_{h+1}^*\left((1,0)\right)-F_{h+1}^*\left((0,0)\right)\right)\\
    &+q^0_{01}\left(q^0_{11}-q^0_{01}\right)\left(F_{h+1}^*\left((1,1)\right)-2F_{h+1}^*\left((1,0)\right)+F_{h+1}^*\left((0,0)\right)\right)\\
    &= R(1)-R(0)+q^0_{01}\left(q^0_{11}-q^0_{01}\right)\left(F_{h+1}^*\left((1,1)\right)-F_{h+1}^*\left((1,0)\right)\right)\\&+q^0_{00}\left(q^0_{11}-q^0_{01}\right)\left(F_{h+1}^*\left((1,0)\right)-F_{h+1}^*\left((0,0)\right)\right)\\
    &\geq R(1)-R(0)+q^0_{01}\left(q^0_{11}-q^0_{01}\right)(R(1)-R(0))+q^0_{00}\left(q^0_{11}-q^0_{01}\right)(R(1)-R(0))\\
    &=(1+q_{11}^0+q_{00}^0)(R(1)-R(0))\\
    &\geq 0.
\end{align*}
Using the same calculation, we can show the case for $\bm{s}\in\mathcal{S}_3^{M-2}$ and Eq.~\eqref{eq:monotonicity-11-case2-sub1-ma}.

Secondly, we consider the sub-case where $\boxed{\bm{q^1_{11}q^0_{01}\geq q^1_{01}q^0_{11}}}$. We want to use backward induction to prove the following statements:
\begin{align}
    \forall t\leq T, &0\leq V_t^*\left((1,0,\bm{s})\right)-V_t^*((0,0,\bm{s}))\leq{R(1)-R(0)}\label{eq:monotonicity-10-case2-sub2-ma}\\
    &0\leq V_t^*\left((1,1,\bm{s})\right)-V_t^*((1,0,\bm{s}))\leq R(1)-R(0)\label{eq:monotonicity-11-case2-sub2-ma}\\
    &V_t^*\left((1,1,\bm{s})\right)-V_t^*((1,0,\bm{s}))\leq V_t^*\left((1,0,\bm{s})\right)-V_t^*((0,0,\bm{s}))\label{eq:monotonicity-1110-case2-sub2-ma}\\
    &\pi_{t+1}^{\text{greedy}}=\pi_{t+1}^*.\label{eq:optimality-lemma3-case2-sub2-ma}
\end{align}
When $t=T$, the statements hold trivially. Suppose this holds for $t\geq h+1$, when $t=h$, we will first show Eq.~\eqref{eq:optimality-lemma3-case2-sub2-ma} under the assumption that Eq.~\eqref{eq:monotonicity-10-case2-sub2-ma}-\eqref{eq:monotonicity-1110-case2-sub2-ma} hold for $t=h+1$. Using the same reasoning, we only need to show Eq.~\eqref{eq:showing-optimality-multiagent}. Again, from the above assumptions and the definition of $F$, we know that 
\begin{align*}
   0&\leq F_{\bm{a}_{h+1},h+1}((1,1))-F_{\bm{a}_{h+1},h+1}((1,0))\leq {R(1)-R(0)}\\
    0&\leq F_{\bm{a}_{h+1},h+1}((1,0))-F_{\bm{a}_{h+1},h+1}((0,0))\leq {R(1)-R(0)}.
\end{align*}
Then combined with the fact that ${{q^1_{11}q^0_{01}\geq q^1_{01}q^0_{11}}}$, it is easy to see that Eq.~\eqref{eq:showing-optimality-multiagent} holds. 

Using the same logic used in the first sub-case, we can prove Eq.~\eqref{eq:monotonicity-10-case2-sub2-ma},\eqref{eq:monotonicity-11-case2-sub2-ma} for $t=h$. The only thing we need to show is Eq.~\eqref{eq:monotonicity-1110-case2-sub2-ma} for $t=h$. Again, when $\bm{s}\in\mathcal{S}_2^{M-2}$, this is the same as the two agent case. Therefore, we will only prove the other two cases under the assumption that Eq.~\eqref{eq:monotonicity-10-case2-sub2-ma}-\eqref{eq:monotonicity-1110-case2-sub2-ma} hold for $t=h+1$ and Eq. \eqref{eq:optimality-lemma3-case2-sub2-ma} holds for $t=h$. We will prove the result for $\bm{s}\in\mathcal{S}_1^{M-2}$, the other cases will hold for similar reasons. We have
\begin{align*}
    &V_h^*\left((1,1,\bm{s})\right)-2V_h^*((1,0,\bm{s}))+V_h^*((0,0,\bm{s}))\\&=\left(q^0_{11}-q^0_{01}\right)\left(F_{h+1}^*\left((1,0)\right)-F_{h+1}^*\left((0,0)\right)\right)\\
    &+q^0_{11}\left(q^0_{11}-q^0_{01}\right)\left(F_{h+1}^*\left((1,1)\right)-2F_{h+1}^*\left((1,0)\right)+F_{h+1}^*\left((0,0)\right)\right)\\
    &-\left(q^0_{11}-q^0_{01}\right)\left(F_{h+1}^*\left((1,0)\right)-F_{h+1}^*\left((0,0)\right)\right)\\
    &-q^0_{01}\left(q^0_{11}-q^0_{01}\right)\left(F_{h+1}^*\left((1,1)\right)-2F_{h+1}^*\left((1,0)\right)+F_{h+1}^*\left((0,0)\right)\right)\\
    &=\left(q^0_{11}-q^0_{01}\right)^2\left(F_{h+1}^*\left((1,1)\right)-2F_{h+1}^*\left((1,0)\right)+F_{h+1}^*\left((0,0)\right)\right)\\
    &\leq 0,
\end{align*}
where the last inequality holds because of the assumption that Eq.~\eqref{eq:monotonicity-1110-case2-sub2-ma} holds for $t=h+1$. 
\paragraph{Case 3:} $q^1_{11}\geq q^1_{01}, q^0_{11}\geq q^0_{01}$. Similar to the two agent case, we can prove the following statements using backward induction:
\begin{align}
    \forall t\leq T, &\frac{R(1)-R(0)}{q^1_{10}+q^1_{01}}\left(1-\left(q^1_{11}-q^1_{01}\right)^{T-t+1}\right)\leq V_t^*\left((1,0,\bm{s})\right)-V_t^*((0,0,\bm{s}))\label{eq:lowerbound-11-case3-ma}\\
    &\frac{R(1)-R(0)}{q^0_{10}+q^0_{01}}\left(1-\left(q^0_{11}-q^0_{01}\right)^{T-t+1}\right)\geq V_t^*\left((1,0,\bm{s})\right)-V_t^*((0,0,\bm{s}))\label{eq:upperbound-11-case3-ma}\\
    &\frac{R(1)-R(0)}{q^1_{10}+q^1_{01}}\left(1-\left(q^1_{11}-q^1_{01}\right)^{T-t+1}\right)\leq V_t^*\left((1,0,\bm{s})\right)-V_t^*((0,0,\bm{s}))\label{eq:lowerbound-10-case3-ma}\\
    &\frac{R(1)-R(0)}{q^0_{10}+q^0_{01}}\left(1-\left(q^0_{11}-q^0_{01}\right)^{T-t+1}\right)\geq V_t^*\left((1,0,\bm{s})\right)-V_t^*((0,0,\bm{s}))\label{eq:upperbound-10-case3-ma}\\
    &\pi_{t+1}^{\text{greedy}}=\pi_{t+1}^*.\label{eq:optimality-lemma3-case3-ma}
\end{align}
The proof will be exactly using the idea used in 2 agent case and the calculation for multiple agents case discussed above, thus will be omitted.
\paragraph{Case 4:} Suppose $q^1_{11}\leq q^1_{01},q^0_{11}\geq q^0_{01},q_{11}^1\leq q_{11}^0$. We can use the same way used in 2 agent case to directly calculate $2V_t^{\Greedy}((1,0,\bm{s}))-V_t^{\Greedy}((0,0,\bm{s}))-V_t^{\Greedy}((1,1,\bm{s}))$ and have the result.
\end{proof}

\section{Technical Details In Section \ref{section:thresholding-bandit-algorithm}}\label{app:proof-thresholding}
We will continue to use the notation defined in Appendix \ref{appendix:proof-of-reduction}.
\subsection{Proof of Lemma \ref{le:enough-good-arm-all-states}}\label{proof:enough-good-arm-all-states}
\sufficientgoodarms*
\begin{proof}
    By Bellman equation, we have
    \begin{equation*}
V^{\bm{\pi}^b}_0\left(\bm{s}_0\right)=\E\left[\sum_{m=1}^Mc_0^m\right]+\E_{\pi_1^b}\left[V^{\bm{\pi}^b}_1(\bm{s}_1)\right].
    \end{equation*}
    Since $V^{\bm{\pi}^b}_0\left(\bm{s}_0\right)=0, V^{\bm{\pi}^b}_1(\bm{s}_1)\geq 0, P_{\bm{\pi}^b}(\bm{s}_0,\bm{s}_1)>0$, we have $V^{\bm{\pi}^b}_1(\bm{s}_1)=0$ for all $\bm{s}_1\in\mathcal{S}^M$. Since again by Bellman equation, we have $V^{\bm\pi^b}_1(\bm{s}_1)\geq \sum_{m=1}^M c_1^m\left(\pi_2^b(s_1^m)\right)$, we have this sum should be 0. This will indicate that there are enough good arms for state $\bm{s}_1$. This will give us the desired result.
\end{proof}
Before proving Theorem \ref{th:sublinearregret}, we first introduce Lemma \ref{le:regret-decomposition} which decomposes the cumulative regret to per-timestep regret.
\begin{lemma}\label{le:regret-decomposition}
    For any $t\leq T$, let per-timestep regret $\text{Regret}_t(\pi,\mathcal{H}_t)$ for the policy $\bm{\pi}$ be defined as:
    \begin{align*}
        \text{Regret}_t(\bm{\pi},\mathcal{H}_t)&=\E_{\pi_{t+1}}\left[\sum_{m=1}^M c_{t}^m(a_{t+1}^m)\mid\mathcal{H}_t\right]+\E_{\pi_{t+1}}\left[\E_{\bm{\pi}^b}\left[\sum_{m=1}^M\sum_{k=t+1}^T c_k^m\left(a_{t+1}^m\right)|\bm{r}_{t+1},\bm{s}_{t+1},\bm{a}_{t+1},\mathcal{H}_t\right]\right]\\&-\E_{\pi_{t+1}^b}\left[\sum_{m=1}^M c_{t}^m\left(a_{t+1}^m\right)\mid \mathcal{H}_t\right]-\E_{\pi_{t+1}^b}\left[\E_{\bm{\pi}^b}\left[\sum_{m=1}^M\sum_{k=t+1}^T c_k^m\left(a_{t+1}^m\right)|\bm{r}_{t+1},\bm{s}_{t+1},\bm{a}_{t+1},\mathcal{H}_t\right]\right].
    \end{align*} 
    Then the cumulative regret $\mathcal{R}(\pi)$ can be decomposed as following:
    \begin{equation*}
        \mathcal{R}(\pi)=\E\left[\sum_{t=0}^T\text{Regret}_t\left(\pi,\mathcal{H}_t\right)\right].
    \end{equation*}
\end{lemma}
\subsection{Proof of Lemma \ref{le:regret-decomposition}}
\begin{proof}
We will prove the result iteratively from $t=1$ to $T$. All we need is calculation. 
    We have
    \begin{align*}
        \mathcal{R}(\bm{\pi})&=\E_{\bm{\pi}}\left[\sum_{m=1}^M\sum_{t=0}^T c_t^m(a_{t+1}^m)|\bm{s}_0\right]-\E_{\bm{\pi}^b}\left[\sum_{m=1}^M\sum_{t=0}^T c_t^m(a_{t+1}^m)|\bm{s}_0\right]\\
        &=\E_{\pi_1}\left[\sum_{m=1}^M c_0^m(a_{1}^m)\mid\mathcal{H}_0\right]+\E_{\pi_1}\left[\E_{\bm{\pi}}\left[\sum_{m=1}^M\sum_{k=1}^T c_k^m\left(a_{k+1}^m\right)|\bm{r}_1,\bm{s}_1,\bm{a}_1,\mathcal{H}_0\right]\right]\\
        &-\E_{\pi_1}\left[\sum_{m=1}^M c_{0}^m(a_1^m)\mid\mathcal{H}_0\right]-\E_{\pi_1}\left[\E_{\bm{\pi}^b}\left[\sum_{m=1}^M\sum_{k=1}^T c_k^m\left(a_{k+1}^m\right)|\bm{r}_1,\bm{s}_1,\bm{a}_1,\mathcal{H}_0\right]\right]\\
        &+\E_{\pi_1}\left[\sum_{m=1}^M c_{0}^m(a_1^m)\mid\mathcal{H}_0\right]+\E_{\pi_1}\left[\E_{\bm{\pi}^b}\left[\sum_{m=1}^M\sum_{k=1}^T c_k^m\left(a_{k+1}^m\right)|\bm{r}_1, \bm{s}_1,\bm{a}_1,\mathcal{H}_0\right]\right]\\
        &-\E_{\pi_1^b}\left[\sum_{m=1}^M c_{0}^m\left(a_1^m\right)\mid\mathcal{H}_0\right]-\E_{\pi_1^b}\left[\E_{\bm{\pi}^b}\left[\sum_{m=1}^M\sum_{k=1}^T c_k^m\left(a_{k+1}^m\right)|\bm{r}_1,\bm{s}_1,\bm{a}_1,\mathcal{H}_0\right]\right]\\
        &=\E\left[\E_{\bm{\pi}}\left[\sum_{m=1}^M\sum_{k=1}^T c_k^m\left(a_{k+1}^m\right)|\mathcal{H}_1\right]-\E_{\bm{\pi}^b}\left[\sum_{m=1}^M\sum_{k=1}^T c_k^m\left(a_{k+1}^m\right)|\mathcal{H}_1\right]\right]\\
        &+\text{Regret}\left(\pi,\mathcal{H}_0\right).
    \end{align*}
    We can continue this trick to $\E_{\bm{\pi}}\left[\sum_{m=1}^M\sum_{k=1}^T c_k^m\left(a_{k+1}^m\right)|\mathcal{H}_1\right]-\E_{\bm{\pi}^b}\left[\sum_{m=1}^M\sum_{k=1}^T c_k^m\left(a_{k+1}^m\right)|\mathcal{H}_1\right]$, this will give us the desired result.
\end{proof}
\subsection{Proof of Theorem \ref{th:sublinearregret}}
\regretbound*
Before we prove the theorem, we first provide a technical lemma showing the concentration property of the empirical mean estimator $\hat{I}_t^m(s)$, which is a direct result from \citealt{10.1214/20-AOS1991}.
\begin{lemma}\label{le:concentration-empirical}
    For any $\alpha>0$, $0\leq t\leq T$, we have
    \begin{equation*}
        \prob\left(\left|\hat{I}_t^m(s)-I^m(s)\right|\geq 1.7\sqrt{\frac{\log\log(2N_{s,m}(t))+0.72\log(10.4/\alpha)}{N_{s,m}(t)}}\right)\leq \alpha.
    \end{equation*}
\end{lemma}
\begin{proof}
   The proof will proceed by first showing that the cumulative regret can be decomposed into per-timestep regret $\text{Regret}_t(\bm{\pi},\mathcal{H}_t)$. Then we will use the standard technique to bound the number of times suboptimal agents are given action 1. 
   
   By Lemma \ref{le:regret-decomposition}, we only need to bound $\text{Regret}_t(\pi,\mathcal{H}_t)$. We have
    \begin{align*}
         \text{Regret}_t(\pi,\mathcal{H}_t)&=\E_{\pi_{t+1}}\left[\sum_{m=1}^M c_{t}^m(a_{t+1}^m)\mid\mathcal{H}_t\right]+\E_{\pi_{t+1}}\left[\E_{\bm{\pi}^b}\left[\sum_{m=1}^M\sum_{k=t+1}^T c_k^m\left(a_{t+1}^m\right)|\bm{r}_{t+1},\bm{s}_{t+1},\bm{a}_{t+1},\mathcal{H}_t\right]\right]\\&-\E_{\pi_{t+1}^b}\left[\sum_{m=1}^M c_{t}^m\left(a_{t+1}^m\right)\mid \mathcal{H}_t\right]-\E_{\pi_{t+1}^b}\left[\E_{\bm{\pi}^b}\left[\sum_{m=1}^M\sum_{k=t+1}^T c_k^m\left(a_{t+1}^m\right)|\bm{r}_{t+1},\bm{s}_{t+1},\bm{a}_{t+1},\mathcal{H}_t\right]\right]\\
         &=\E_{\pi_{t+1}}\left[\sum_{m=1}^M c_{t}^m(a_{t+1}^m)\mid\mathcal{H}_t\right]+\E_{\pi_{t+1}}\left[\E_{\bm{\pi}^b}\left[\sum_{m=1}^M\sum_{k=t+1}^T c_k^m\left(a_{t+1}^m\right)|\bm{r}_{t+1},\bm{s}_{t+1},\bm{a}_{t+1},\mathcal{H}_t\right]\right]\\
         &\leq \E\left[\sum_{I^m(s)\leq\gamma}\Delta_s^m\mathbb{I}\left\{a_{t+1}^m=1,s_t^m=1\right\}\right]+\E_{\pi_{t+1}}\left[V_{t+1}^{\bm{\pi}^b}\left(\bm{s}_{t+1}\right)\mid \bm{s}_{t+1},\bm{a}_{t+1},\mathcal{H}_t\right]\\
         &=\E\left[\sum_{I^m(s)\leq\gamma}\Delta_s^m\mathbb{I}\left\{a_{t+1}^m=1,s_t^m=1\right\}\right],
    \end{align*}
    where the second and the third equality holds by Lemma \ref{le:enough-good-arm-all-states}. Then we have the cumulative regret is upper bounded by
    \begin{equation*}
        \sum_{I^m(s)\leq\gamma}\Delta_s^m\E\left[\sum_{t=0}^T \mathbb{I}\left\{a_{t+1}^m=1,s_t^m=s\right\}\right].
    \end{equation*}

    We have
\begin{align}
    \E\left[\sum_{t=0}^T \mathbb{I}\left\{a_{t+1}^m=1,s_t^m=s\right\}\right]&=\E\left[\sum_{t=0}^T \mathbb{I}\left\{a_{t+1}^m=1,s_t^m=s,\mathrm{random}\right\}\right]\label{eq:first-term-regret}\\&+\E\left[\sum_{t=0}^T \mathbb{I}\left\{a_{t+1}^m=1,s_t^m=s,\mathrm{not~random }\right\}\right]\label{eq:second-term-regret},
\end{align}
where $\mathrm{random}$ stands for doing random exploration.
We will bound the two terms separately. For Eq.~\eqref{eq:first-term-regret}, we have
\begin{align}
   &\E\left[\sum_{t=0}^T \mathbb{I}\left\{a_{t+1}^m=1,s_t^m=s,\mathrm{random}\right\}\right]\notag\\
   &=\E\left[\sum_{t=0}^T \mathbb{I}\left\{a_{t+1}^m=1,s_t^m=s,\mathrm{random},\exists (m',s'), s_t^{m'}=s',N_{s',m'}^{(t)}\leq N_{s_t^{m''},m''}^{(t)},\forall m''\right\}\right]\notag\\
   &\leq \sum_{m'}\sum_{s'}\E\left[\sum_{t=0}^T \mathbb{I}\left\{a_{t+1}^m=1,s_t^m=s,\mathrm{random}, s_t^{m'}=s',N_{s',m'}^{(t)}\leq N_{s_t^{m''},m''}^{(t)},\forall m''\right\}\right]\notag\\
   &\leq \sum_{m'}\sum_{s'}\E\left[\sum_{t=0}^T \mathbb{I}\left\{\mathrm{random}, s_t^{m'}=s',N_{s',m'}^{(t)}\leq N_{s_t^{m''},m''}^{(t)},\forall m''\right\}\right]\notag\\
   &\leq \sum_{m'}\sum_{s'}\E\left[\sum_{k=0}^\infty \mathbb{I}\left\{\mathrm{random},N_{s',m'}^{(\tau_k^{m'}(s'))}\leq N_{s_t^{m''},m''}^{(\tau_k^{m'}(s'))},\forall m'',\tau_k^{m'}(s')\leq T\right\}\right]\notag\\
   &\leq \sum_{m'}\sum_{s'}\E\left[\sum_{k=0}^T \mathbb{I}\left\{\mathrm{random},N_{s',m'}^{(\tau_k^{m'}(s'))}\leq N_{s_t^{m''},m''}^{(\tau_k^{m'}(s'))},\forall m''\right\}\right]\notag\\
   &=\sum_{m'}\sum_{s'}\sum_{k=0}^T\prob\left(\mathrm{random}\mid N_{s',m'}^{(\tau_k^{m'}(s'))}\leq N_{s_t^{m''},m''}^{(\tau_k^{m'}(s'))},\forall m''\right)\prob\left(N_{s',m'}^{(\tau_k^{m'}(s'))}\leq N_{s_t^{m''},m''}^{(\tau_k^{m'}(s'))},\forall m''\right)\notag\\
   &\leq \sum_{m'}\sum_{s'}\sum_{k=0}^T\prob\left(\mathrm{random}\mid N_{s',m'}^{(\tau_k^{m'}(s'))}\leq N_{s_t^{m''},m''}^{(\tau_k^{m'}(s'))},\forall m''\right)\notag\\
   &\leq\sum_{m'}\sum_{s'}\sum_{k=0}^T\frac{D}{k\Delta^2}\notag\\
   &\leq SMD\frac{\log(T)}{\Delta^2},\label{eq:regret-theorem-first-term}
\end{align}
where $\tau_k^m(s):=\inf\{t:N_{s,m}^{(t)}=k\}$.

For Eq.~\eqref{eq:second-term-regret}, we have
\begin{align*}
    &\E\left[\sum_{t=0}^T \mathbb{I}\left\{a_{t+1}^m=1,s_t^m=s,\mathrm{not~random}\right\}\right]\\
    &=\E\left[\sum_{t=0}^T \mathbb{I}\left\{a_{t+1}^m=1,s_t^m=s,\mathrm{not~random},\exists (m',s')\in\mathcal{G},\hat{I}_t^m(s)\geq \hat{I}_t^{m'}(s'),s_t^{m'}=s'\right\}\right]\\
    &\leq \sum_{(m',s')\in\mathcal{G}}\E\left[\sum_{t=0}^T \mathbb{I}\left\{a_{t+1}^m=1,s_t^m=s,\hat{I}_t^m(s)\geq \hat{I}_t^{m'}(s'),s_t^{m'}=s'\right\}\right]
\end{align*}
For each $(m',s')\in\mathcal{G}$, we have
\begin{align}
    &\E\left[\sum_{t=0}^T \mathbb{I}\left\{a_{t+1}^m=1,s_t^m=s,\hat{I}_t^m(s)\geq \hat{I}_t^{m'}(s'),s_t^{m'}=s'\right\}\right]\notag\\
    &\leq \E\left[\sum_{t=0}^T \mathbb{I}\left\{a_{t+1}^m=1,s_t^m=s,\hat{I}_t^m(s)\geq \gamma+\frac{\Delta}{2},s_t^{m'}=s'\right\}\right]\label{eq:bad-arm-bad}\\
    &+\E\left[\sum_{t=0}^T \mathbb{I}\left\{a_{t+1}^m=1,s_t^m=s,\hat{I}_t^{m'}(s')\leq \gamma+\frac{\Delta}{2},s_t^{m'}=s'\right\}\right]\label{eq:good-arm-bad}.
\end{align}
Again, for Eq.~\eqref{eq:bad-arm-bad}, we have
\begin{align*}
   & \E\left[\sum_{t=0}^T \mathbb{I}\left\{a_{t+1}^m=1,s_t^m=s,\hat{I}_t^m(s)\geq \gamma+\frac{\Delta}{2},s_t^{m'}=s'\right\}\right]\\
   &\leq \E\left[\sum_{t=0}^T \mathbb{I}\left\{s_t^m=s,\hat{I}_t^m(s)\geq \gamma+\frac{\Delta}{2}\right\}\right]\\
   &=\E\left[\sum_{k=0}^\infty \mathbb{I}\left\{s_{\tau_k^m(s)}^m=s,\hat{I}_{\tau_k^m(s)}^m(s)\geq \gamma+\frac{\Delta}{2}\right\},\tau_k^m(s)\leq T\right]\\
   &\leq\E\left[\sum_{k=0}^T \mathbb{I}\left\{\hat{I}_{\tau_k^m(s)}^m(s)\geq \gamma+\frac{\Delta}{2}\right\}\right]\\
   &=\E\left[\sum_{k=0}^T \mathbb{I}\left\{N_{s,m}^{(\tau_k^m(s))}(r)\in[k_1 t^k_{\mathrm{random}},k_2t^k_\mathrm{random}],\hat{I}_{\tau_k^m(s)}^m(s)\geq \gamma+\frac{\Delta}{2}\right\}\right]\\
   &+\E\left[\sum_{k=0}^T \mathbb{I}\left\{N_{s,m}^{(\tau_k^m(s))}(r)\notin[k_1 t^k_{\mathrm{random}},k_2t^k_\mathrm{random}],\hat{I}_{\tau_k^m(s)}^m(s)\geq \gamma+\frac{\Delta}{2}\right\}\right]\\
   &\leq \E\left[\sum_{k=0}^T \mathbb{I}\left\{N_{s,m}^{(\tau_k^m(s))}(r)\in[k_1 t^k_{\mathrm{random}},k_2t^k_\mathrm{random}],\hat{I}_{\tau_k^m(s)}^m(s)\geq \gamma+\frac{\Delta}{2},t^k_\mathrm{random}\geq k_3^k\right\}\right]\\
   &+\E\left[\sum_{k=0}^T \mathbb{I}\left\{N_{s,m}^{(\tau_k^m(s))}(r)\notin[k_1 t^k_{\mathrm{random}},k_2t^k_\mathrm{random}]\right\}\right]+\E\left[\sum_{k=0}^T \mathbb{I}\left\{t_\mathrm{random}^k\leq k_3^k\right\}\right],
\end{align*}
where $N_{s,m}^{(\tau_k^m(s))}(r):=\sum_{l=1}^k \mathbb{I}\left\{a_{\tau_l^m(s)}^m=0,s_{\tau_l^m(s)}^m=s,\mathrm{random}\right\}$ denotes the total number of times that action 0 is given to agent $m$ when $m$ is in state $s$ and the algorithm is doing random exploration, $t^k_{\mathrm{random}}:=\sum_{l=1}^m\sum_{l=1}^k \mathbb{I}\left\{\mathrm{random}\right\}$ denotes the total number of times when agent $m$ is in state $s$ and the algorithm is doing random explorations. We let $k_1=\max\left\{\frac{M-B}{2M},\frac{2M-3B}{2M}\right\}, k_2=\min\left\{\frac{3(M-B)}{2M},\frac{2M-B }{2(M-B)},k_3^k=\frac{C\log(k)}{2\min\{\Delta^2,1/2\}}\right\}$

For the first term, we have
\begin{align*}
    &\E\left[\sum_{k=0}^T \mathbb{I}\left\{N_{s,m}^{(\tau_k^m(s))}(r)\in[k_1 t^k_{\mathrm{random}},k_2t^k_\mathrm{random}],\hat{I}_{\tau_k^m(s)}^m(s)\geq \gamma+\frac{\Delta}{2},t^k_\mathrm{random}\geq k_3^k\right\}\right]\\
    &\leq\E\left[\sum_{k=0}^T \mathbb{I}\left\{N_{s,m}^{(\tau_k^m(s))}(r)\in[k_1 t^k_{\mathrm{random}},k_2t^k_\mathrm{random}],N_{s,m,1}^{(\tau_k^m(s)}\geq (1-k_2)t_{\mathrm{random}}^k,\hat{I}_{\tau_k^m(s)}^m(s)\geq \gamma+\frac{\Delta}{2},t^k_\mathrm{random}\geq k_3^k\right\}\right]\\
    &\leq \E\left[\sum_{k=0}^T \mathbb{I}\left\{N_{s,m}^{(\tau_k^m(s))}(r)\geq k_1k_3^k,N_{s,m,1}^{(\tau_k^m(s)}\geq (1-k_2)k_3^k,\hat{I}_{\tau_k^m(s)}^m(s)\geq \gamma+\frac{\Delta}{2}\right\}\right]\\
    &\leq \E\left[\sum_{k=0}^T \mathbb{I}\left\{N_{s,m}^{(\tau_k^m(s))}(r)\geq k_1k_3^k,N_{s,m,1}^{(\tau_k^m(s)}\geq (1-k_2)k_3^k,\left|\hat{I}_{\tau_k^m(s)}^m(s)-I^m(s)\right|\geq \frac{\Delta}{2}\right\}\right]\\
    &\leq \sum_{k=0}^T \frac{1}{k^2}\\
    &\leq 2,
\end{align*}
where the second last inequality holds due to Lemma~\ref{le:concentration-empirical}. Similarly, we can prove the last two terms are upper bounded by 2 using Lemma~\ref{le:concentration-empirical}. 

Using the same way, we can upper bound Eq.~\eqref{eq:good-arm-bad} by 6. Therefore, Eq.~\eqref{eq:second-term-regret} can be upper bounded as
\begin{align}
&\E\left[\sum_{t=0}^T \mathbb{I}\left\{a_{t+1}^m=1,s_t^m=s,\text{not~random}\right\}\right]\notag\\
    &=\sum_{(m',s')\in\mathcal{G}}\E\left[\sum_{t=0}^T\mathbb{I}\left\{a_{t+1}^m=1,s_t^m=s,\hat{I}_t^m(s)\geq\hat{I}_t^{m'}(s'),s_t^{m'}=s'\right\}\right]\notag\\
    &\leq 12|\mathcal{G}|.\label{eq:regret-theorem-second-term}
\end{align}
Further assume that $k\left|\Delta\right|\geq \Delta_{s}^m$, for all $I^m(s)\leq\gamma$. Then combining Eq.~\eqref{eq:regret-theorem-first-term} and Eq.~\eqref{eq:regret-theorem-second-term} will give us the desired result.

We also have
\begin{align*}
    \mathcal{R}(\pi)&\leq kB\Delta T+\frac{k|\mathcal{S}|MD(M|\mathcal{S}|-|\mathcal{G}|)}{\Delta}\log T\\
    &\leq k\sqrt{BMD|\mathcal{S}|(M|\mathcal{S}|-|\mathcal{G}|)T\log T}.
\end{align*}
\end{proof}
\section{Implementation Detail}\label{sec:implementation-detail}
All the experiments are conducted on MacBook Pro with Apple M3 Pro chip and 18 GB memory.
\subsection{Regret Curve}\label{subsec:implementation-detail-regret}
\begin{figure}[!t]
\begin{center}
\centerline{\includegraphics[width=0.4\columnwidth]{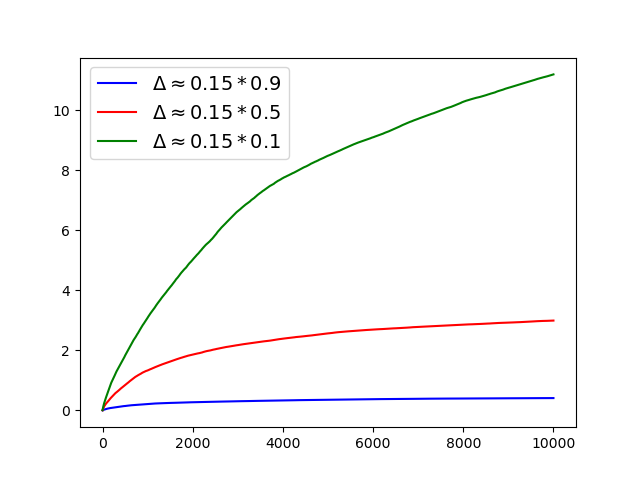}}
\caption{Regret Curve as horizon $T$ grows.}
\label{figure:regret-curve}
\end{center}
\end{figure}
To numerically 
verify Theorem \ref{th:sublinearregret}, 
we set 
$M=2$, 
$B=1$, and 
$\mathcal{S}=\{0,1\}$. We set the expected reward of the state as $R^m(s)=s$. We randomly generated $10$ instances and chose three $\gamma$ values satisfying Assumption~\ref{asssumption:enough-good-arm}, labeling each $\gamma$ by the corresponding gap $\Delta$ between the threshold and the smallest incremental reward. Specifically, we will randomly generate a number between $[0,1]$ independently and assign this number to be $P_{a}^m(s,1), \forall a,m,s\in\{0,1\}$. 

Without loss of generality, assume that $I^1(1)\geq I^1(0)\geq I^2(1)\geq I^2(0)$. We choose $\gamma$ to be between $I^1(0)$ and $I^2(0)$. Specifically, let $\theta\in(0,1)$ be a tunable parameter. We set
$\gamma:=\theta I^2(1)+(1-\theta)I^2(0)$. To make sure that the gap between $\gamma$ and $I^2(1)$ is not too small, we choose those instances where $I^2(1)-I^2(0)\geq 0.15$. 
To generate the three different $\gamma$ values in Figure~\ref{figure:regret-curve}, we selected
three different choices of $\theta$:
$0.1,0.5,0.9$, yielding the observed $\Delta$ values.

For each $\gamma$, the regret was averaged over $100$ repetitions per instance (Figure~\ref{figure:regret-curve}). 
Figure \ref{figure:regret-curve} demonstrates sublinear regret for all choices of \(\gamma\), with regret stabilizing at a constant as the horizon length increases, consistent with Theorem \ref{th:sublinearregret}. Further, as $\Delta$ decreases, regret increases as it becomes harder to distinguish whether the incremental reward is above the threshold or not. This matches the results in Theorem~\ref{th:sublinearregret}.

\subsection{Cumulative Reward}\label{subsec:exp-detail-cumulative}
We first describe the benchmarks we chose in detail:
\begin{itemize}
    \item Oracle Whittle: A variant of Whittle's index policy with access to the ground truth transitions and rewards, optimized for maximizing discounted reward (as described in Appendix \ref{app:whittle}). 
    \item Oracle Greedy:
    The greedy policy we defined (Algorithm~\ref{alg:lcb-greedy}) with knowledge of the ground truth incremental reward.
    \item OMR: A modified LP-based policy for maximizing finite-horizon reward.
    \item WIQL \citep{biswas2021learn}:  A Q-learning based approach to
learn
the value function of each arm at each state by interacting
with the RMAB instance.
We use the same learning rate ($\alpha$) as in \citet{biswas2021learn}.
\item UCWhittle-penalty \citep{wang2023optimistic}: It constructs the confidence interval of the unknown transition probability and then chooses one to calculate Whittle's index policy. Specifically, it constructs a confidence interval and chooses the transition probability that will maximize $V_\lambda^m(s)$ (Eq.~\eqref{eq:dynamic-programming-discounted}) for some adaptively updated $\lambda$.
\item R$(\text{MA})^2$B: It uses the explore-then-commit framework. It first randomly gives action and constructs the confidence interval of the unknown transition. After the exploration period, it uses the estimated transition to calculate the OMR policy. 
\end{itemize}

In all of the settings, the discount factor $\beta$ was set to $0.9$.
We note that though $\beta=1$ was chosen in \citet{biswas2021learn}, $\beta=0.9$ has better empirical performance, which aligns with the choice in 
\citet{wang2023optimistic}.

Then, we will discuss more on the implementation of the algorithms.
First of all 
for the UCWhittle families introduced in \citet{wang2023optimistic}, the algorithms
reinitialize the initial state after certain time-steps. We do not include this episodic structure as this contradicts our goal of maximizing the finite-horizon cumulative reward.
We keep the rest structures of their algorithms as the same.

We additionally note that the code provided by
\citet{wang2023optimistic} contains errors and could not be used to 
reproduce the results shown in \citet{wang2023optimistic}.
A mistake was found in
Line 73 of uc$\_$whittle.py that they provided. Specifically, they use the current state of an agent as the index to get the current action of this agent. 
As a result, we build our simulation environment from scratch. 

{Furthermore, we note that the Whittle's index could be negative. Therefore, we make modifications to the binary search routine of getting Whittle's index. Specifically, we let the initial lower bound be $-10$ to ensure a more accurate calculation of Whittle's index.}

We also note that in the original R$(\text{MA})^2$B algorithm, it assumes that they can reinitialize at any state for arbitrary number of times. This makes their exploration criteria, which will only stop when sufficient number of samples are collected for each agent, state, action pair, possible. We think this assumption is not impractical and do not use this structure. Therefore, we also change the algorithm so that the exploration period will end after the same number of time-steps instead of number of samples.

\paragraph{Randomly Generated Instances With 10 States }\label{paragraph:experiments-10-states}
 \begin{figure*}[ht]
       \centering
\includegraphics[width=\linewidth]{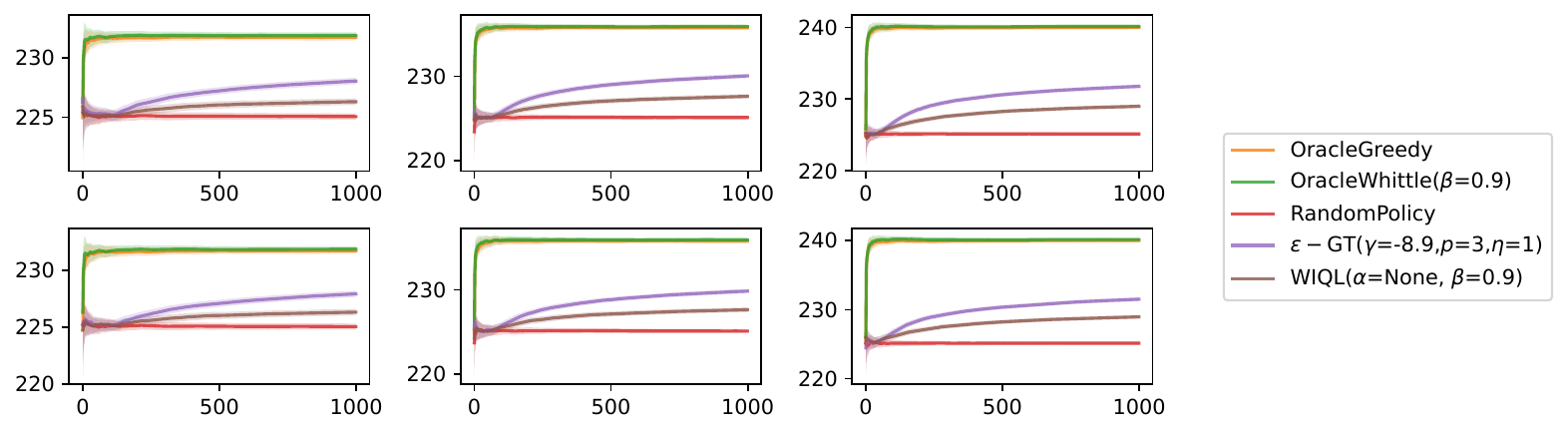}
        \caption{Average (over time) cumulativere reward for budgets $B=5$ (left), $10$ (middle), $20$ (right), averaged over 50 instances with 13 repetitions each. Top: noiseless reward; bottom: noisy reward. Ten states.}
        \label{fig:sub7}
         \vskip -0.1in
    \end{figure*}
    In this paragraph, we increase the size of the state space to 10 states to randomly generate 50 instances to test the performance of $\epsilon$-GT comparing to the benchmarks. Similarly, we also test two settings: 1) a noiseless case where the reward is deterministic, and 2) a noisy case where the rewards at state $s\in\mathcal{S}$ follow a normal distribution with mean $R^m(s)$ and variance 0.1. We would like to note that since UCWhittlePv and RRMABB behave like random policy for 2 states, we do not include that in the 10 states setting.

    As shown in Figure~\ref{fig:sub7}, $\epsilon$-GT with threshold $\gamma=-8.9$ consistently outperform WIQL in both noiseless and noisy setting. Further, we would like to emphasize that the gap between $\epsilon$-GT and WIQL is larger than the 2 states setting, demonstrating that our algorithm works well in larger states setting and adapting to the real-life setting better.
\subsection{Ablation Study}\label{subsec:ablation}
\begin{figure*}[ht]
       \centering
\includegraphics[width=\linewidth]{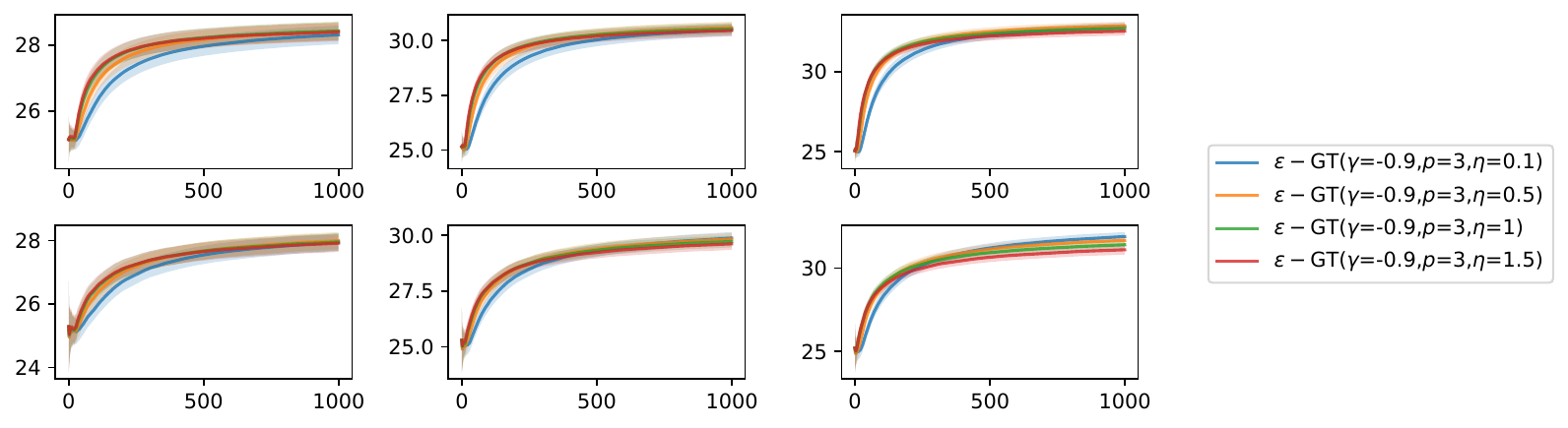}
        \caption{Average (over time) cumulative reward for budgets $B=5$ (left), $10$ (middle), $20$ (right), averaged over 50 binary state instances with 13 repetitions each. Top: noiseless reward; bottom: noisy reward.}
        \label{fig:sub9}
         \vskip -0.1in
    \end{figure*}

\begin{figure*}[ht]
       \centering
\includegraphics[width=\linewidth]{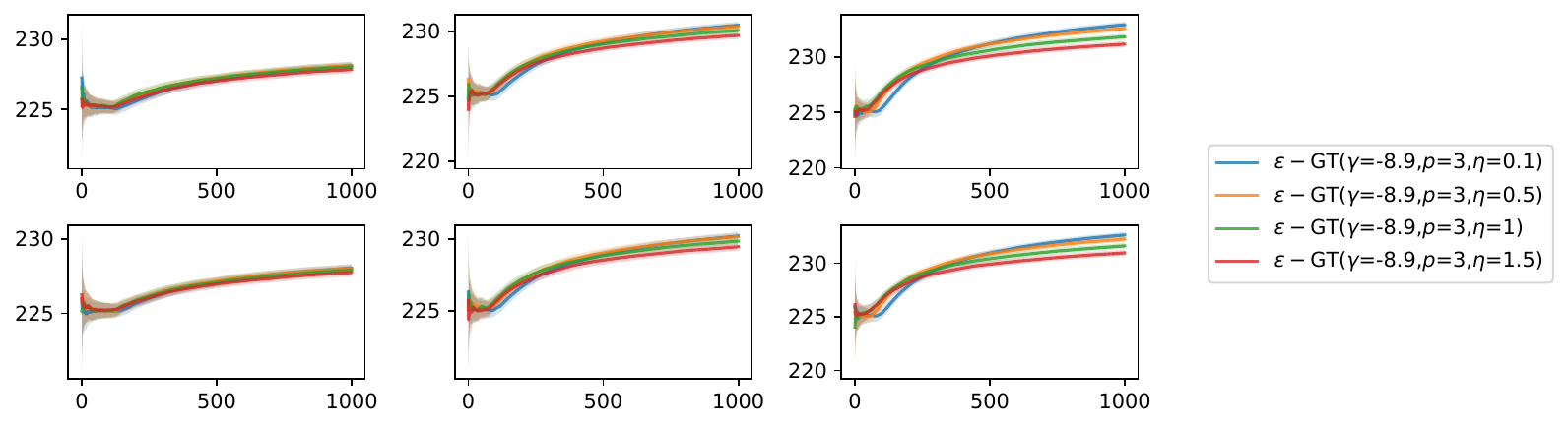}
        \caption{Average (over time) cumulative reward for budgets $B=5$ (left), $10$ (middle), $20$ (right), averaged over 50 10 state instances with 13 repetitions each. Top: noiseless reward; bottom: noisy reward.}
        \label{fig:sub12}
         \vskip -0.1in
    \end{figure*}

\begin{figure*}[ht]
       \centering
\includegraphics[width=\linewidth]{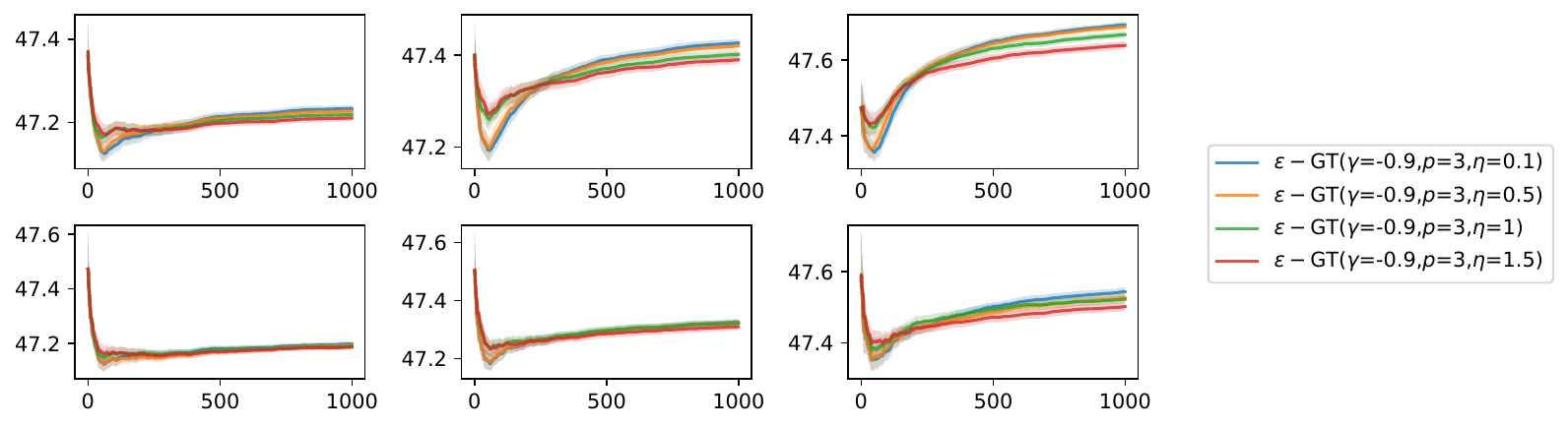}
        \caption{Average (over time) cumulative reward for budgets $B=5$ (left), $10$ (middle), $20$ (right), averaged over 100 repetitions of real dataset. Top: noiseless reward; bottom: noisy reward.}
        \label{fig:sub13}
         \vskip -0.1in
    \end{figure*}

\begin{figure*}[ht]
       \centering
\includegraphics[width=\linewidth]{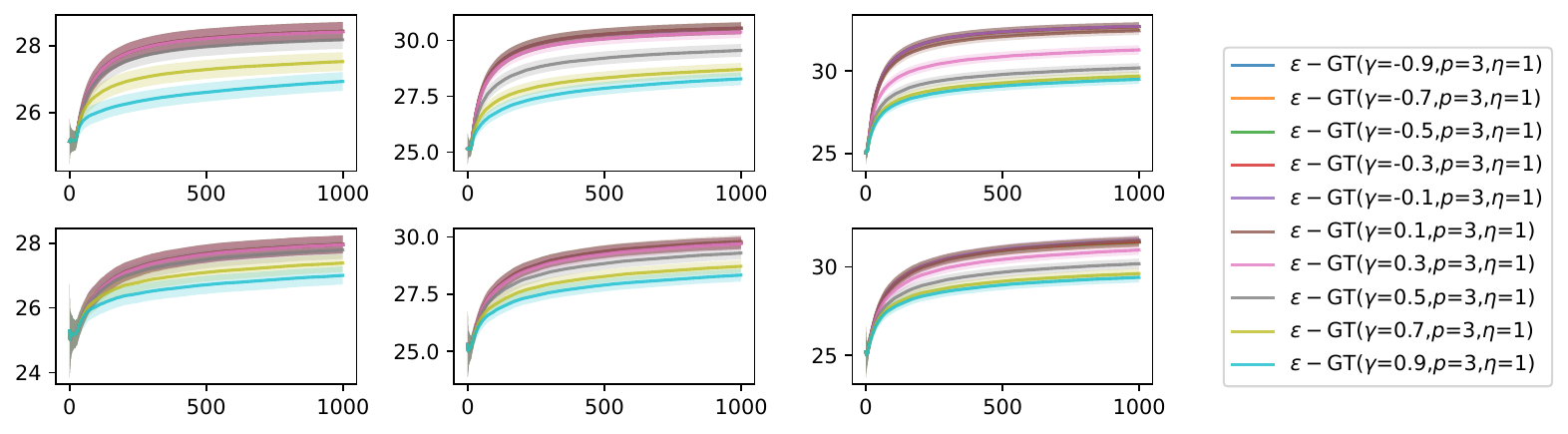}
        \caption{Average (over time) cumulative reward for budgets $B=5$ (left), $10$ (middle), $20$ (right), averaged over 50 binary state instances with 13 repetitions each. Top: noiseless reward; bottom: noisy reward. }
        \label{fig:sub8}
         \vskip -0.1in
    \end{figure*}

\begin{figure*}[ht]
       \centering
\includegraphics[width=\linewidth]{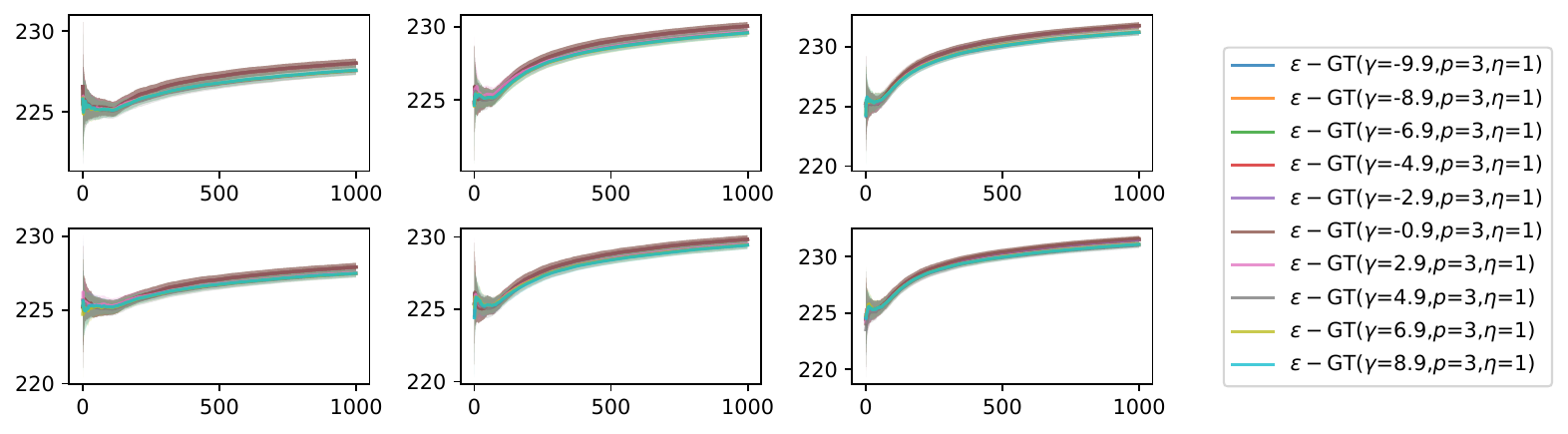}
        \caption{Average (over time) cumulative reward for budgets $B=5$ (left), $10$ (middle), $20$ (right), averaged over 50 ten state instances with 13 repetitions each. Top: noiseless reward; bottom: noisy reward. }
        \label{fig:sub10}
         \vskip -0.1in
    \end{figure*}

\begin{figure*}[ht]
       \centering
\includegraphics[width=\linewidth]{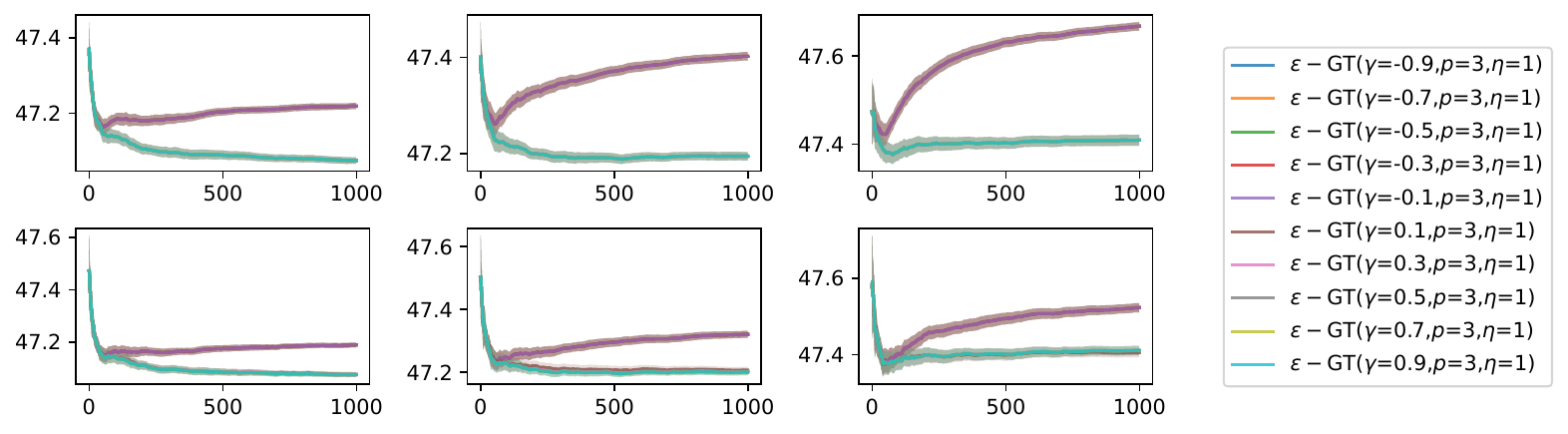}
        \caption{Average (over time) cumulative reward for budgets $B=5$ (left), $10$ (middle), $20$ (right), averaged over 100 repetitions of real dataset. Top: noiseless reward; bottom: noisy reward.}
        \label{fig:sub11}
         \vskip -0.1in
    \end{figure*}

In this section, we will test the performance of Alg.~\ref{alg:lcb-greedy} with different choices of $\eta$ and $\gamma$. 
\paragraph{Choice of $\eta$:}
We will choose four choices of $\eta\in\{0.1,0.5,1,1.5\}$. We will fix the $\gamma$ and $p$ to align with the choice of $\gamma$ used for experiments in Section~\ref{sec:experiments}. As shown in Figures~\ref{fig:sub9}-\ref{fig:sub13}
 Alg.~\ref{alg:lcb-greedy} is stable over different choices of $\eta$ for all of the synthetic binary state, ten state instances and real instance we have tested.
\paragraph{Choice of $\gamma$}\label{paragraph:experiments-gamma}
In this paragraph, we will compare $\epsilon-$GT with different threshold. We will fix other hyper-parameters to align with choices used in Section~\ref{sec:experiments}. As shown in Figure~\ref{fig:sub8}-Figure~\ref{fig:sub11}, for all instances, as $\gamma$ increases, the convergence rate of Alg.~\ref{alg:lcb-greedy} is slower. This translates to a lower average reward. Also, we observe that among all instances, a low threshold that satisfies Assump.~\ref{asssumption:enough-good-arm} will achieve the best performance.

\end{document}